\documentclass[a4paper,11pt,twoside,openright]{report}

\usepackage[utf8]{inputenc}
\usepackage[T1]{fontenc}
\usepackage[english, polish]{babel}
\usepackage{multirow}
\usepackage{makecell}
\usepackage{array}
\usepackage{appendix}

\makeatletter
\AtBeginEnvironment{appendices}{
  \clearpage
  \begingroup
    \let\ps@plain\ps@empty
    \appendixpage
  \endgroup
  \renewcommand{\thesection}{\Roman{section}}
  
  \let\tf@toc\tf@app
  \addtocontents{app}{\protect\setcounter{tocdepth}{1}}
  \immediate\write\@auxout{\string\let\string\tf@toc\string\tf@app^^J}
}
\makeatother

\usepackage{amsmath, amsfonts, amsthm, latexsym} 

\usepackage[final]{pdfpages} 

\usepackage{csquotes} 
\usepackage[
  authordate,
  backend=biber,
  maxbibnames=3,     
  minbibnames=1,
  maxcitenames=2,    
  giveninits=true,   
  doi=true,
  url=true,
  isbn=false,
  eprint=true
]{biblatex-chicago}
\AtEveryBibitem{%
  \clearfield{issn}%
  \clearfield{isbn}%
  \clearfield{series}%
  \clearfield{month}%
  \clearfield{urldate}%
  \clearfield{note}
  \clearfield{address}%
  \clearlist{location}
}

\AtEveryBibitem{%
  \iffieldundef{doi}{}{%
    \clearfield{url}%
  }%
}

\usepackage{commath} 

\usepackage{xurl}
\usepackage{graphicx}
\usepackage{alltt}
\usepackage{lmodern}
\usepackage{booktabs}
\usepackage{longtable}
\usepackage{array}
\usepackage{tabularx}
\usepackage{microtype}
\usepackage{float}
\usepackage[useregional]{datetime2}
\usepackage{multirow}
\usepackage{subcaption}
\usepackage{enumitem}
\usepackage{arydshln} 
\usepackage{setspace}

\usepackage{algorithm}
\usepackage{algpseudocode}
\usepackage{adjustbox}
\usepackage{listings}
\usepackage{xcolor}
\usepackage{nameref}

\renewcommand{\appendixtocname}{List of appendices}

\makeatletter
\let\oldappendix\appendices

\renewcommand{\appendices}{%
  \clearpage
  \renewcommand{\thesection}{\Roman{section}}
  \let\tf@toc\tf@app
  \addtocontents{app}{\protect\setcounter{tocdepth}{1}}
  \immediate\write\@auxout{%
    \string\let\string\tf@toc\string\tf@app^^J
  }
  
  \let\originalpsplain\ps@plain      
  \let\ps@plain\ps@empty             
  \oldappendix                       
  \let\ps@plain\originalpsplain      
}%

\newcommand{\listofappendices}{%
  \begingroup
  \renewcommand{\contentsname}{\appendixtocname}
  \let\@oldstarttoc\@starttoc
  \def\@starttoc##1{\@oldstarttoc{app}}
  \tableofcontents 
  \endgroup
}

\makeatother

\usepackage[inner=20mm, outer=20mm, bindingoffset=10mm, top=25mm, bottom=25mm]{geometry} 

\allowdisplaybreaks         

\usepackage{indentfirst}    
\usepackage{fancyhdr}
\fancypagestyle{plain}{
  \fancyhf{}
  \fancyfoot[LE,RO]{\thepage}

}

\usepackage{titlesec}
\titleformat{\chapter}
  {\normalfont\Large \bfseries}
  {\thechapter.}{1ex}{\Large}

\titleformat{\section}
  {\normalfont\large\bfseries}
  {\thesection.}{1ex}{}
\titlespacing{\section}{0pt}{30pt}{20pt}

\titleformat{\subsection}
  {\normalfont \bfseries}
  {\thesubsection.}{1ex}{}

\def\cleardoublepage{\clearpage\if@twoside
\ifodd\c@page\else\hbox{}\thispagestyle{empty}\newpage
\if@twocolumn\hbox{}\newpage\fi\fi\fi}

\usepackage{etoolbox}
\makeatletter
\patchcmd{\l@chapter}
  {\hfil}
  {\leaders\hbox{\normalfont$\m@th\mkern \@dotsep mu\hbox{.}\mkern \@dotsep mu$}\hfill}
  {}{}
\makeatother

\usepackage{titletoc}
\makeatletter
\titlecontents{chapter}
  [0pt]
  {}
  {\bfseries \thecontentslabel.\quad}
  {\bfseries}
  {\bfseries\leaders\hbox{\normalfont$\m@th\mkern \@dotsep mu\hbox{.}\mkern \@dotsep mu$}\hfill\contentspage}

\titlecontents{section}
  [1em]
  {}
  {\thecontentslabel.\quad}
  {}
  {\leaders\hbox{\normalfont$\m@th\mkern \@dotsep mu\hbox{.}\mkern \@dotsep mu$}\hfill\contentspage}

\titlecontents{subsection}
  [2em]
  {}
  {\thecontentslabel.\quad}
  {}
  {\leaders\hbox{\normalfont$\m@th\mkern \@dotsep mu\hbox{.}\mkern \@dotsep mu$}\hfill\contentspage}
\makeatother

\makeatletter
\newtheoremstyle{definition}
{3ex}
{3ex}
{\upshape}
{}
{\bfseries}
{.}
{.5em}
{\thmname{#1}\thmnumber{ #2}\thmnote{ (#3)}}
\makeatother

\theoremstyle{definition}

\usepackage[hidelinks, pdftex]{hyperref} 

\newcommand{\tytul}{Łączenie tradycyjnych metod wyjaśnialności z multimodalnymi i wielojęzycznymi modelami: analiza oparta na XAI}
\renewcommand{\title}{Bridging Traditional Explainability Methods and Multimodal Multilingual Models: An XAI-Based Analysis}

\begin{document}
\selectlanguage{english}

\includepdf[pages=-]{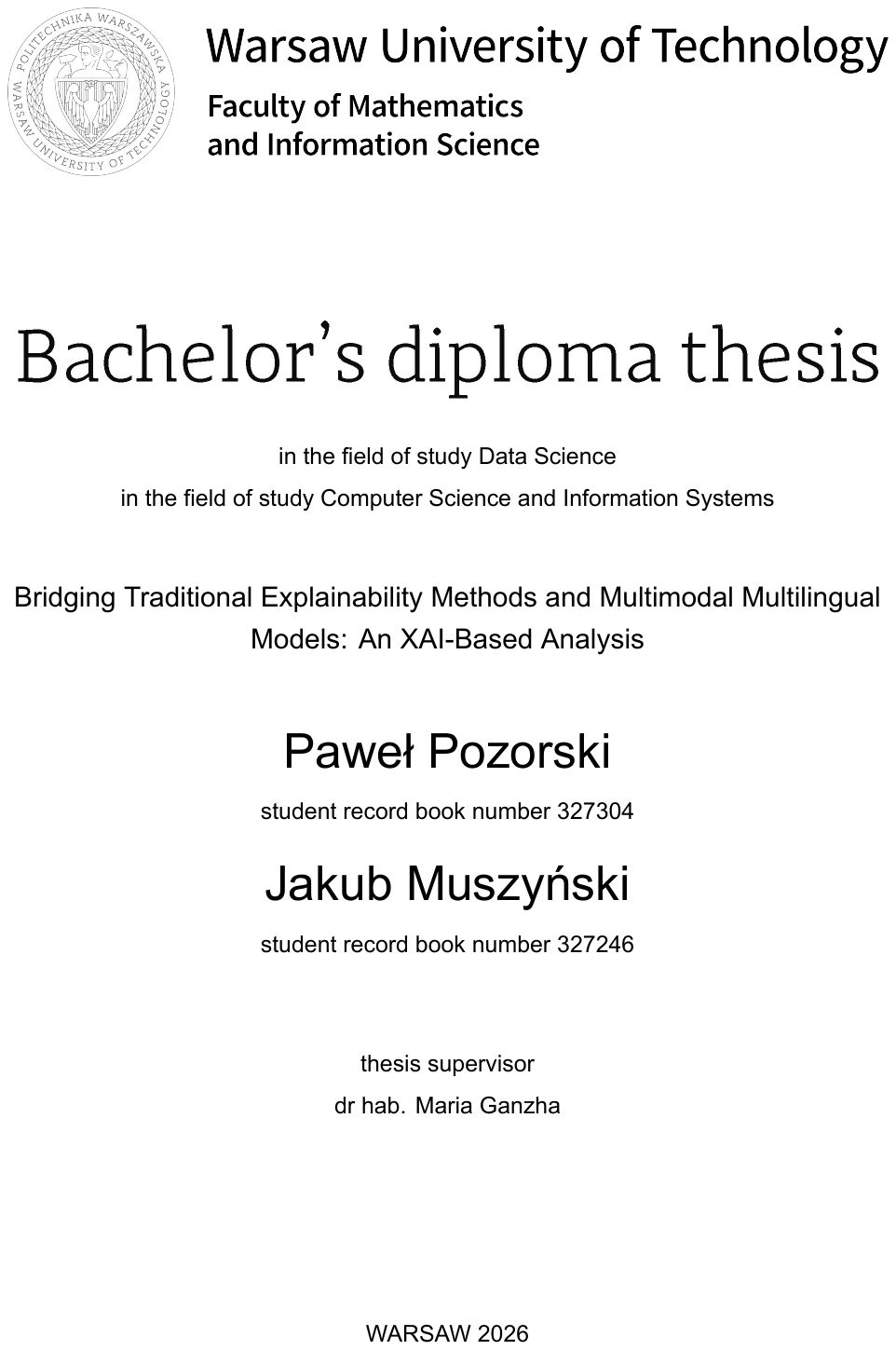} 

\null\thispagestyle{empty}\newpage


%
%
%


{ \fontsize{12}{14} \selectfont
\begin{abstract}

\begin{center}
\title
\end{center}

Multimodal Large Language Models (MLLMs) can integrate text and audio to interpret context in interactive conversations. However, the mechanisms by which information from different modalities shapes model behavior remain difficult to analyze. Shapley values (SV) are widely used for local, model-agnostic explainability in simple text-based conversations, yet their direct application to multimodal data is nontrivial due to cross-channel dependencies, dialogue structure, and the prohibitive computational cost of native audio tokenization.

This work introduces a multimodal extension of Shapley values (SV), where units of information -- such as text tokens and audio segments -- are treated as cooperative features. To make the approach feasible under real-world constraints, we pair this formulation with efficient estimation methods: exact Shapley computation for short inputs, and sampling-based approximations using Monte Carlo permutation and stratified sampling with Neyman allocation, balancing variance against a strict computational budget. Additionally, to address the granularity mismatch between text and audio, we propose Spectrogram-Guided Phonetic Alignment (SGPA), a pre-processing method that maps dense audio streams to interpretable, word-aligned segments.

As an applied contribution, we provide a model-agnostic Python package for computing and visualizing multimodal Shapley values for text and audio. A companion GUI enables interactive inspection of attributions, side-by-side modality visualization, and method-specific estimates of computational cost. Furthermore, we curate resources derived from the VoiceBench and Infinity Instruct datasets, encompassing diverse modality configurations and multilingual scenarios. These resources are used in validation experiments which demonstrate that input modality appears to be a significant driver of attribution volatility, while syntactic importance proxies often fail to predict model attention across languages. \\

\end{abstract}
}

\null\thispagestyle{empty}\newpage

{\selectlanguage{polish} \fontsize{12}{14}\selectfont
\begin{abstract}

\begin{center}
\tytul
\end{center}

Wielomodalne duże modele językowe (MLLM) łączą tekst i dźwięk, aby interpretować kontekst w interaktywnych konwersacjach. Do tej pory zrozumienie, w jaki sposób informacje pochodzące z różnych modalności wpływają na zachowanie modelu, pozostaje mocno ograniczone. Wartości Shapleya (SV) są powszechnie stosowane jako lokalna, niezależna od modelu metoda wyjaśnialności w prostych rozmowach tekstowych, jednak ich bezpośrednie zastosowanie w środowisku wielomodalnym jest nietrywialne ze względu na zależności między różnymi kanałami, strukturę dialogu oraz wysoki koszt obliczeniowy natywnej tokenizacji audio.

W niniejszej pracy przedstawiono wielomodalne rozszerzenie wartości Shapleya, w którym jednostki informacji, takie jak tokeny tekstowe i segmenty dźwięku, są traktowane jako współpracujące cechy. Aby uczynić to podejście praktycznym, połączono je z wydajnymi metodami estymacji: dokładnym obliczaniem wartości Shapleya dla krótkich wejść oraz aproksymacjami opartymi na próbkowaniu Monte Carlo i stratyfikowanym próbkowaniu z alokacją Neymana. Dodatkowo, aby rozwiązać problem różnicy w ziarnistości między tekstem a dźwiękiem, zaproponowano metodę Spectrogram-Guided Phonetic Alignment (SGPA), która mapuje gęste strumienie audio na interpretowalne segmenty wyrównane do słów.

Jako wkład praktyczny opracowano pakiet w języku Python, umożliwiający obliczanie i wizualizację wielomodalnych wartości Shapleya niezależnie od wybranego modelu. Towarzyszący mu interfejs graficzny (GUI) pozwala na interaktywną analizę atrybucji, wizualizację modalności oraz estymację kosztów obliczeniowych. Ponadto przygotowano zbiory danych oparte na zestawach VoiceBench i Infinity Instruct, obejmujące różne konfiguracje modalne i scenariusze wielojęzyczne. Wykorzystano je w eksperymentach walidacyjnych, które wykazały, że modalność wejściowa jest znaczącym czynnikiem wpływającym na zmienność atrybucji, a syntaktyczne miary ważności słów często nie pokrywają się z uwagą modelu w różnych językach.

\noindent \textbf{Słowa kluczowe:} Wielomodalne duże modele językowe, Wielojęzyczne duże modele językowe, Wartości Shapleya, Wyjaśnialna sztuczna inteligencja
\end{abstract}
}

%
%
%
\null\thispagestyle{empty}\newpage
\pagenumbering{gobble}
\tableofcontents
\thispagestyle{empty}


\null\thispagestyle{empty}\newpage
\pagestyle{fancy}
\pagenumbering{arabic}
\setcounter{page}{11}

%

\chapter*{Artificial Intelligence usage declaration}
\label{sec:ai_usage}

We hereby declare that:

\begin{enumerate}[label=\textbf{\arabic*)}]
    \item We \textbf{have} used IT tools to generate the content of the manuscript of this thesis.
    
    \item We \textbf{have} used IT tools to generate the code of the software developed for this thesis.

    \item We take full responsibility for all content in the thesis, including both the manuscript and the software developed for it.
\end{enumerate}

\begin{table}[htbp]
    \caption*{Artificial Intelligence (AI) usage details.}
    \centering
    \renewcommand{\arraystretch}{1.4}
    \begin{tabularx}{\textwidth}{@{} >{\raggedright\arraybackslash}X >{\raggedright\arraybackslash}X @{}}
        \toprule
        \textbf{The scope of the use of IT tools to generate manuscript} & \textbf{The scope of the use of IT tools to generate software code} \\
        \midrule
        \begin{itemize}[leftmargin=*, nosep, topsep=0pt, before=\vspace{-\baselineskip}]
            \item Replacement of a domain linguistic specialist to support ourselves in the validation of experiment results from a linguistic perspective.
        \end{itemize} & 
        \begin{itemize}[leftmargin=*, nosep, topsep=0pt, before=\vspace{-\baselineskip}]
            \item Support in writing unit tests.
            \item Suggestions for refactoring code for Graphical User Interface (GUI) frontend and \textit{mllm-shap} package.
            \item Additional (to ours) reviews within GitHub Pull Request.
        \end{itemize} \\
        \bottomrule
    \end{tabularx}
\end{table}

\newpage \ \null\thispagestyle{empty}\newpage

\chapter{Introduction}
\label{sec:introduction}

In recent years, intelligent systems capable of interacting with humans through text and voice have become vital tools for organizations aiming to reduce operational costs and improve customer accessibility. From in-house chatbots that streamline employee onboarding, to text-based virtual assistants replacing traditional customer service teams and voice-driven support lines, such technologies now play a central role in modern business operations. They deliver scalable, cost-efficient, and always-available services that can be seamlessly adapted to evolving organizational needs.

Research aimed at improving large language models (LLMs) has focused largely on expanding domain knowledge and mitigating hallucinations and other reasoning errors \parencite{llmdirectionsoverview}. Techniques such as model fine-tuning, Retrieval-Augmented Generation (RAG) \parencite{lewis2021retrievalaugmentedgenerationknowledgeintensivenlp}, Prompt Engineering \parencite{schulhoff2025promptreportsystematicsurvey}, Few-Shot Prompting \parencite{brown2020languagemodelsfewshotlearners}, and Chain-of-Thought (CoT) \parencite{wei2023chainofthoughtpromptingelicitsreasoning} have demonstrated considerable potential. However, evaluating their real-world impact remains challenging, as many corner cases and inconsistencies frequently persist in production environments.

Understanding the factors that influence model outputs is therefore essential for improving reliability and ensuring the safe integration of LLMs into business-critical systems. Given the costs associated with training and adapting LLMs for specific tasks, it is crucial to analyze how changes in a model's internal knowledge affect its final outputs. Consequently, comprehensive interpretability and robustness analyses are vital for enabling LLMs to serve as cost-effective, state-of-the-art, and trustworthy alternatives to human decision-makers.

Addressing this challenge requires more than algorithmic innovation -- it demands a complete \emph{research instrument} capable of systematic, reproducible study. While Shapley values (SV), popularized in machine learning by \textcite{lundberg2017unifiedapproachinterpretingmodel}, provide a principled framework for identifying which parts of an input exert the greatest influence on model outputs, applying them to multimodal LLMs (MLLMs) that process both text and audio introduces distinct technical barriers: exponential computational complexity, mismatched granularity across modalities, and the need for reproducible execution under hardware constraints.

\newpage
This work develops such an instrument -- an integrated explainability platform comprising three core components: (1) a model-agnostic Python package (\texttt{mllm-shap}) implementing efficient SV approximation methods adapted for variable-length multimodal sequences, (2) reproducible experimental infrastructure with configuration management, checkpointing, and artifact versioning, and (3) an interactive web interface enabling near real-time attribution visualization and session management. The platform addresses the audio segmentation challenge through Spectrogram-Guided Phonetic Alignment (SGPA), a novel preprocessing method that reduces coalition space by 10--50$\times$ while preserving semantic interpretability, making previously intractable computations feasible on consumer hardware.

The package integrates seamlessly with the \texttt{transformers} framework \parencite{wolf2020transformers} and is publicly available via \texttt{pip}\footnote{\url{https://pypi.org/project/mllm-shap/}}. It implements multiple SV approximation strategies, including Monte Carlo sampling by \textcite{goldshmidt2024tokenshapinterpretinglargelanguage} and stratified allocation via Neyman methods \parencite{shapleyapproximations, nayman}. To our knowledge, this is the first work to provide a complete, reproducible framework for SV-based explainability of text and audio MLLMs.

To validate the platform's capabilities, we conduct controlled experiments across modalities (text-only, audio-only, mixed-modality) and languages (English, Spanish, French) on the LFM2-Audio-1.5B model (Sections~\ref{subsec:experiments__multi_modal_context}--\ref{subsec:experiments__multi_language_context}). While hardware constraints limited this initial study to a single model, these experiments demonstrate both the platform's technical feasibility and its utility for characterizing MLLM behavior -- establishing proof-of-concept findings on how input modality and language affect attribution patterns.

The analysis focuses on direct audio-to-audio models, which operate on native or latent audio tokens rather than cascaded speech $\rightarrow$ text $\rightarrow$ speech architectures \parencite{yang2025largelanguagemodelsmeet, Lee_2022}. This design choice ensures experimental consistency and reflects current industry trends toward end-to-end multimodal processing. Although other modalities such as images have been investigated in prior SV work \parencite{goldshmidt2024tokenshapinterpretinglargelanguage}, the focus on text and audio reflects both market priorities for voice-enabled systems and the specific technical challenges of aligning discrete text units with continuous audio signals.

The remainder of this work is organized as follows. Chapter~\ref{sec:business_goal} establishes the business case for multimodal explainability, while Chapter~\ref{sec:literature} reviews related work on XAI for generative models. Chapter~\ref{sec:requirements} presents specific challenges related to the topic and general approach to solving them. Chapter~\ref{sec:theory} presents the theoretical foundations, and Chapter~\ref{sec:solution} describes the platform implementation. Finally, Chapters~\ref{sec:experiments_setup} through~\ref{sec:experiments_results} document the experimental protocol and findings, with conclusions in Chapter~\ref{sec:summary}.

\chapter{Motivation}
\label{sec:business_goal}

The increasing reliance on intelligent systems that interpret and generate human communication has made transparency in their decision-making processes a critical business concern. Organizations are deploying Large Language Models (LLMs) and their multimodal variants (MLLMs) across domains such as customer support, healthcare, and education. As these systems evolve from purely textual models into multimodal frameworks capable of processing speech, audio, and other continuous signals, their internal mechanisms grow more complex and harder to interpret. 

Focusing on end-user deployment performance, many companies invest substantial effort and resources into enhancing models’ reasoning capabilities and domain-specific knowledge -- either through the techniques discussed in Chapter~\ref{sec:introduction} or by enabling models to interact with internal systems via various tools (e.g., see \textcite{chen-etal-2024-towards-tool}). These improvements, however, are typically made under the assumption that LLMs reason in the same way humans do -- that is, that models can be provided with knowledge and engaged through interaction much like a human counterpart. Validating this assumption remains an open challenge that must be addressed to ensure such enhancements are performed safely, cost-effectively, and successfully. In the absence of model explainability, the most reliable way to assess whether a new prompt performs better than an existing one is through A/B testing -- an approach that is both expensive and time-consuming. Similar limitations apply to other enhancement techniques as well.

From a business perspective, lack of explainability introduces operational and strategic risks beyond those related to enhancing models' performance. Decisions made by these systems can affect customer experience, regulatory compliance, and even safety-critical operations. Without insight into how a model weighs input features or interprets multimodal context, organizations face challenges in auditing performance, mitigating biases, and ensuring accountability. Explainability, therefore, is not just a research goal -- it is essential for trustworthy deployment, regulatory compliance, and long-term value creation.

\newpage
MLLMs integrate multiple sensory and linguistic streams -- such as text and audio -- to construct richer contextual understanding. While this integration enhances performance, it simultaneously increases model opacity. Unlike unimodal systems, MLLMs may exhibit subtle cross-modal failures, including inconsistencies between modalities or disproportionate reliance on a single channel, which are considerably harder for human supervisors to detect and interpret. Understanding these interactions is therefore essential for developing systems that remain robust across languages, environments, and modes of communication.

Explainability also benefits end users and researchers. In Retrieval-Augmented Generation (RAG) systems, it allows tracing the model’s reasoning back to specific source data, offering deeper insights beyond a mere list of retrieved documents.

Tools based on Shapley values (SV) offer quantitative and visual insights into how each modality contributes to model predictions. They allow developers, researchers, and business stakeholders to diagnose behavior, uncover sources of bias, and optimize performance. Such explainability forms the foundation for improving reliability, refining data pipelines, and aligning model outputs with human expectations.

The platform described in this work was designed to bridge the gap between technical explainability and practical business application. By implementing explainability directly within an interactive platform, the project enables:

\begin{itemize}
    
    \item \textbf{Transparency:} Real-time interpretability of MLLM outputs builds user and stakeholder trust in AI-driven decisions.
    
    \item \textbf{Accountability:} Quantifiable attributions allow organizations to trace outcomes to specific inputs, aiding in auditability and compliance.
    
    \item \textbf{Optimization:} Insights from explainability analyses support data quality improvements and fine-tuning strategies, reducing cost and improving performance consistency.
    
    \item \textbf{Scalability:} By standardizing explainability workflows for text and audio modalities, the system establishes a foundation for integrating explainable AI practices into broader enterprise pipelines.
    
\end{itemize}

In summary, the business goal behind this research is to enhance the interpretability of next-generation language systems that operate beyond text. Through the development 
of an explainability platform for MLLMs, this work provides a framework for responsible innovation -- one that transforms complex neural architectures into transparent, measurable, and trustworthy tools for real-world applications.

\chapter{Literature Overview}
\label{sec:literature}

As machine learning models are increasingly deployed in high-stakes domains like healthcare, finance, and law, the need for Explainable Artificial Intelligence (XAI) has become critical. These models often operate as ``black boxes'', making it challenging for end users to comprehend decision-making processes. By providing interpretable explanations, XAI bridges the gap between complex model logic and human reasoning to ensure trust and accountability \parencite{explainableaihistory}.

Explanations allow stakeholders to validate decisions, detect bias, ensure compliance with regulatory standards, and enhance human–machine collaboration \parencite{explainableaihistory}.

These considerations are especially critical for Large Language Models (LLMs), which pose additional challenges due to their extreme complexity and limited reproducibility of results. As highlighted by \textcite{herrera2025makingsenseunsensiblereflection}, XAI may become a foundational requirement for deploying LLMs in high-risk, human-centered environments subject to strict legal regulations.

The rapid evolution of LLM architectures and the emergence of direct multimodal processing have necessitated a literature review focused primarily on the period between 2023 and 2025. This narrow temporal window is intentional; the field of XAI for generative models is currently undergoing a paradigm shift where traditional methods are being fundamentally reassessed or replaced by techniques capable of handling the high-dimensional, variable-length nature of modern transformer-based systems. Consequently, the following sections prioritize state-of-the-art developments that address these contemporary challenges, while relevant foundational literature is revisited in subsequent chapters of this thesis to provide specific theoretical grounding.

The complexity of LLMs, their training processes, and the tasks they are applied to make LLM-based systems susceptible to various challenges, including hallucinations \parencite{Huang2025}. Their growing popularity -- both in applications and research -- has empowered the development of numerous explainability methods \parencite{zhao2023explainabilitylargelanguagemodels}. Many of these methods are tied to model architecture or system behavior, including Chain-of-Thought (CoT) and local explanation techniques based on attention analysis.

\newpage
While system-aware techniques can provide more informative and computationally efficient explanations, this work focuses on black-box approaches, interacting only through inputs and outputs. This enables a versatile, architecture-agnostic, and easily extensible package.

Among popular black-box XAI methods are Local Interpretable Model-agnostic Explanations (LIME) and SHapley Additive exPlanations (SHAP) \parencite{explainableaihistory}. Although LIME has been successfully applied to LLMs in some cases \parencite{dehghani2025explaininglargelanguagemodels}, it introduces additional user-side complexity. Generating perturbations of the original input to create similarity-weighted samples for the surrogate model requires selecting both a perturbation algorithm and a similarity measure. To maintain a lightweight and user-friendly system with minimal hyperparameters, this work focuses on SHAP as the primary method for black-box explainability. While questions of output similarity -- analogous to LIME’s similarity measures for input perturbations -- remain, the overall user-side algorithmic complexity is reduced.

SVs have proven effective in general XAI applications \parencite{lundberg2017unifiedapproachinterpretingmodel}, though efforts to apply them to LLMs have yielded mixed outcomes, with only limited success reported by \textcite{explainableaihistory}. More recent approaches, such as \textcite{goldshmidt2024tokenshapinterpretinglargelanguage}, demonstrate stronger alignment with human judgments and generate more interpretable outputs. Building on these foundations, the present work extends Shapley-based explainability to multi-turn conversational LLMs capable of processing both audio and text modalities.

Efficient SV approximation is crucial for practical deployment, as Multimodal LLMs (MLLMs) are computationally expensive. Without approximations that reduce model calls while maintaining accuracy and low variance, these methods cannot support near real-time applications.

A Monte Carlo (MC)-based approach was proposed by \textcite{goldshmidt2024tokenshapinterpretinglargelanguage}, where, depending on the sampling budget, all single-token combinations and a random subset of all possible token subsets are evaluated. Including first-order combinations (where exactly one player is removed from the grand coalition, see Chapter~\ref{sec:shapley_values}) as well as empty masks and full masks, allowed the authors to achieve $90\%$ accuracy with only $30\%$ of the total $2^n - 1$ possible combinations, providing a significant performance boost (See Section~\ref{subsec:shapley_values__monte_carlo} for details).

A more promising approach was proposed by \textcite{shapleyapproximations}. Instead of computing each player's marginal contribution to all possible coalitions they employ complementary contributions (CC) of each coalition to all remaining players outside that coalition, a formulation equivalent to the traditional SV definition. The CC of a coalition $S$ can be reused to compute the SV for all players within that coalition. This property, combined with a strategic allocation of coalitions to be evaluated using the Neyman method \parencite{nayman}, yields scalability far beyond Monte Carlo approaches while maintaining lower variance.

\chapter{Challenges and Design Constraints}
\label{sec:requirements}

This work develops a research instrument -- a software platform designed to make Shapley value (SV) analysis of multimodal language models \emph{feasible, reproducible, and scalable}. The core challenge is straightforward: computing exact SV for realistic inputs is mathematically intractable, yet research validity demands methodological rigor. This chapter outlines the constraints that shaped our design and the requirements that emerged from building a tool capable of supporting systematic empirical study.

Unlike ad-hoc research scripts, an instrument must be \emph{reusable beyond the initial study}. The decisions documented here -- from segmentation strategies to approximation methods to infrastructure design -- reflect a single goal: enabling controlled, reproducible experiments on multimodal attribution at a scale that would otherwise be impossible.

Only after satisfying these rigorous criteria did we conduct the initial research phase, applying the framework to validate whether SV can effectively characterize model behavior. While these experiments offer compelling insights and possess intrinsic value as an exploration of Multimodal Large Language Models (MLLMs) explainability, current hardware limitations necessitated a restricted scope, preventing the attainment of broadly generalizable results at this stage. Nevertheless, this work was a primary objective of our inquiry; it serves as a vital proof-of-concept that demonstrates the utility of our approach and establishes the precise methodological groundwork required for more resource-intensive, large-scale studies in the future.

\section{The Fundamental Constraint: Exponential Complexity}
\label{subsec:requirements__exponential}

Shapley value computation requires evaluating all $2^n$ subsets of input features. For a $50$-token prompt, this means $10^{15}$ model evaluations. Even at 1 millisecond per call, exact computation would take 30 years.

This mathematical fact forces three foundational design choices:

\begin{enumerate}[nosep]
    \item \textbf{Approximation is unavoidable.} We must use sampling-based estimators (Section~\ref{sec:shapley_values}) and rigorously validate their accuracy-cost trade-offs (Section~\ref{subsec:shapley_values__comparison}).
    
    \item \textbf{Segmentation determines feasibility.} Defining what counts as a ''feature'' changes $n$ by orders of magnitude. For audio, native tokenization yields 100+ units per sentence; our Spectrogram-Guided Phonetic Alignment (SGPA) method (Chapter~\ref{sec:experiments__sgpa}) reduces this to $5$--$10$ word-aligned segments -- the difference between intractable and feasible.
    
    \item \textbf{Cost must be measurable.} Research requires knowing not just ''what is the result'' but ''how much did it cost to compute.'' Every experiment logs evaluation counts, wall-clock time, and approximation parameters (Section~\ref{sec:experiments__infrastructure}).
\end{enumerate}

\section{The Audio Challenge: When Native Tokenization Breaks}
\label{subsec:requirements__audio}

While text discretizes naturally into semantically grounded units (e.g., words or sub-words), raw audio lacks inherent semantic boundaries. A standard $3$-second utterance typically transcribes to roughly $10$ words, yet produces over $150$ encoder frames under native model tokenization. This discrepancy creates a fundamental bottleneck for interpretability. 

Treating raw encoder frames as independent features leads to three primary failures:
\begin{itemize}
    \item \textbf{Dimensionality Explosion:} The coalition space for Shapley value (SV) computation increases from $2^{10}$ to $2^{150}$ (a factor of $\approx 10^{42}$), rendering exact calculation impossible.
    \item \textbf{Semantic Dilution:} Individual audio frames (often $20$ms) lack standalone meaning; resulting attributions become granular ''noise'' rather than interpretable insights.
    \item \textbf{Cross-Modal Incompatibility:} Direct comparison between text and audio explanations becomes invalid due to the misalignment of their underlying units.
\end{itemize}

To bridge this gap, we introduce Spectrogram-Guided Phonetic Alignment (SGPA) (see Section~\ref{sec:experiments__sgpa}). Rather than utilizing native tokenization, SGPA acts as a specialized preprocessing layer that segments audio into word-level units using acoustic cues and forced alignment. This provides three key advantages:
\begin{enumerate}[nosep]
    \item \textbf{Tractability:} Reduces the feature count $n$ by a factor of $10$--$50$, enabling computationally feasible experiments (Table~\ref{tab:experiments__sgpa__performance}).
    \item \textbf{Alignment:} Synchronizes audio segments with text units to facilitate valid cross-modal comparisons.
    \item \textbf{Interpretability:} Produces attributions that correspond to recognizable phonetic and linguistic structures.
\end{enumerate}

\newpage
It is important to note that SGPA fundamentally alters the ''cooperative game'' under analysis. By computing SV over word-aligned segments rather than native encoder frames, we acknowledge a deliberate trade-off: we prioritize \textit{interpretability} and \textit{tractability} over a ''pure'' model-native representation. The implications and potential biases introduced by this segmentation are addressed in Section~\ref{subsec:experiments__sgpa__diagnostics}.

\section{The Reproducibility Challenge: Infrastructure for Science}
\label{subsec:requirements__reproducibility}

To ensure scientific validity, experimental results must be independently verifiable. Our infrastructure is designed to mitigate the ''black box'' nature of deep learning research by enforcing three pillars of reproducibility. For details, refer to Section~\ref{sec:experiments__infrastructure}.

\paragraph{Pillar 1: Comprehensive Run Specification}
A result is only meaningful if the environment that produced it is fully documented. Every experimental execution in our framework automatically generates a \texttt{spec.json} file capturing the complete provenance of the run: version control of resources, stochastic control for data sampling and model generation (random seeds), and the exact hyperparameters used for execution logic -- ranging from approximation methods and budgets to hardware-specific parameters.

\paragraph{Pillar 2: Fault-Tolerant, Stateful Execution}
Large-scale interpretability studies are prone to intermittent failures, such as Application Programming Interface (API) timeouts or graphic card out-of-memory errors. To prevent compute waste and avoid \textit{selection bias} -- where only short-running or ``simple'' samples successfully complete -- our system implements per-sample checkpointing. This allows experiments to resume from the last completed index with deterministic seed management, ensuring the final output is identical to a continuous, uninterrupted run.

\paragraph{Pillar 3: Decoupled Artifact Analysis}
To facilitate rapid iteration without re-executing expensive model calls, we strictly separate data collection from analysis. Raw outputs are stored as structured artifacts containing both the result and its associated metadata. This allows researchers to regenerate figures, perform statistical tests, or adjust visualization parameters using only these frozen artifacts, ensuring the analysis phase remains computationally cheap and non-destructive.

\section{From Constraints to Architecture}

The design requirements outlined in Appendix~\ref{app:requirements} are not independent silos; rather, they interact to define the platform’s foundational architecture. This holistic design allows the system to function as a cohesive research instrument where each component reinforces the others:

\begin{itemize}
    \item \textbf{Approximation Methods} (Section~\ref{sec:shapley_values}): Offer a controlled spectrum of cost-fidelity trade-offs, supported by rigorous error bounds and convergence diagnostics.
    
    \item \textbf{Spectrogram-Guided Phonetic Alignment (SGPA)} (Section~\ref{sec:experiments__sgpa}): Acts as the primary dimensionality reducer, transforming high-resolution audio into semantically grounded segments to make complex experiments computationally feasible.
    
    \item \textbf{Modular Package Design} (Section~\ref{sec:package}): Decouples model logic from attribution logic, enabling seamless ablation studies and objective cross-model comparisons.
    
    \item \textbf{Robust Infrastructure} (Section~\ref{sec:experiments__infrastructure}): Automates the overhead of reproducibility, logging, and fault tolerance, ensuring that experiments are verifiable at scale.
    
    \item \textbf{Systematic Validation} (Chapter~\ref{sec:experiments_results}): Demonstrates that the platform can reliably execute large-scale studies ($100$+ samples) within realistic time and compute constraints.
\end{itemize}

\noindent This framework is not merely ``research code'' written to execute a specific task; it is a specialized scientific instrument designed to enable a class of audio interpretability experiments that would otherwise be infeasible. The following chapters document the design and implementation of these components (Sections~\ref{sec:shapley_values}--\ref{sec:gui}), followed by empirical validation (Chapter~\ref{sec:experiments_results}) and a discussion of current limitations (Chapter~\ref{sec:summary}).

\chapter{Theory}
\label{sec:theory}

This chapter establishes the theoretical framework for our research. Section~\ref{sec:shapley_values} introduces Shapley Values (SV) alongside their approximation methods, providing the basis for our interpretability approach. Section~\ref{sec:experiments__sgpa} details Spectrogram-Guided Phonetic Alignment (SGPA), our novel audio tokenization algorithm. SGPA enables audio explainability while strictly adhering to the \textbf{NFR-1 (Usability)} and \textbf{NFR-2 (Performance)} requirements (refer to Appendix~\ref{app:requirements}).

\section{Shapley Values for Multimodal Large Language Models}
\label{sec:shapley_values}

Shapley values (SV) were originally introduced by \textcite{RM-670-PR} as a method for fairly distributing both gains and costs among multiple actors cooperating within a coalition.  Table~\ref{tab:shapley_values__notation_definition} presents the notation used in this section, while Table~\ref{tab:shapley_values__axioms} defines the axioms through which the SV formalizes the notion of ``fairness.'' In the context of their application to Multimodal Large Language Models (MLLMs), the \textit{players} can be interpreted as input tokens or equivalent units, and the \textit{utility function} $U(S)$ can be understood as the similarity between the model’s output generated from the \textit{coalition} of tokens $S$ and an ideal response.

\begin{table}[htbp]
\centering
\caption{Notation Definition. For the scope of this work, \textit{model} refers to Multimodal Large Language Models (MLLMs).}

\begin{tabular}{p{2cm} p{10cm}}

\toprule

\textbf{Notation} & \textbf{Definition} \\

\midrule

$n$ & The number of players. \\
$m$ & Total number of samples (budget). \\
$U(\cdot)$ & Utility function. \\
$z_i$ & The $i$-th player, $i \in \{1, \dots, n\}$. \\
$S$ & A coalition (subset of players), $S \subset \{z_1, \dots, z_n\}$. \\
$N$ & The grand coalition (all players). \\
$R_S$ & A model response generated from input represented by coalition $S$, represented by a vector of output tokens. \\
$I_S$ & Model activations on the last layer, generated from input represented by coalition $S$, represented by a vector of real values. \\
$S_n$ & An $n$-coalition -- a coalition of $n$ players. \\
$S_i$ & An $i$-positive coalition -- a coalition containing player $z_i$. \\
$S_{-i}$ & An $i$-negative coalition -- a coalition not containing player $z_i$. \\
$S_l^i$ & An $(l, i)$-coalition -- a coalition of $l$ players where player $z_i$ is present. \\
$\Theta^j$ & The set of all $j$-coalitions. \\
$\Theta^{(l, i)}$ & The set of all $(l, i)$-coalitions. \\
$SV$ & The vector of Shapley values (SV) for all players. \\
$\overline{SV}$ & The vector of approximated SVs for all players. \\
$SV_i$ & SV of player $z_i$. \\
$\overline{SV_i}$ & Approximate SV of $z_i$. \\
$CC(S)$ & $U(S) - U(N \setminus S)$ -- complementary contribution of $S$. \\

\bottomrule

\end{tabular}
\label{tab:shapley_values__notation_definition}
\end{table}

\begin{table}[htbp]
\caption{Axioms of the Shapley values (SV) defining its notion of ''fairness''. It has been proven that it is the only function to meet those criteria.}
\centering
\renewcommand{\arraystretch}{1.1}
\setlength{\tabcolsep}{6pt}
\begin{tabular}{p{0.16\textwidth} p{0.72\textwidth}}

\toprule

\textbf{Axiom} & \textbf{Description} \\

\midrule

\textbf{Efficiency} &
All players’ credits sum to the grand coalition’s value:
$\sum_{i=1}^n SV_i = U(N) - U(\emptyset)$. \\[2pt]

\textbf{Symmetry} &
If two players $z_i$, $z_j$ ($i \neq j$) are interchangeable,
i.e.\ $\forall_{S:\, z_i, z_j \notin S}\; U(S \cup \{z_i\}) = U(S \cup \{z_j\})$,
then $SV_i = SV_j$. \\[2pt]

\textbf{Null player} &
If player $z_i$ contributes no value,
$\forall_{n = 1 \dots n}\; U(S_n^i) = U(S_n^i \setminus \{z_i\})$,
then $SV_i = 0$. \\[2pt]

\textbf{Linearity} &
Credits for linear combinations of games behave linearly:
$\forall_{a,b \in \mathbb{R},\, i,j = 1 \dots n}\;
SV_{a z_i} + SV_{b z_j} = SV_{a z_i + b z_j}$,
where $a z_i$ is the scaled contribution of player $z_i$. \\

\bottomrule

\end{tabular}
\label{tab:shapley_values__axioms}
\end{table}

The exact definition of SV goes as follows:

\[
SV_i = \sum_{S \subseteq N \setminus \{z_i\}} 
\frac{|S|!\,(n - |S| - 1)!}{n!} \,
\big(U(S \cup \{z_i\}) - U(S)\big)
\]

\noindent where \( U(S \cup \{z_i\}) - U(S) \) represents the \textit{marginal contribution} of player \( z_i \) to coalition \( S \), and 
\( \frac{|S|!\,(n - |S| - 1)!}{n!} \) is a \textit{weighting factor} -- the probability that player \( z_i \) joins the game after exactly \(|S|\) other players. Intuitively, this means that coalitions that are less likely to occur in random joining receive smaller weights, ensuring that the SV reflects an expected marginal contribution across all possible game orders.

\newpage
To compute the entire $SV$, we need to evaluate the utility function $U(\cdot)$ for all possible coalitions $S$. This requires performing $2^n$ potentially expensive evaluations, which becomes impractical for production use in the context of MLLMs, where the number of input tokens can be very large. State-of-the-art models operate over context windows containing thousands of tokens, and each model inference (and thus each evaluation of $U(\cdot)$) is both time- and energy-intensive.

While methods such as KernelSHAP \parencite{lundberg2017unifiedapproachinterpretingmodel} and TreeSHAP \parencite{yang2022fasttreeshapacceleratingshap}, popular in SHAP\footnote{\url{https://pypi.org/project/shap/}} package -- Python's most widely used SV-based Explainable Artificial Intelligence (XAI) tool, provide efficient approximations of the SV, they are inherently designed for models operating on fixed-length feature vectors, where the number and position of input features remain constant across all samples.

In contrast, MLLMs process variable-length sequences of tokens, meaning that both the number of input elements and their contextual dependencies can change dynamically between inputs. This variability violates the core assumption of these approximation methods that each feature represents a consistent and independent dimension of the input space.

As a result, KernelSHAP and TreeSHAP cannot be directly applied to MLLMs, since they lack mechanisms to handle variable-length token representations and context-dependent feature interactions that are fundamental to modern language models.

Another clear demonstration of the limitations of existing SV approximations in the context of Natural Language Processing (NLP) (and consequently MLLMs) applications was presented by \textcite{mosca-etal-2022-shap}. They outline active research directions in this field, with the most crucial for our purposes being \textbf{C1} (methods tailored to different input data), \textbf{C2} (methods for explaining different model architectures), and \textbf{C5} (methods for estimating SV more efficiently). As they emphasize in their work, no existing implementation satisfies all of these criteria.

Section~\ref{subsec:shapley_values__utility_func} outlines used utility function in SV calculation. Sections~\ref{subsec:shapley_values__monte_carlo} to~\ref{subsec:shapley_values__hierarchical} describe the approximation techniques to address the computational complexity of SV for MLLMs, with final benchmarks presented in Section~\ref{subsec:shapley_values__comparison}.

\subsection{Utility functions}
\label{subsec:shapley_values__utility_func}

Currently, most systems that compute semantic similarity between embedding vectors rely on the \textit{cosine similarity} (CS) measure (such as \cite{reimers}), for example, in context vector retrieval in Retrieval-Augmented Generation (RAG) architectures\footnote{\url{https://docs.pinecone.io/guides/get-started/overview}}. There are also successful applications of this measure in Large Language Models (LLMs) explainability using Shapley values (SV) \parencite{goldshmidt2024tokenshapinterpretinglargelanguage}.

CS is defined as the cosine of the angle between two non-zero vectors $x$ and $y$ in an inner product space:
\[
CS(x, y) = \frac{x \cdot y}{\|x\| \, \|y\|} = \frac{\sum_{i=1}^{n} x_i y_i}{\sqrt{\sum_{i=1}^{n} x_i^2} \, \sqrt{\sum_{i=1}^{n} y_i^2}}.
\]
It measures directional alignment rather than magnitude, making it scale-invariant and particularly suitable for comparing high-dimensional embedding vectors.

\textcite{zhou-etal-2022-problems} demonstrated that CS tends to underestimate the semantic similarity of frequent words compared to human judgments. This limitation has important implications for downstream Natural Language Processing (NLP) tasks, as cosine-based metrics are commonly used to measure embedding similarity. Their findings also indicate that word frequency in pre-training data shapes the geometry of embedding spaces -- low-frequency words form more compact clusters, while frequent words are distributed more broadly. Such frequency-dependent distortions can bias semantic representations and influence model behavior.

Similarly, \textcite{Steck_2024} showed, using linear matrix factorization models that allow analytical examination, that CS is highly sensitive to the choice of model and regularization technique, and in some cases can even become mathematically meaningless.

This work adopts CS in the calculation of the utility function, although existing literature provides notable criticism of its reliability. This choice is dictated by the absence of an evidently superior alternative, as well as by its widespread adoption in existing systems.

To limit the latent space of all possible text embeddings, from which vectors $R_S$ and $R_N$ originate, we apply the TF–IDF function (\cite{tfidf-wikipedia}, usage inspired by the work of \textcite{goldshmidt2024tokenshapinterpretinglargelanguage}) in order to consider only relevant words.

Term Frequency–Inverse Document Frequency (TF–IDF) is a numerical statistic intended to reflect how important a word is to a document within a collection or corpus. It is calculated as the product of two terms: the \textit{term frequency} (TF), which measures how frequently a term appears in a document, and the \textit{inverse document frequency} (IDF), which measures how rare the term is across all documents. Formally:

\[
\text{TF–IDF}(t, d, D) = TF(t, d) \times IDF(t, D)
\]

\noindent where \(
TF(t, d) = \frac{f_{t,d}}{\sum_{t' \in d} f_{t',d}}, \quad
IDF(t, D) = \log\frac{N}{|\{d \in D : t \in d\}|}
\), \( f_{t,d} \) is the number of times term \( t \) appears in document \( d \), \( N \) is the total number of documents, and \( |\{d \in D : t \in d\}| \) is the number of documents containing term \( t \). 

Words that occur frequently in a single document but rarely across the corpus receive higher TF–IDF scores, emphasizing their relative importance in that specific context.

\newpage
We use implementation of TF-IDF from scikit-learn\footnote{\url{https://scikit-learn.org/stable/modules/generated/sklearn.feature_extraction.text.TfidfVectorizer.html}} Python package.

The final package supports following utility functions:

\begin{description}
    \item[U1$(S) = CS(I_S, I_N)$] 

    This activation function relies on the model’s internal state, thereby slightly breaking the assumption of treating the model as a black box -- since it requires access to the final hidden layer activations. However, it allows CS to be computed over vectors that capture contextual information rather than just the final token sequence. This function will not be used in experiments, as it is not suitable for comparing different models.

    \item[U2$(S) = CS(e(R_S), e(R_N))$]

    where $e(d)$ represents the embedding of the detokenized output $R_S$. The embedding can be extracted using the same model’s initial layers (not applicable when comparing different models) or via a third-party embedding model.

    \item[U3$(S) = CS(h(R_S), h(R_N))$]
    
        where $h(d)$ represents the TF–IDF vector of document $d$:
    \[
        h(d) = [\textit{TF–IDF}(t_1, d, D), \dots, \textit{TF-IDF}(t_k, d, D)]
    \]
    with $D = \{R_S \mid S \subset N\}$ being the set of all model responses generated for every possible subset $S$ of the full input $N$.  
    The vocabulary $T = (t_1, \dots, t_k)$ is defined as $T = \bigcup D$, i.e., the set of all unique terms appearing in any response within $D$.  
    Thus, each $h(R_S)$ and $h(R_N)$ correspond to TF–IDF representations of model responses, allowing cosine similarity to quantify how closely the response to a subset $S$ resembles the response to the full input~$N$ at a cost of limiting its contextual knowledge.
\end{description}

\textbf{U1} can be useful in scenarios where comparing SV across different models is not required, as it captures the semantic context directly from the model’s internal representations. However, it suffers from a large latent space and complete dependence on the model’s internal knowledge, which can lead to unstable or unintuitive results when applied to weaker models. 

\textbf{U2} addresses this dependence by leveraging external embeddings, enabling more generalizable comparisons. Nonetheless, it still operates within a high-dimensional similarity space while lacking access to the internal semantic context of the model, which may reduce interpretability. 

Therefore, this work adopts \textbf{U3} as the primary utility function for experiments and sets it as the default configuration for the entire package. It provides a balance by enabling cross-model comparison while operating within a reduced and well-defined feature space for cosine similarity, thus improving the robustness and interpretability of results. Refer to Section~\ref{subsec:shapley_values__comparison} for details.

\subsection{Monte Carlo (MC) approximation}
\label{subsec:shapley_values__monte_carlo}

\textcite{goldshmidt2024tokenshapinterpretinglargelanguage} proposed a Monte Carlo (MC)–inspired approximation method for calculating Shapley values (SV) for Large Language Models (LLMs):

\[
\overline{SV_i} = p_i - m_i
\]

\noindent where 
\[
p_i = \frac{1}{k} \sum_{j=1}^{k} U(P_j), \quad P = (P_1, \dots, P_k)
\]
is the mean value of utility function over all sampled coalitions $P$ containing player $z_i$, and 
\[
n_i = \frac{1}{m - k} \sum_{j=1}^{m - k} U(N_j), \quad N = (N_1, \dots, N_{m - k})
\]
is the mean value of utility function over all sampled coalitions $N$ that do not contain player $z_i$. Coalitions are sampled using the MC method.

Following the authors, we constrain the sampling process to always include the \textit{first-order omission coalition}, i.e., coalitions $S_i^{n-1}$ in which player $z_i$ is excluded. The authors demonstrated that this approach achieves a mean approximation accuracy exceeding $85\%$ for any $m > n$. Selecting the appropriate number of samples thus represents a trade-off between computational efficiency and approximation accuracy.

For the purposes of this work, we note that increasing the conversation length n or the contextual complexity intuitively produces a more challenging environment for the model, characterized by a lower density of influential tokens. Natural language, for example, contains many low-information tokens such as stopwords (''a'', ''an'', ''the''), which contribute minimally to semantic meaning. 

Consequently, we hypothesize that the baseline approximation accuracy -- when relying solely on the first-order omission method -- will decline under such conditions. This, in turn, necessitates larger sampling budgets and motivates the search for more efficient approximation strategies.

We introduce two methods of sampling: 
\begin{itemize}
    \item \textit{Standard} approach, in which coalitions are sampled purely at random aside from the initial subset constructed according to the first-order omission methodology,  
    \item \textit{Limited} variant, which follows the original authors with a minor modification to prevent sampling the same coalition multiple times for computational efficiency. 
\end{itemize}

\newpage
Although model responses may vary due to stochasticity, we assume that utilizing a different coalition within the same sampling budget provides a greater informational benefit. Therefore, it is recommended to use the final package with the model configured for the most deterministic inference settings. A similar division into two implementations -- with and without uniqueness constraints -- was applied to all other approximation techniques.

Algorithm~\ref{alg:shapley_values__monte_carlo__alg} shows the final logic with a space complexity of $\mathcal{O}(m)$, a time complexity in respect to number of possibly expensive calls to the utility function $U$ of $O(m)$ (exactly $m$ calls).

\begin{algorithm}[htbp]
\caption{Monte Carlo (MC)-inspired algorithm for approximation of Shapley values (SV).}
\label{alg:shapley_values__monte_carlo__alg}
\begin{algorithmic}[1]

\State \textbf{Input:} Set of players $N = \{z_1, \dots, z_n\}$, utility function $U$, number of samples $m$, 
use\_first\_order\_omission, limited
\State \textbf{Output:} Approximate SV

\State Initialize empty vectors $R$, $S_{\text{sampled}}$ of size $m$, $idx \gets 0$

\If{use\_first\_order\_omission}
    \State $m \gets \max{(n, m)}$ \Comment{Omission requires budget of at least n}
    \ForAll{$z_i \in N$}
        \State $S \gets N \setminus \{z_i\}$ \Comment{First-order coalition for $z_i$}
        \State $R[idx] \gets U(S)$
        \State $S_{\text{sampled}}[idx] \gets S$
        \State $idx \gets idx + 1$
    \EndFor
\EndIf

\While{$idx < m$} \Comment{Spent remaining budget using random sampling}
    \State Sample a random coalition $S \subseteq N$
    \If{limited \textbf{and} $S \in S_{\text{sampled}}$}
        \State \textbf{continue} \Comment{Skip if already sampled}
    \EndIf
    \State $R[idx] \gets U(S)$
    \State $S_{\text{sampled}}[idx] \gets S$
    \State $idx \gets idx + 1$
\EndWhile

\ForAll{$z \in N$} \Comment{Calculate final approximations}
    \State $P \gets$ indexes of all coalitions from $S_{\text{sampled}}$ containing $z$
    \State $N \gets$ indexes of all coalitions from $S_{\text{sampled}}$ not containing $z$
    \State $p_z \gets \frac{1}{|P|} \sum_{idx \in P} R[idx]$ \Comment{Mean output for $i$-positive}
    \State $n_z \gets \frac{1}{|N|} \sum_{idx \in N} R[idx]$ \Comment{Mean output for $i$-negative}
    \State $\overline{SV}_z \gets p_z - n_z$
\EndFor

\State \Return $\overline{SV} = (\overline{SV}_1, \dots, \overline{SV}_n)$

\end{algorithmic}
\end{algorithm}

\newpage
\subsection{Complementary Contributions (CC) Monte Carlo approximation}
\label{subsec:shapley_values__complementary}

\textcite{shapleyapproximations} has proven that the original Shapley values (SV) definition can be rewritten as 

\begin{equation}
\label{equation:shapley_values__complementary__definition}
SV_i = \frac{1}{n} \sum_{S \subseteq N \setminus \{z_i\}} \frac{CC(S \cup \{z_i\})}{{n - 1 \choose |S|}}
\end{equation}

\noindent The core advantage of this formulation is that each \textit{complementary contribution} $CC(S)$ can be used in the calculation of $SV_i$ for every player $z_i \in S$, whereas the original \textit{marginal contribution} $U(S \cup \{z_i\}) - U(S)$ can be used only in the computation of $SV_i$. Moreover, as $CC(S) = -CC(N \setminus S)$, $CC(S)$ can contribute to the computation of every entry in the $SV$ vector.

Equation~\ref{equation:shapley_values__complementary__definition} can be further rewritten as

\[
SV_i = \frac{1}{n} \sum_{j=1}^n SV_{(i, j)},
\]

\noindent where $SV_{(i, j)} = \sum_{S \in \Theta^{(i, j)}} \frac{CC(S)}{{n - 1 \choose j - 1}}$ is the expected complementary contribution of $(i, j)$-coalitions. Such stratification optimizes the computation of the final approximation of $SV$, where, given a matrix $M \in \mathbb{N}^{n \times n}$ of sample sizes aligned to corresponding (i, j)-coalitions, we denote

\[
\overline{SV_i} = \frac{1}{n} \sum_{j = 1}^n \overline{SV}_{(i, j)}, \quad 
\overline{SV}_{(i, j)} = \frac{1}{M_{(i, j)}} C_{(i, j)},
\]

\noindent where $C_{(i, j)} = \sum_{k = 1}^{M_{(i, j)}} CC(S_k)$,  $S_1, \dots, S_{M_{(i, j)}}$ are randomly sampled from $\Theta^{(i, j)}$ and \\ $\sum_{i, j = 1}^n M_{(i, j)} = m$.

This follows from the fact that $SV_{(i, j)}$ is the expected value of a random variable $CC_{(i, j)}$ with a uniform distribution over $\{CC(S) \mid S \in \Theta^{(i, j)} \}$. The original authors proved this formula to be an unbiased estimator of $SV_i$.

Algorithm~\ref{alg:shapley_values__complementary__alg} presents the final formulation, featuring a space complexity of $O(n^2 + m)$ and a time complexity of $O(m)$ with respect to the utility function $U$ (specifically, $2m$ calls). Compared to Algorithm~\ref{alg:shapley_values__monte_carlo__alg}, this approach theoretically requires a smaller m to achieve the same approximation accuracy as the method described in Section~\ref{subsec:shapley_values__monte_carlo} (see Section~\ref{subsec:shapley_values__comparison} for further discussion).

For simplicity, we refer to this algorithm as the CC-based algorithm for SV computation, even though it is also Monte Carlo (MC)-inspired; in contrast, MC-based algorithms in the existing literature typically rely on marginal contributions. For consistency with algorithm proposed in Section~\ref{alg:shapley_values__monte_carlo__alg}, sampling is also limited to unique coalitions only.

\begin{algorithm}[b!]
\caption{Complementary Contributions (CC)-based algorithm for approximation of Shapley values (SV).}
\label{alg:shapley_values__complementary__alg}
\begin{algorithmic}[1]

\State \textbf{Input:} Set of players $N = \{z_1, \dots, z_n\}$ with defined order, utility function $U$, number of samples $m$, 
use\_first\_order\_omission, limited
\State \textbf{Output:} Approximate SV

\State Initialize empty vectors $R$, $S_{\text{sampled}}$ of size $m$, matrices $C$, $M$ of size $n \times n$ full of zeros, $idx \gets 0$

\While{$idx < m$} 
    \State Sample a random coalition $S \subseteq N$
    \If{limited \textbf{and} $S \in S_{\text{sampled}}$}
        \State \textbf{continue} \Comment{Skip if already sampled}
    \EndIf
    \State $R[idx] \gets U(S) - U(N \setminus S)$
    \State $S_{\text{sampled}}[idx] \gets S$
    \State $idx \gets idx + 1$
\EndWhile

\For{$k = 1$ to $m$} \Comment{Update matrices M and C}
    \State $i \gets |S[idx]|$ \Comment{Number of players in that coalition}
    \For{$j = 1$ to $i$}
        \State $z_k \gets S_{\text{sampled}}[idx][j]$ \Comment{$k$ - index of that player in $N$}
        \State $M_{(k, i)} \gets M_{(k, i)} + 1$
        \State $C_{(k, i)} \gets C_{(k, i)} + R[idx]$
    \EndFor
    \For{$j = i + 1$ to $n$}
        \State $z_k \gets (N \setminus S_{\text{sampled}}[idx])[j]$
        \State $M_{(k, n - i)} \gets M_{(k, n - i)} + 1$
        \State $C_{(k, n - i)} \gets C_{(k, n - i)} - R[idx]$
    \EndFor
\EndFor

\ForAll{$i \in N$} \Comment{Calculate final approximations}
    \State $\overline{SV}_i \gets \frac{1}{n} \sum_{j = 1}^{n}
        \begin{cases}
            \frac{C_{(i, j)}}{M_{(i, j)}}, & \text{if } M_{(i, j)} \neq 0 \\
            0, & \text{otherwise}
        \end{cases}$
\EndFor

\State \Return $\overline{SV} = (\overline{SV}_1, \dots, \overline{SV}_n)$

\end{algorithmic}
\end{algorithm}

\subsection{Neyman approximation for Complementary Contributions (CC)}
\label{subsec:shapley_values__neyman}

The sum of variances of the Shapley values (SV) approximated according to Algorithm~\ref{alg:shapley_values__complementary__alg} is \parencite{shapleyapproximations}:

\begin{equation}
\label{equation:shapley_values__neyman__variance}
\sum_{i=1}^{n} \operatorname{Var}\!\left[\overline{SV}_{i}\right]
= \frac{1}{n^{2}} \sum_{i=1}^{n} \sum_{j=1}^{n}
    \frac{\sigma_{i,j}^{2}}{M_{(i,j)}}.
\end{equation}

\noindent where $\sigma_{i,j}^2$ is the variance of $CC_{(i,j)}$ (refer to Section~\ref{subsec:shapley_values__complementary}). 

To minimize this sum, and therefore make our approximation less variable, we need to sample data taking into account the reference impact of the matrix $M$ on the final variance.

\newpage
Difficulty arising from complex relationships between entries of $M$ can be solved by replacing $M_{i,j}$ with its expected value. Let $m_j$ be the size of the sample set drawn from $\Theta^j$. After drawing single samples $S \in \Theta^j$ we can estimate $CC_N(S)$ that will be used in $\overline{SV}_{i,j}$ for $z_i \in S$ and in $\overline{SV}_{i,n-j}$ for $z_i \in N \setminus S$. Therefore $\mathbb{P}(X \in CC_{(i,j)}) = \mathbb{P}(z_i \in S) = \frac{j}{n}$ where $X$ is a random variable reflecting the drawn sample. Since we draw them independently, we have $\mathbb{E}[M_{(i,j)}] = \frac{1}{n} m_{\max(j,n-j)}$ after drawing $m_{\max(j,n-j)}$ samples. Substituting this expectation into Equation~\ref{equation:shapley_values__neyman__variance}, we have:

\[
\sum_{i=1}^{n} \operatorname{Var}\!\left[\overline{SV}_{i}\right]
= \frac{1}{n^{2}} \sum_{i=1}^{n} \sum_{j=1}^{n} 
    \frac{\sigma_{i,j}^{2}}{m_{\max(j,\, n-j)}}
= \underbrace{
    \frac{1}{n} \sum_{j=\lceil n/2 \rceil}^{n-1} 
    \frac{
        \sum_{i=1}^{n} 
        \left( 
            \frac{\sigma_{i,j}^2}{j} + 
            \frac{\sigma_{i,n-j}^2}{n-j} 
        \right)
    }{m_j}
}_{h(m_{\lceil n/2 \rceil}, \dots, m_n)}.
\]

\noindent The last equality is due to re-indexing. The minimization problem can be stated as follows:

\[
    \min_{\substack{m_{\lceil n/2 \rceil}, \dots, m_n \\ \sum_{i = \lceil n/2 \rceil}^{n} m_i = 1}}
    h(m_{\lceil n/2 \rceil}, \dots, m_n)
\]

\noindent As \textcite{shapleyapproximations} has shown, using Lagrange multipliers the optimal solution is
\begin{equation}
\label{equation:shapley_values__neyman__m}
    m_j = \frac{m \sqrt{\sum_{i=1}^{n} \left( \frac{\sigma_{i,j}^2}{j} + \frac{\sigma_{i,n-j}^2}{n-j} \right)}}{\sum_{k=\lceil n/2 \rceil}^{n} \sqrt{\sum_{i=1}^{n} \left( \frac{\sigma_{i,k}^2}{k} + \frac{\sigma_{i,n-k}^2}{n-k} \right)}}.
\end{equation}
For estimation of $\sigma_{i,j}^2$ we use Bessel's correction:

\begin{equation}
\label{equation:shapley_values__neyman__c}
    \hat{\sigma^2}_{i, j} = \frac{\overline{C}}{M - 1} - \frac{C^2}{M(M - 1)}
\end{equation}

\noindent where $\overline{C}_{(i, j)} = \sum_{k = 1}^{M_{(i, j)}} CC(S_k)$.

Algorithm~\ref{alg:shapley_values__neyman__alg} presents the final formulation, with a space complexity of $O(n^2)$ and a time complexity of $O(m)$ with respect to the utility function $U$ (specifically, $2m$ calls). The key distinctions from Algorithm~\ref{alg:shapley_values__complementary__alg} include sampling only coalitions of size $\ge \lceil n/2 \rceil$ (besides initial phase) and dynamically determining the sampling budget for each j-coalition, subject to a global limit of m. This time it cannot be limited to sampling only unique coalitions, as it might be impossible depending on the value of $m_{\text{init}}$. One might spot that it is prone to making many redundant calls to $U$ for small coalitions. It does not necessarily pose performance issues, as in our case they can be easily cached.

For subsequent use, we define $m_{\text{init}} = \max(2,\ \left\lfloor \frac{m}{2n^2} \right\rfloor)$, following the choice made by the original authors.

\begin{algorithm}[H]
\caption{Neyman-based algorithm for approximation of Shapley values (SV) using Complementary Contributions (CC).}
\label{alg:shapley_values__neyman__alg}
\begin{algorithmic}[1]

\State \textbf{Input:} Set of players $N = \{z_1, \dots, z_n\}$ with defined order, initial sample count $m_{\text{init}} > 1$, utility function $U$, number of samples $m$
\State \textbf{Output:} Approximate SV

\State Initialize vector $\hat{M}$ of size $n$, matrices $C$, $\overline{C}$, $M$ of size $n \times (n + 1)$ full of zeros, $c \gets -1$

\Comment{Initial sampling}
\While{not $\forall_{(i, j)} \quad M \ge m_{\text{init}}$}
    \For{$i = 1$ to $n$}
        \For{$j = 0$ to $n$}
            \If{$M_{(i,j)} < m_{\text{init}}$} \Comment{Until matrix $M$ has all entries $\ge m_{\text{init}}$}
                \State Sample a random coalition $S \in \Theta^j$
                \State $u \gets U(S) - U(N \setminus S)$
                \If{$j \ne 0$}
                    \ForAll{$z_k \in S$} \Comment{$k$ - index of that player in $N$, from 1}
                        \State $C_{(k,j)} \gets C_{(k,j)} + u$
                        \State $\overline{C}_{(k,j)} \gets \overline{C}_{(k,j)} + u^2$
                        \State $M_{(k,j)} \gets M_{(k,j)} + 1$
                    \EndFor
                \Else
                    \State $\dots$ \Comment{Do the same but for entire 0th column of all matrices}
                \EndIf

                \If{$j \ne n$}
                    \ForAll{$z_i \in N \setminus S$}
                        \State $C_{(k, n - j)} \gets C_{(k, n - j)} - u$
                        \State $\overline{C}_{(k,n-j)} \gets \overline{C}_{(k,n-j)} + u^2$
                        \State $M_{(k, n - j)} \gets M_{(k, n - j)} + 1$
                    \EndFor
                \Else
                    \State $\dots$ \Comment{Do the same but for entire 0th column of all matrices}
                \EndIf
                \State $m \gets m -1$
            \EndIf
        \EndFor
    \EndFor
\EndWhile

\State Compute estimated $\hat{\sigma^2}$ according to Formula~\ref{equation:shapley_values__neyman__c} \Comment{Element wise multiplication}

\For{$j = \lceil n/2 \rceil$ to $n$}
    \State $\hat{M}_j \gets \text{approximation according to Formula~\ref{equation:shapley_values__neyman__m}}$
\EndFor

\Comment{Sample for remaining estimated budget}
\For{$j = \lceil n/2 \rceil$ to $n$} \Comment{Just ''large'' coalitions}
    \For{$k = 1$ to $\hat{M}_j$} \Comment{Sample remaining number of times}
        \State $\dots$ \Comment{Same logic as earlier lines $8$ - $19$}
    \EndFor
\EndFor

\ForAll{$i \in N$} \Comment{Calculate final approximations}
    \State $\overline{SV}_i \gets \frac{1}{n} \sum_{j = 1}^{n} \frac{C_{(i, j)}}{M_{(i, j)}}$
\EndFor

\State \Return $\overline{SV} = (\overline{SV}_1, \dots, \overline{SV}_n)$

\end{algorithmic}
\end{algorithm}

\subsection{Hierarchical Approach}
\label{subsec:shapley_values__hierarchical}

\begin{figure}[b!]
    \centering
    \includegraphics[width=0.85\linewidth]{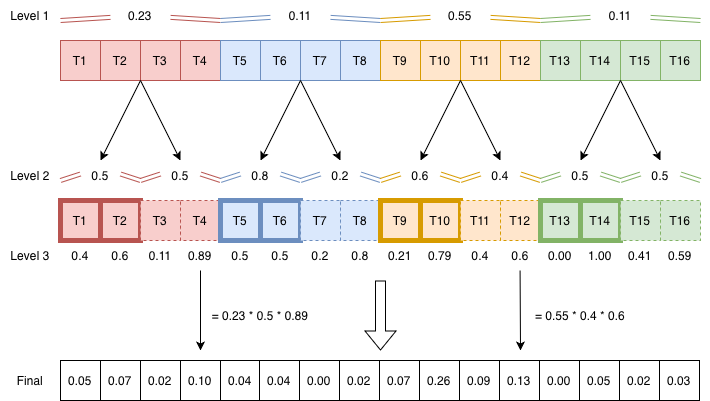}
    \caption{Example decomposition for $n = 16$ for Large Language Model's (LLM's) input tokens. We first group adjacent tokens by 4 (\textit{Level 1} - 4 groups in total indicated by different colors), then each of the subgroups is divided into 2 (\textit{Level 2} - 8 groups in total, each in different color and frame border). Final Shapley values (SV) are in the lowest frame. All floating-point numbers represent SV calculated for corresponding group / token.}
    \label{fig:shapley_values__hierarchical__example}
\end{figure}

Inputs to Multimodal Large Language Models (MLLMs) can be very large, particularly in multimodal conversations. To ensure our methods remain computationally feasible (as required by \textbf{NFR-1} and \textbf{NFR-2} -- refer to Appendix~\ref{app:requirements}), we propose grouping adjacent players to reduce the overall complexity. To preserve approximation accuracy, each group is processed recursively -- the Shapley value (SV) of a group is computed by further decomposing it into smaller subgroups. The final SV for an individual player is then obtained as the product of the SVs along its corresponding path in this hierarchical decomposition tree. Figure~\ref{fig:shapley_values__hierarchical__example} showcases exemplary decomposition process, whereas Algorithm~\ref{alg:shapley_values__hierarchical__alg} formalizes the logic. The total number of calls to the explaining algorithm is at most 
$2\left\lceil \frac{n}{k_{\max}} \right\rceil - 1$ 
(and at least $\left\lceil \frac{n}{k_{\max}} \right\rceil$), 
giving time complexity with respect to the utility function $U$
$O\!\left(c_F\,\frac{n}{k_{\max}}\right)$, 
with recursion depth $\left\lceil \log_{k_{\max}} n \right\rceil$. Here, $n$ is the number of players/tokens; $k_{\max}$ is the maximum final (leaf) group size; $U$ is the utility function; $c_F$ is the number of $U$-evaluations performed inside a single call to the explainer $F$ (e.g., $c_F=m$ for Monte Carlo (MC, refer to Section~\ref{subsec:shapley_values__monte_carlo}) and $c_F=2m$ for Complementary Contributions (CC, refer to Section~\ref{subsec:shapley_values__complementary}) or Neyman (refer to Section~\ref{subsec:shapley_values__neyman}). Additional space complexity comes only from recursion stack size.

\begin{algorithm}[H]
\caption{Hierarchical approximation of Shapley values (SV).}
\label{alg:shapley_values__hierarchical__alg}
\begin{algorithmic}[1]

\State \textbf{Input:} Set of players $N = \{z_1, \dots, z_n\}$ with defined order, SV approximation algorithm $F$, maximum final group size $k_{\text{max}}$
\State \textbf{Output:} Approximate SV $\overline{SV}_i$ for all $z_i \in N$

\Function{ComputeSV}{$G$}
    \State $k \gets |G|$
    \State $l \gets \lceil log_{k_{max}}(k) \rceil$ \Comment{Number of subgroups}
    \If{$l \le 1$} \Comment{Final group: compute SV directly}
        \State \Return $F(G)$ \Comment{SV for final group}
    \EndIf
    \State Divide $G$ into $l$ contiguous subgroups $G_1, \dots, G_l$ of same size \Comment{$G_l$ might be smaller}
    \State Initialize vector $\overline{SV}$ of size k, full of zeros
    \For{$j = 1$ to $l$}
        \State $\overline{SV}_{G_j} \gets F(G_j) * $ \Call{ComputeSV}{$G_j$} \Comment{Recursive computation, $\overline{SV}_{G_j}$ is a subvector of $\overline{SV}$ indexed by tokens from $G_j$}
    \EndFor
    \State \Return $\overline{SV}$
\EndFunction

\State $\overline{SV} \gets$ \Call{ComputeSV}{$N$}
\State \Return $\overline{SV}$

\end{algorithmic}
\end{algorithm}

\section{Spectrogram-Guided Phonetic Alignment (SGPA)}
\label{sec:experiments__sgpa}

A practical difficulty when applying Shapley value (SV) analysis to Multimodal Large Language Models (MLLMs) is the mismatch in granularity between text and audio inputs. Text is naturally discretized into words or tokens, whereas audio is represented as a continuous waveform that audio encoders convert into dense frame- or token-level sequences. In typical MLLM pipelines, the resulting audio representation contains several times more units than the corresponding transcript, which is quantified for our use case precisely in Appendix~\ref{app:compute_analysis}.

Running SV analysis at the native audio-unit level is impractical for two reasons. First, the number of coalitions grows as $2^n$ with the number of players $n$, so increasing the number of input units quickly makes exact computation infeasible and substantially increases the cost of standard Monte Carlo estimators. Second, segmenting audio purely by time or by raw token counts can place boundaries inside phonetic transitions (e.g., co-articulation). Such cuts may create discontinuities during masking, introducing audible artifacts and out-of-distribution perturbations that can degrade attribution fidelity.

To address these constraints, we introduce Spectrogram-Guided Phonetic Alignment (SGPA), a preprocessing procedure that maps a word-level transcript to time-aligned audio segments. After alignment, each word (or word span) defines one SV player, which keeps the game size manageable while preserving semantic interpretability.

\subsection{Methodological framework}

SGPA combines model-based alignment with local signal-based boundary refinement. It consists of three stages: (i) transcript decomposition, (ii) alignment based on a data-driven model, and (iii) spectrogram-guided boundary refinement followed by (iv) word-level aggregation.

\paragraph{Stage 1: Transcript Decomposition}
We first decompose the transcript into words and characters and record the character-to-word membership. This bookkeeping allows us to estimate character-level time boundaries and then merge them into word-level segments in a final aggregation step.

\paragraph{Stage 2: Initial Alignment via CTC}
To obtain an initial estimate of where transcript characters occur in the audio stream, we use Connectionist Temporal Classification (CTC) \parencite{graves2006connectionist}. Unlike dynamic time warping (DTW) \parencite{dwt}, CTC does not require a reference audio signal and can align an unsegmented input sequence (audio) to a target sequence (text).

We use \texttt{Wav2Vec2-XLSR-53} \parencite{baevski2020wav2vec} to support the multilingual setting of our experiments (English, Spanish, French). Given an audio waveform $A$, the model produces an emission matrix $\mathbf{E} \in \mathbb{R}^{T \times V}$, where $T$ is the number of time frames and $V$ is the vocabulary size.

We then apply Viterbi decoding over $\mathbf{E}$ to extract the most likely character path $\pi^*$:
\begin{equation}
    \pi^* = \arg\max_{\pi} P(\pi \mid \mathbf{E}) .
\end{equation}

From $\pi^*$ we obtain approximate start and end frame indices for each character, which are converted to timestamps using the model's effective frame stride. These boundaries are suitable as coarse anchors but are not guaranteed to coincide with acoustically stable cut points.

\paragraph{Stage 3: Spectrogram-Guided Boundary Refinement}
We treat the CTC-derived timestamps as candidate boundaries and refine them using local spectral cues to reduce the likelihood of cutting through transient phonetic events. Speech commonly alternates between high-energy steady regions (e.g., vowels) and lower-energy regions or rapid transitions (e.g., closures, silences).

For each candidate boundary $t_{\mathrm{est}}$, we define a search window $W = [t_{\mathrm{est}} - \delta,\ t_{\mathrm{est}} + \delta]$. Within $W$ we compute:
\begin{enumerate}
    \item \textbf{Short-time energy $E[n]$:} Root Mean Square (RMS) energy per frame as a measure of amplitude.
    \item \textbf{Spectral flux $SF[n]$:} The change in the magnitude spectrum between consecutive frames, computed from the Short Time Fourier Transform (STFT) \parencite{mcfee2015librosa}.
\end{enumerate}

We select the refined boundary as the point in $W$ with minimal weighted cost:
\begin{equation}
    t_{\mathrm{final}} = \arg\min_{t \in W} \left(\alpha\, E[t] + \beta\, SF[t]\right) .
\end{equation}
Intuitively, this favors cut points near low-energy and spectrally stable regions (often short pauses or inter-word transitions), which reduces the risk of introducing artifacts during the masking operations required for Shapley values (SV) estimation. We empirically determined that $\alpha = 0.8$ and $\beta = 0.2$ yielded the best results for our use case based on English samples. Further investigation is required into the cross-lingual impact of these hyperparameters.

\paragraph{Stage 4: Word-Level Aggregation}
Finally, we aggregate refined character-level boundaries into word-level segments using the transcript decomposition from Stage 1. This granularity aligns with the primary unit of human interpretation (words) and reduces the player count relative to raw audio units, making SV estimation (including the Neyman estimator used in our experiments) computationally feasible.

\begin{figure}[htbp]
    \centering
    \includegraphics[width=\textwidth]{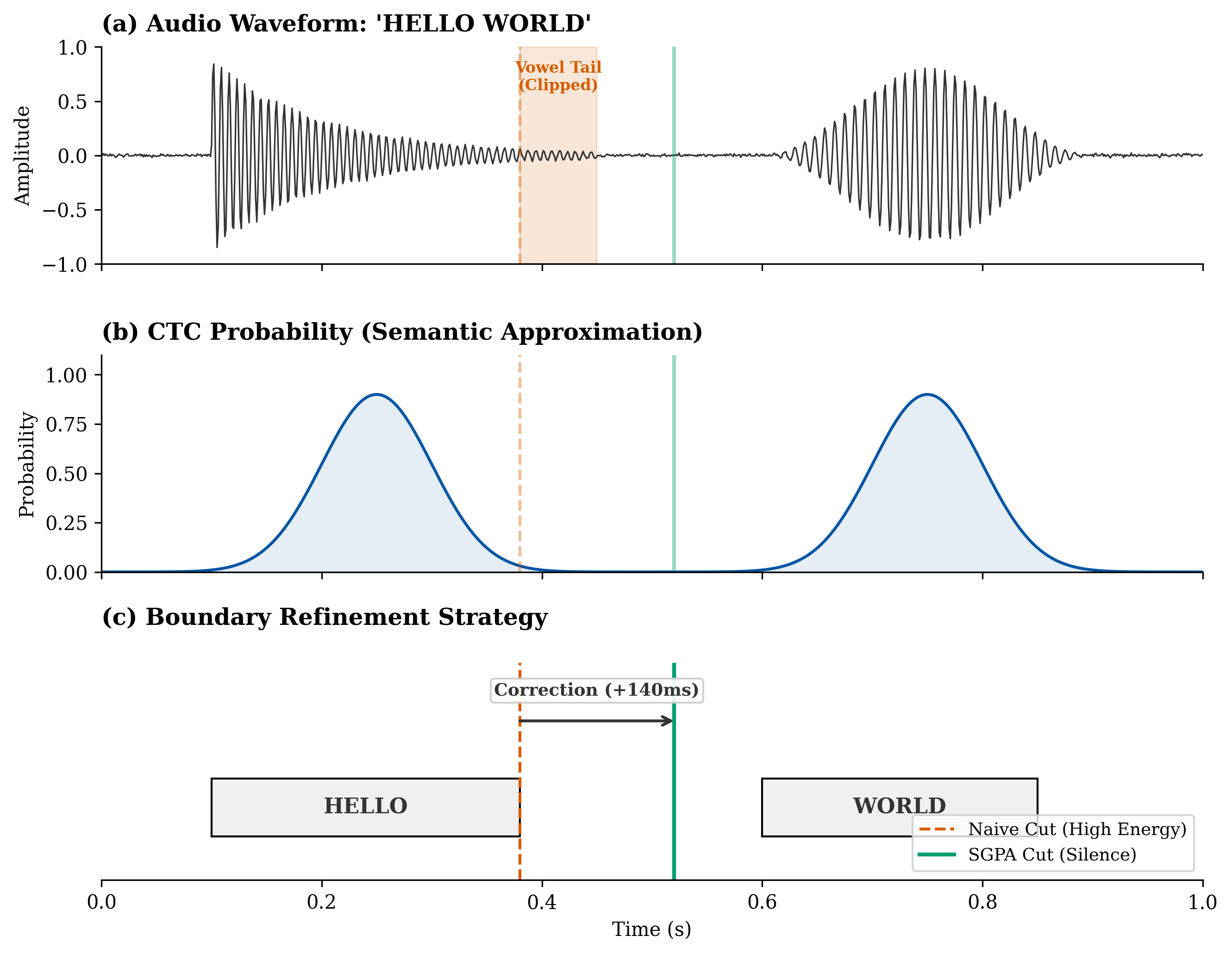}
    \caption{Schematic representation of the Spectrogram-Guided Phonetic Alignment (SGPA) framework. The Waveform (a) shows the raw audio. The Connectionist Temporal Classification (CTC) Probability (b) indicates the model activation for a specific character. The Alignment View (c) shows the refinement: the CTC boundary (dashed red) is approximate, while the refined boundary (solid green) shifts to a nearby low-energy region between words.}
    \label{fig:sgpa_schematic}
\end{figure}

Figure~\ref{fig:sgpa_schematic} illustrates the SGPA pipeline, where CTC boundaries are shifted to acoustically stable regions to avoid high-energy vowels and prevent audible artifacts.

\chapter{Solution}
\label{sec:solution}

The previous chapters established \emph{what} must be built, \emph{why} certain design decisions are unavoidable (Chapter~\ref{sec:requirements}), as well as \emph{how} we approached the problem from the theoretical side (Chapter~\ref{sec:theory}). This chapter describes how we \emph{built} it -- presenting the three integrated components that transform the theoretical framework into a functioning research instrument.

The platform architecture consists of:

\begin{enumerate}[nosep]
    \item \textbf{Core attribution engine} (Section~\ref{sec:package}): A Python package (\texttt{mllm-shap}) implementing the Shapley value (SV) computation methods from Chapter~\ref{sec:shapley_values}, with connectors for multimodal model integration and utilities for filtering, normalization, and result export.
    
    \item \textbf{Experimental infrastructure} (Section~\ref{sec:experiments__infrastructure}): A reproducible execution pipeline with configuration management, checkpointing, artifact versioning, and optional observability -- enabling systematic studies at scale under hardware constraints.
    
    \item \textbf{Interactive interface} (Section~\ref{sec:gui}): A web-based graphical user interface (GUI) that wraps the attribution engine, providing real-time visualization, session management, and cost estimation for exploratory research.
\end{enumerate}

These components are \textbf{mutually dependent}. The package provides the computational primitives; the infrastructure orchestrates large-scale batch execution with fault tolerance; the GUI serves as both a debugging tool during development and a demonstration of near-real-time feasibility enabled by SGPA segmentation (Chapter~\ref{sec:experiments__sgpa}). Section~\ref{sec:tests__components} validates the package's modular design through exhaustive configuration testing, while Section~\ref{subsec:shapley_values__comparison} benchmarks the approximation methods to justify default parameter choices used throughout the platform. Finally, Section~\ref{subsec:experiments__sgpa__diagnostics} details our investigation of SGPA method and its limitations.

This chapter follows a bottom-up structure: we begin with the core package (the foundation), describe the infrastructure that enables reproducible experiments, present the GUI as a user-facing layer, and conclude with validation of the platform's key components.

\section{Package}
\label{sec:package}

The project includes a Python package that operationalizes all Shapley-value (SV) algorithms introduced in Section~\ref{sec:shapley_values} within a unified, reproducible pipeline for Multimodal Large Language Models (MLLMs). The package is designed to support the thesis experiments end-to-end: it standardizes model interaction through connectors, enforces consistent feature-unit tracking (modality/role), implements mask generation and caching, evaluates utility functions over model responses, and exposes a modular family of SV estimators (precise and approximate) tailored to multimodal conversational settings.

A key engineering objective was to ensure that experimental comparisons remain interpretable: estimator ablations should differ only in the approximation method, while masking policies, connector behavior, and utility evaluation remain fixed. The package also prioritizes reproducibility and traceability (deterministic inference defaults, artifact logging, and Confidence Interval (CI)-gated testing). The detailed details of the full Application Programming Interface (API) (classes, module paths, and configuration parameters) are provided in the Appendix~\ref{app:package_reference}.

\subsection{Design goals and scope}
\label{sec:package__goals}

The package is scoped around three goals aligned with the thesis contributions:

\begin{enumerate}
    \item \textbf{Unified multimodal interaction.} Interaction with MLLMs is encapsulated behind connector interfaces, enabling consistent handling of text and audio inputs/outputs and preserving modality metadata required for explainability.
    \item \textbf{Estimator comparability under controlled budgets.} Multiple SV estimators are implemented under a shared pipeline, enabling fair comparisons between precise SV computation and approximation methods (Monte Carlo, complementary contributions, Neyman allocation, and hierarchical variants - refer to Section~\ref{sec:shapley_values}).
    \item \textbf{Engineering robustness.} The implementation emphasizes deterministic defaults, input validation, caching, Confidence Interval-gated tests, and user-facing documentation to make experiments reproducible and extensible.
\end{enumerate}

The package is not positioned as a general-purpose interpretability framework. Instead, it targets the multimodal conversational setting of this thesis, where feature units must retain modality and role information, and where computational feasibility requires explicit budget handling.

\subsection{Architecture and data flow}
\label{sec:package__arch}

\begin{figure}[htbp]
    \centering
    \includegraphics[width=0.7\linewidth]{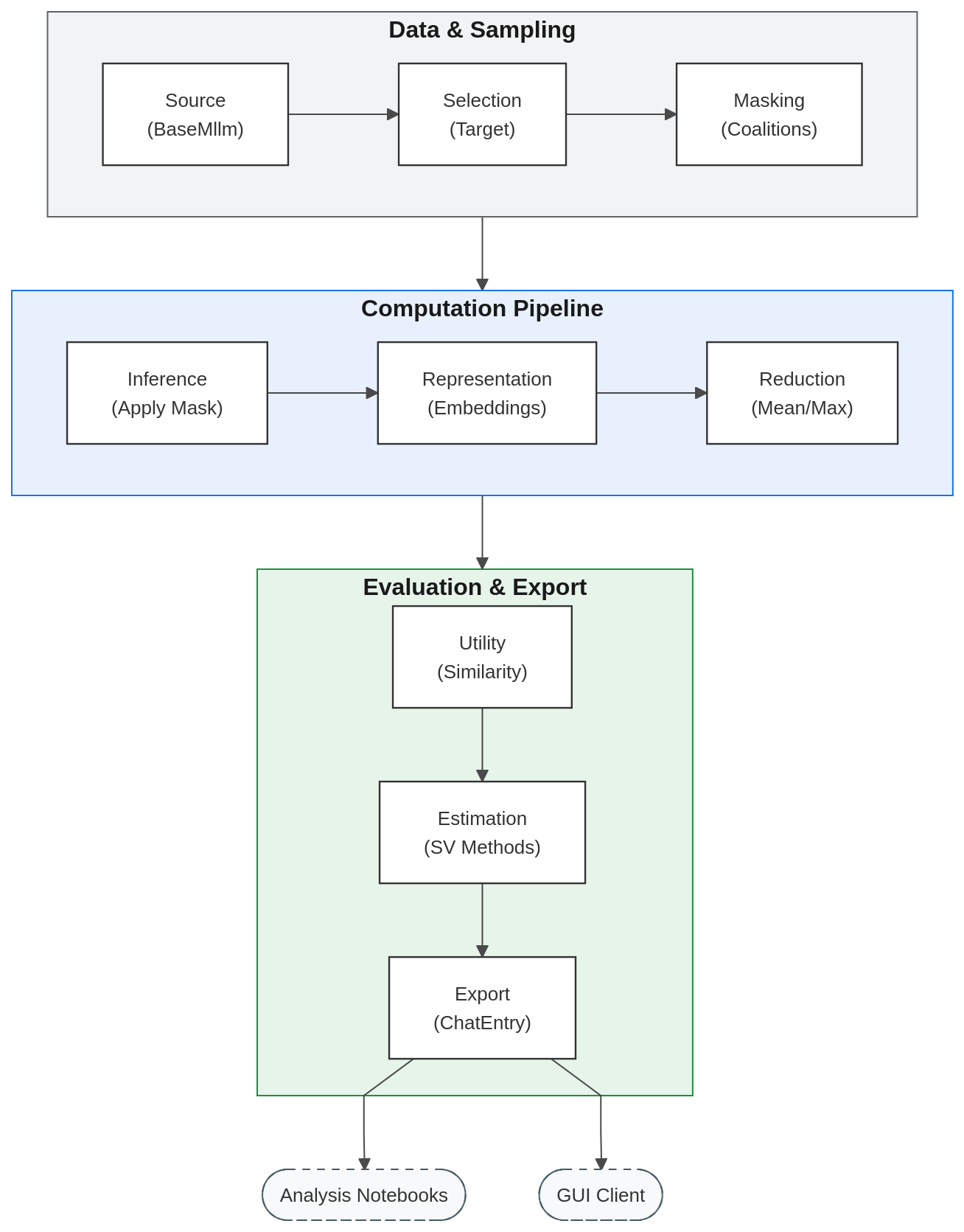}
    \caption{High-level package pipeline used in all experiments: chat construction and feature-unit tracking, mask-based coalition generation, connector-based inference, utility evaluation, and Shapley values (SV) estimation with export for analysis / Graphical User Interface (GUI).}
    \label{fig:package__arch}
\end{figure}

Figure~\ref{fig:package__arch} summarizes the execution flow used throughout the experiments. Given a source chat and a target model response, the explainer constructs feature units (tokens aligned with modality and role), generates and applies coalition masks, replays masked chats through a model connector, evaluates the utility function on the resulting responses, and aggregates marginal contributions into Shapley values. The final output remains a chat-like structure with SV values aligned to each token, which supports both analysis and Graphical User Interface (GUI) visualization.

\subsection{Extensibility and configuration strategy}
\label{sec:package__extensibility}

The implementation is intentionally modular to support controlled ablations:

\begin{itemize}
    \item \textbf{Connector boundary (black-box compliance).} Model integration is isolated in connector classes (Appendix~\ref{app:package_reference}, Section~\ref{sec:package__func}), responsible only for translating between package-level chat objects and model-specific Input / Output (I/O) formats. This design supports the black-box setting (\textbf{RR-1}, Appendix~\ref{app:requirements}).
    \item \textbf{Swappable utility evaluation.} Utility functions are implemented via interchangeable similarity measures and embedding strategies (Appendix~\ref{app:package_reference}, Section~\ref{sec:package__func}). This allows the same estimator to be evaluated under different representational assumptions (e.g., TF--IDF vs embedding-based cosine similarity - refer to Section~\ref{subsec:shapley_values__utility_func}).
    \item \textbf{Estimator plug-ins.} All SV explainers implement a shared callable interface so that the experimental pipeline can switch between precise SV and approximation algorithms without modifying connector or masking code (Appendix~\ref{app:package_reference}, Section~\ref{sec:package__func}).
\end{itemize}

This separation is crucial in the experimental chapters: when comparing estimator behavior, the only intended variable is the estimator itself rather than the underlying model integration or utility implementation.

\subsection{Engineering quality: reproducibility, testing, and documentation}
\label{sec:package__quality}

\textbf{Reproducibility.} By default, supported models generate using deterministic parameters (\texttt{temperature} $=0$, \texttt{top-}$k=1$), enforced via a default configuration object (Appendix~\ref{app:package_reference}, Section~\ref{sec:package__func}). This reduces variance originating from sampling and isolates estimator-induced variability in experiments.

\begin{figure}[t!]
    \centering
    \includegraphics[width=1\linewidth]{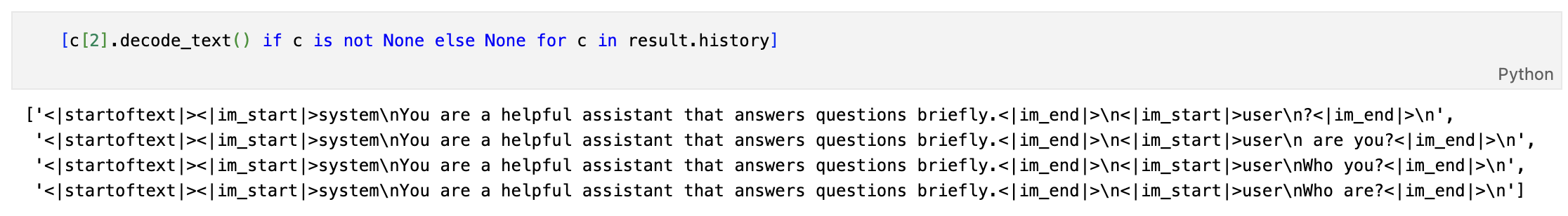}
    \caption{Manual test on correctness of masks generation with token filtering criteria and role-based explanation enabled.}
    \label{fig:package__docs__masks_tests}
\end{figure}

\textbf{Testing and CI gates.} The repository includes (i) manual notebooks for exploratory validation and experiment replication, (ii) an automated unit test suite integrated into Continuous Integration / Deployment (CI/CD), and (iii) quality gates (formatting, linting, static analysis, and security checks). These practices are used to ensure that experimental results are not coupled to ad-hoc local modifications. Figure~\ref{fig:package__docs__masks_tests} illustrates a manual correctness test of mask generation under filtering and role-based explanation.

\begin{figure}[t!]
    \centering
    \includegraphics[width=0.75\linewidth]{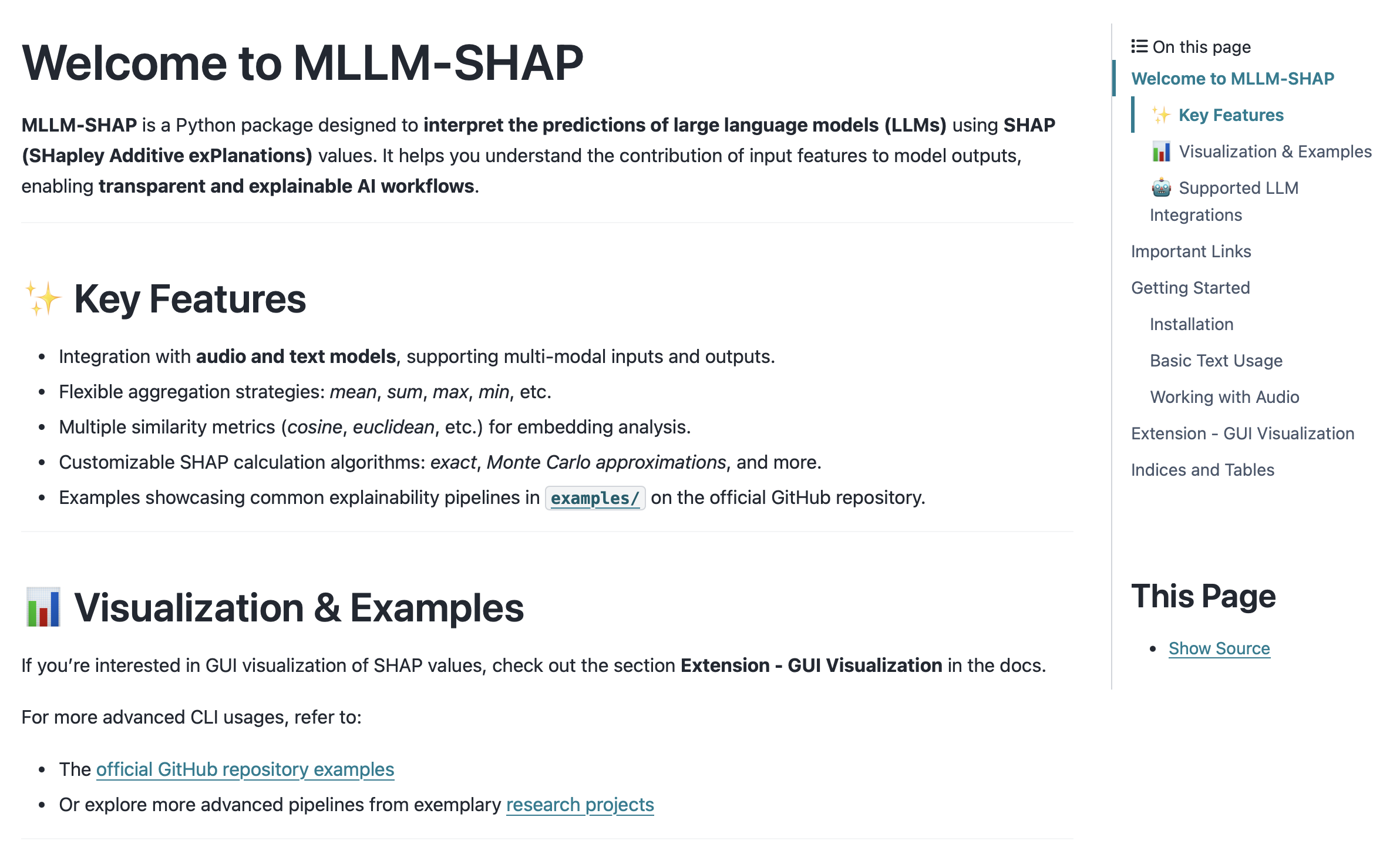}
    \caption{Beginning of the documentation home page. It includes a brief package description and highlights key features. The page structure is shown on the right.}
    \label{fig:package__doc_s_home}
\end{figure}

\textbf{User-facing documentation.} The package exposes a stable public API and provides Sphinx-generated documentation with examples and a GUI guide (Appendix~\ref{app:package_reference}, Section~\ref{sec:package__docs}). The documentation home page is shown in Figure~\ref{fig:package__doc_s_home}.

\subsection{Requirement traceability and package outputs}
\label{sec:package__traceability}

A detailed requirements mapping is provided in Appendix~\ref{app:package_reference} (Section~\ref{sec:package__align}). In summary, the connector abstraction supports multimodal I/O (\textbf{FR-A2}) and the black-box integration pattern (\textbf{RR-1}), while modality/role tracking preserves feature-unit semantics (\textbf{RR-2}). Multiple approximation algorithms address estimator requirements for long sequences (\textbf{RR-3}), and deterministic defaults and CI-gated validation support reproducibility and reliability (\textbf{NFR-3}, \textbf{NFR-5}).

The output format is intentionally analysis-friendly: each explained conversation can be exported as a token-level table retaining modality/role metadata and SV values. An example output is shown in Table~\ref{tab:package__align__example}.

\begin{table}[t!]
    \centering
    \caption{Package output for multi-turn conversation, showing turns $1$ and $3$. System messages (Role $2$) are not set to be explained in that scope -- they are not explained with \texttt{NaN}. SV stands for Shapley value.}
    \label{tab:package__align__example}
    \begin{tabular}{ccccc}
        \toprule
        \textbf{ID} & \textbf{Token} & \textbf{SV} & \textbf{Role} & \textbf{Turn} \\
        \midrule
        0 & \texttt{<|im\_start|>} & NaN & 2 & 1 \\
        1 & \texttt{user} & NaN & 2 & 1 \\
        2 & \text{ } & NaN & 2 & 1 \\
        3 & \texttt{Who} & 0.231312 & 0 & 1 \\
        4 & \texttt{are} & 0.000000 & 0 & 1 \\
        5 & \texttt{you} & 0.229981 & 0 & 1 \\
        6 & \texttt{?} & NaN & 0 & 1 \\
        7 & \texttt{<|im\_end|>} & NaN & 2 & 1 \\
        8 & \text{ } & NaN & 2 & 1 \\
        \midrule
        9 & \texttt{<|im\_start|>} & NaN & 2 & 3 \\
        10 & \texttt{user} & NaN & 2 & 3 \\
        11 & \text{ } & NaN & 2 & 3 \\
        12 & \texttt{Can} & 0.154570 & 0 & 3 \\
        13 & \texttt{you} & 0.167566 & 0 & 3 \\
        14 & \texttt{repeat} & 0.216570 & 0 & 3 \\
        15 & \texttt{?} & NaN & 0 & 3 \\
        16 & \texttt{<|im\_end|>} & NaN & 2 & 3 \\
        17 & \text{ } & NaN & 2 & 3 \\
        \bottomrule
    \end{tabular}
\end{table}

For completeness, full API documentation, module paths, and implementation details are provided in Appendix~\ref{app:package_reference} (Sections~\ref{sec:package__func}--\ref{sec:package__tests}).

\section{Experimental Infrastructure}
\label{sec:experiments__infrastructure}

\begin{figure}[htbp]
    \centering
    \includegraphics[width=\textwidth]{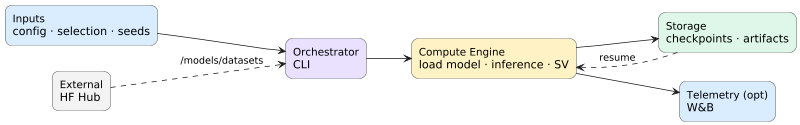}

    \caption[End-to-end experimental infrastructure: configuration expansion, execution engine, checkpoints, artifact store, and optional W\&B (Weights\&Biases) telemetry.]{End-to-end experimental infrastructure: configuration expansion, execution engine, checkpoints, artifact store, and optional W\&B (Weights\&Biases)\footnotemark \vspace{0.2cm}  telemetry.}
    
    \label{fig:infra__architecture}
\end{figure}

This section describes the software and runtime infrastructure that enables reliable, repeatable, and observable experimentation across all studies reported in Chapter~\ref{sec:experiments_results}. The design goals were: (i) reproducibility under constrained compute, (ii) observability of cost and attribution metrics, (iii) fault tolerance via resumable runs, and (iv) portability to a standard Linux workstation (see Appendix~\ref{app:requirements_inventory}). A high-level view of the end-to-end stack is shown in Figure~\ref{fig:infra__architecture}.

\footnotetext{\url{https://wandb.ai/site/}}

\newpage
\addtocontents{toc}{\protect\setcounter{tocdepth}{1}}

\subsection{Configuration and Reproducibility}
\label{subsec:infra__config}

\addtocontents{toc}{\protect\setcounter{tocdepth}{2}}

All experiments are declared as structured configurations and expanded into concrete runs:
\begin{itemize}
    \item \textbf{Experiment declaration:} An \texttt{ExperimentSet} defines dataset, selection, generation, and Shapley values (SV) computation settings. Each user-declared \texttt{ExplainerVariant} is materialized into one or more concrete runs (e.g., \texttt{exact}, or \texttt{Monte Carlo} with specific \texttt{num\_samples}/\texttt{fraction}); see the orchestration path in Figure~\ref{fig:infra__architecture}.
    \item \textbf{Seeding:} Deterministic data selection uses a fixed \texttt{shuffle\_seed}; model calls are run with stable inference settings to minimize stochasticity (Appendix~\ref{app:requirements}).
    \item \textbf{Spec snapshot:} Every run writes an immutable \texttt{spec.json} to disk (Section~\ref{subsec:infra__artifacts}), capturing the entire configuration, device, and software versions for replayability (layout in Figure~\ref{fig:infra__artifacts}).
\end{itemize}

In addition, a system prompt is chosen, as it is needed by methods such as Neyman stratification (described in Section~\ref{subsec:shapley_values__neyman}). For the purposes of experiment execution, we use a standard system prompt, used for example by OpenAI\footnote{As part of few-shot prompt, \url{https://platform.openai.com/docs/guides/prompt-engineering}}: \textit{``You are a helpful assistant''}. This prompt is fed as system-level instruction to the model in every run, regardless of the method used, to ensure consistency of output across methods and runs.

\addtocontents{toc}{\protect\setcounter{tocdepth}{1}}

\subsection{Data Access and Version Pinning}
\label{subsec:infra__data}

\addtocontents{toc}{\protect\setcounter{tocdepth}{2}}

Datasets are fetched from the Hugging Face Hub with explicit revision pinning. Downloads are blocked unless the revision matches a $40$-hex commit Secure Hash Algorithm (SHA); users may override via \texttt{ALLOW\_UNPINNED\_HF\_DOWNLOAD=1} for exploratory work. This guarantees that results in Chapter~\ref{sec:experiments_results} can be reproduced byte-for-byte against the same dataset snapshot. Sentence-level alignment and language checks are performed during preprocessing (see Section~\ref{subsec:experiments__datasets}).

\addtocontents{toc}{\protect\setcounter{tocdepth}{1}}

\subsection{Execution Engine}
\label{subsec:infra__engine}

\addtocontents{toc}{\protect\setcounter{tocdepth}{2}}

The run loop is orchestrated by the runner:
\begin{enumerate}
    \item \textbf{Variant build:} Models are loaded on the chosen device; an \texttt{Explainer} is constructed with \textit{Exact}, \textit{Monte Carlo}, or \textit{Neyman}-style estimators. For Monte Carlo grids, the base explainer is re-instantiated per \texttt{num\_samples} or \texttt{fraction}.
    \item \textbf{Row iteration:} A deterministic iterator yields rows according to selection bounds and (optional) shuffling. Audio modality is resolved (\textit{male}/\textit{female}), and text prompts are extracted from the selected column.
    \item \textbf{Attribution call:} The explainer is invoked with generation settings; per-sample wall time is recorded. Conversation objects are serialized; Shapley values (SV) are sanitized (\texttt{NaN}/\texttt{Inf} handling) and summarized per modality.
    \item \textbf{Checkpointing:} After each sample, a durable checkpoint (\texttt{checkpoint.json}) is updated. On resume, completed indices are merged with those inferred from disk to avoid re-computation.
\end{enumerate}

\begin{figure}[h!]
    \centering
    \includegraphics[width=0.35\textwidth]{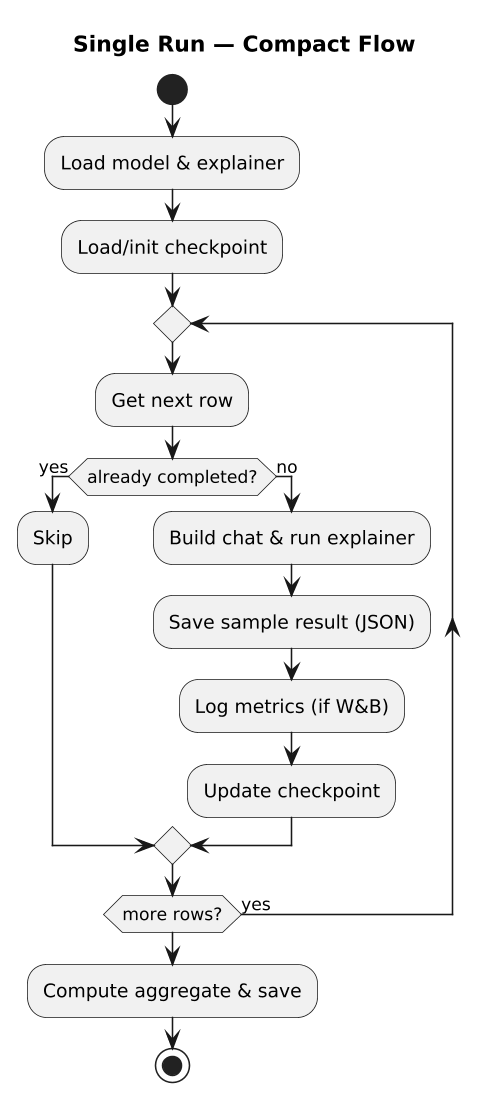}
    \caption{Execution Directed Acyclic Graph (DAG) for a single run: build explainer $\rightarrow$ iterate rows $\rightarrow$ attribute $\rightarrow$ persist result $\rightarrow$ update checkpoint. Failure-safe edges resume from the last persisted state.}
    \label{fig:infra__dag}
\end{figure}

The execution flow for a single run is summarized in Figure~\ref{fig:infra__dag}.

\newpage
\addtocontents{toc}{\protect\setcounter{tocdepth}{1}}

\subsection{Observability and Logging}
\label{subsec:infra__observability}

\addtocontents{toc}{\protect\setcounter{tocdepth}{2}}

\begin{figure}[h!]
    \centering
    \includegraphics[width=\textwidth]{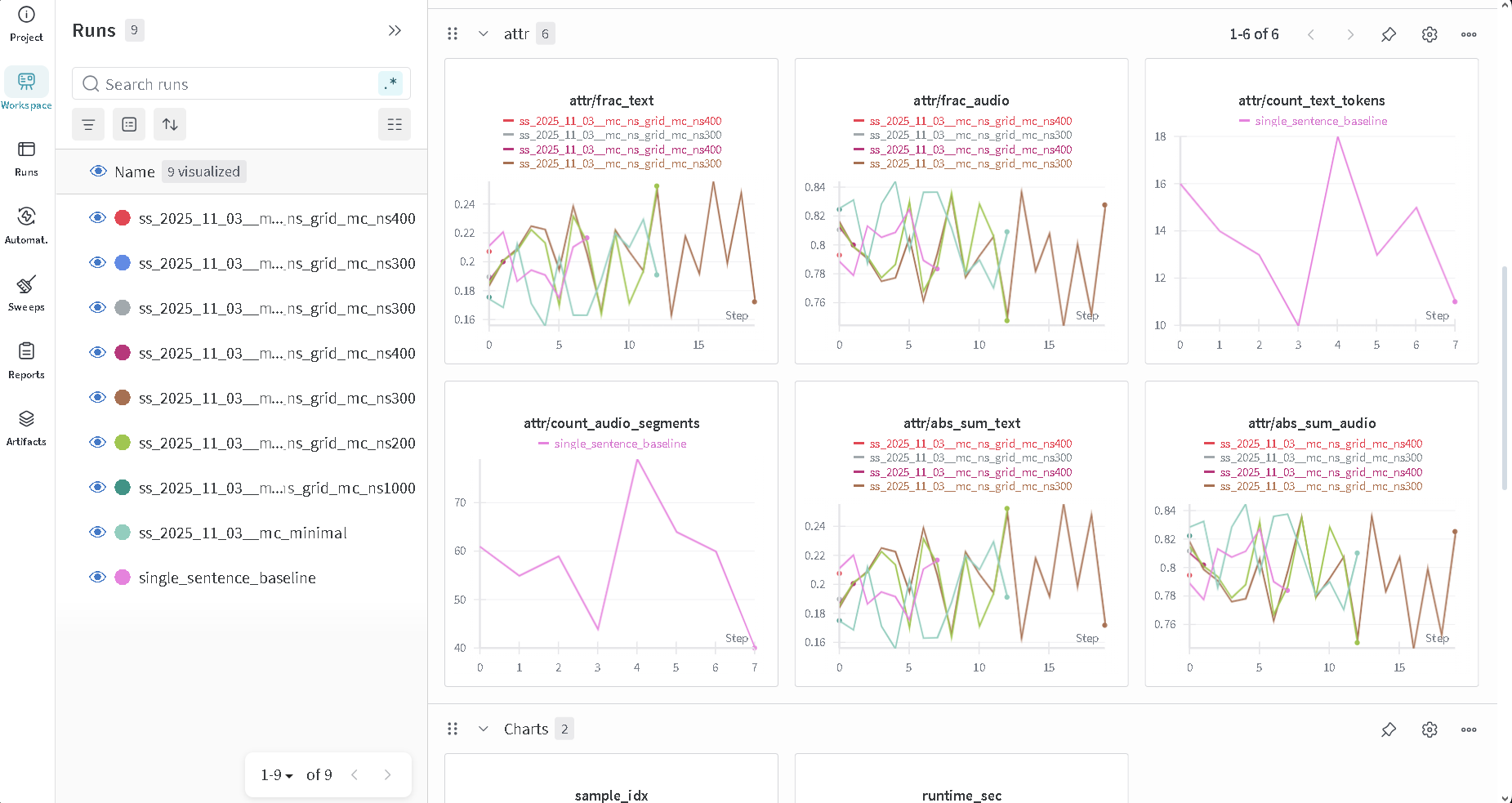}
    \caption{Optional observability: example W\&B (Weights\&Biases) dashboard panels (progress counters, runtime distributions, modality fractions) when telemetry is enabled.}
    \label{fig:infra__wandb}
\end{figure}

Two complementary channels record outcomes:
\begin{itemize}
    \item \textbf{On-disk artifacts} (authoritative): per-sample JSON results, a run-level \texttt{spec.json}, and an aggregated \texttt{summary/aggregate\_metrics.json} with averages over runtime and modality fractions.
    \item \textbf{Optional W\&B (Weights\&Biases) telemetry}: a thin wrapper performs dynamic imports (no hard dependency), controlled by a configuration switch and environment mode. When enabled, the system logs progress counters, runtime distributions, and aggregated attribution fractions; the final summary is uploaded as an artifact.
\end{itemize}
Illustrative dashboards are shown in Figure~\ref{fig:infra__wandb}.

\addtocontents{toc}{\protect\setcounter{tocdepth}{1}}

\subsection{Artifact Layout and Resume Semantics}
\label{subsec:infra__artifacts}

\addtocontents{toc}{\protect\setcounter{tocdepth}{2}}

\begin{figure}[h!]
    \centering
    \includegraphics[width=0.83\textwidth]{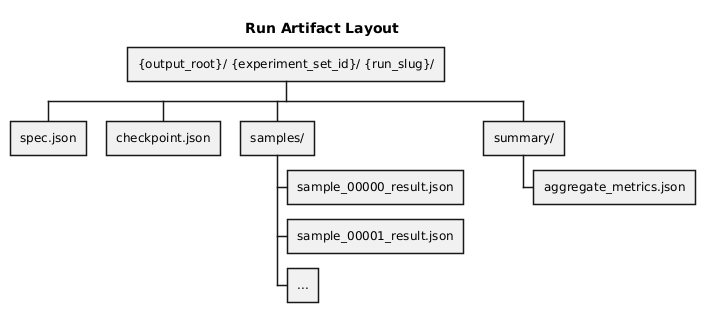}
    \caption{On-disk artifact layout per run showing \texttt{spec.json}, \texttt{checkpoint.json}, per-sample results, and aggregated metrics.}
    \label{fig:infra__artifacts}
\end{figure}

Every concrete run writes to:
\[
\texttt{\{output\_root\}/\{experiment\_set\_id\}/\{run\_slug\}/}
\]
with the following structure:
\begin{itemize}
    \item \texttt{spec.json} $\rightarrow$ frozen configuration and environment snapshot.
    \item \texttt{checkpoint.json} $\rightarrow$ \texttt{completed\_indices}, \texttt{next\_index}, timestamps.
    \item \texttt{samples/} $\rightarrow$ one \texttt{sample\_\{idx\}\_result.json} per processed row (conversation + per-sample metrics).
    \item \texttt{summary/aggregate\_metrics.json} $\rightarrow$ run-level aggregates (count, total/avg runtime, average modality fractions).
\end{itemize}
If a run is interrupted, resuming merges on-disk \texttt{samples/} with the checkpoint to skip already-finished indices. The directory layout is depicted in Figure~\ref{fig:infra__artifacts}.

\section{Graphical User Interface}
\label{sec:gui}

To bridge the gap between the Shapley value (SV) framework and practical model diagnostics, a companion Graphical User Interface (GUI) was developed. The GUI enables an interactive workflow aligned with the experimental procedures used in Chapter~\ref{sec:experiments}: the user configures a model and explanation method, interacts with the model through text or audio, and requests explanations that are computed by the backend and visualized immediately in the interface. In this way, SV computation becomes an exploratory loop rather than a purely code-driven process.

The application is intended primarily as a research companion: it supports near real-time analysis (subject to the computational cost of SV estimation), persistent session tracking, and export of explanation artifacts for reporting and downstream analysis. Input modalities are limited to text and audio. The full implementation is available as open-source on GitHub\footnote{\url{https://github.com/mvishiu11/shap-mllm-explainer}}.

\subsection{System architecture}
\label{sec:system_arch}

\begin{figure}[t!]
    \centering
    \includegraphics[width=0.6\textwidth]{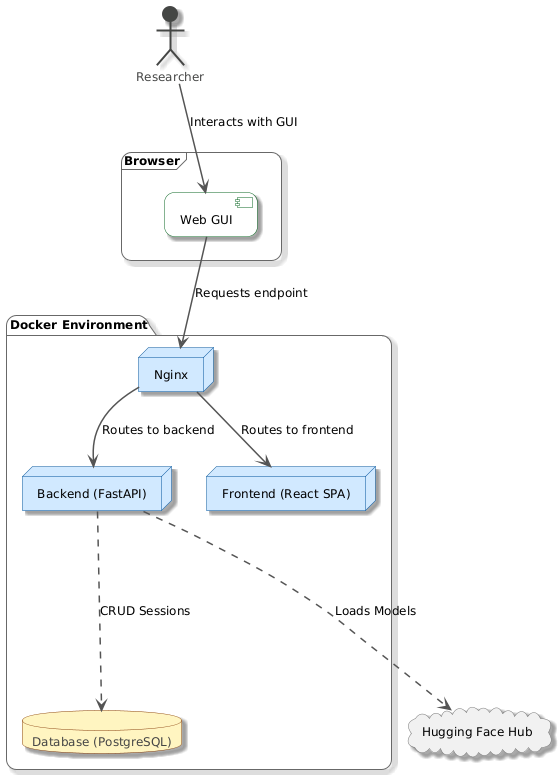}
    \caption{High-level system architecture of the interactive Shapley values (SV) explanation tool, showing the separation between the React frontend and the FastAPI backend, which handles model loading, inference, and SV computation via a Representational State Transfer Application Programming Interface (REST API).}
    \label{fig:gui_architecture}
\end{figure}

The GUI follows a client--server design that separates an interactive frontend from a computationally intensive backend. This separation ensures that the browser remains responsive while long-running inference and SV estimation are performed asynchronously on the server. The high-level architecture is shown in Figure~\ref{fig:gui_architecture}, where the backend exposes a Representational State Transfer Application Programming Interface (REST API) responsible for model loading, inference, and explanation computation.

From an engineering perspective, this design supports three objectives relevant to the thesis: (i) responsiveness during expensive computations, (ii) reproducible and traceable experiment sessions through persistent storage, and (iii) a stable interface for exporting results used throughout this work.

\subsection{Features and user workflow}
\label{sec:gui_workflow}

\begin{figure}[t!]
\centering
    \includegraphics[width=0.8\textwidth]{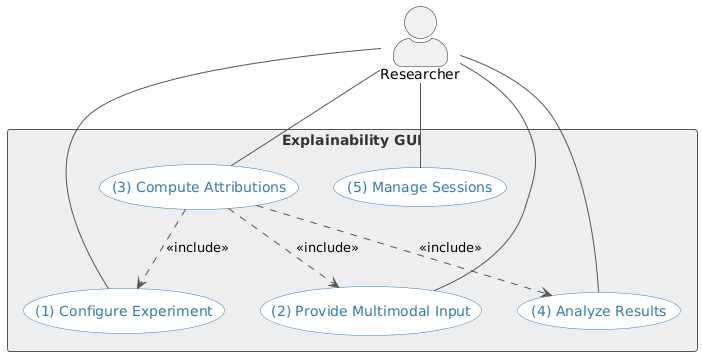}
    \caption{Unified Modeling Language (UML) Use Case diagram illustrating the primary interactions a researcher can perform with the Graphical User Interface (GUI), from configuration and interaction to visualization and session management.}
    \label{fig:gui_use_case}
\end{figure}

\begin{figure}[t!]
    \centering
    \includegraphics[width=\textwidth]{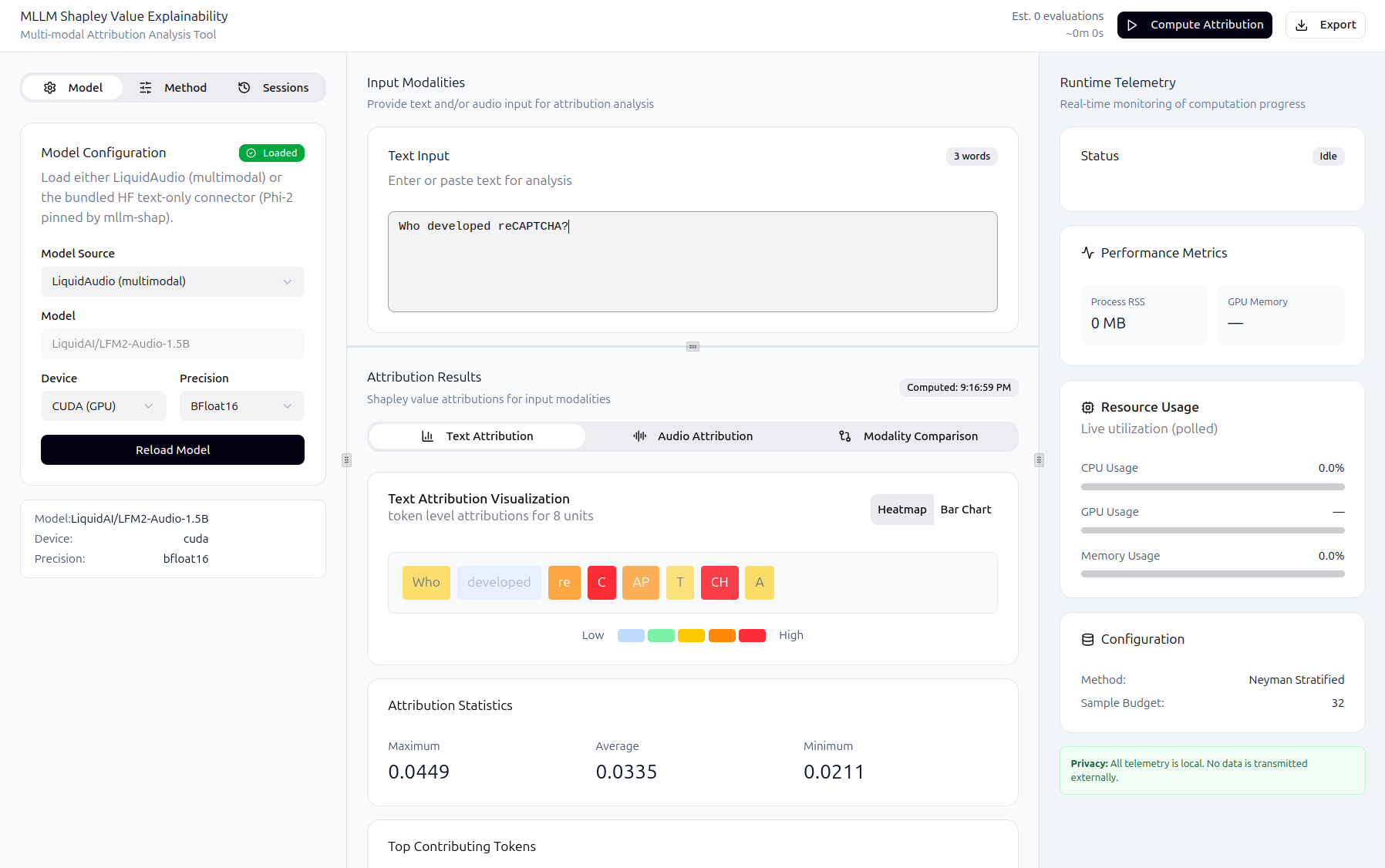}
    \caption{Screenshot of the main application interface, demonstrating the layout: (A) Configuration panels for model and method, (B) Chat interface for multimodal (text/audio) interaction, and (C) Visualization panel displaying text attribution heatmaps and audio attribution plots.}
    \label{fig:gui_screenshot}
\end{figure}

The GUI is designed around a single workflow that guides the user from configuration to explanation and inspection (Figure~\ref{fig:gui_use_case}). The interface is divided into three panels (Figure~\ref{fig:gui_screenshot}): configuration of the model and method, chat interaction (text/audio), and visualization of attributions.

\begin{figure}[t!]
    \centering
    \includegraphics[width=\textwidth]{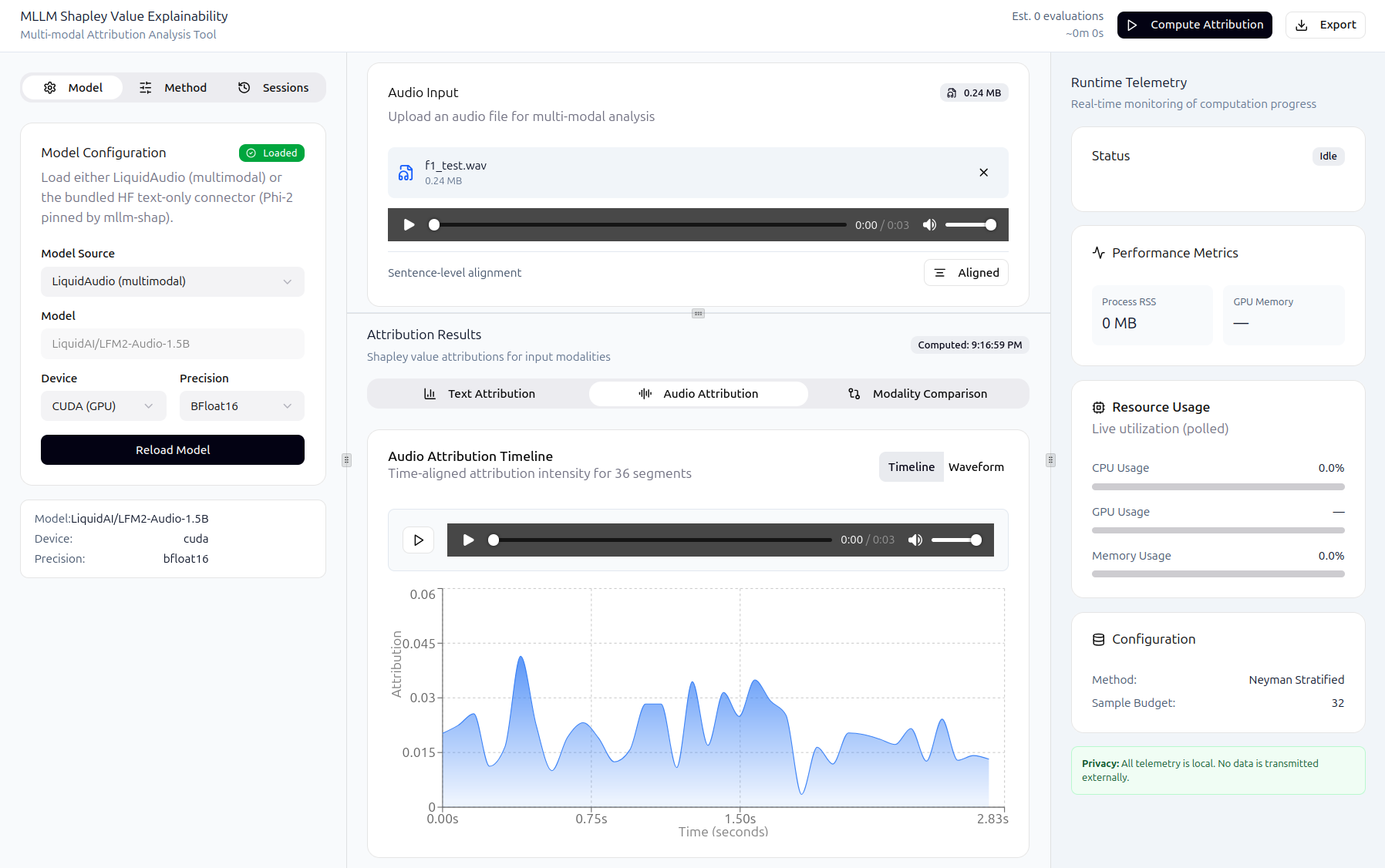}
    \caption{Screenshot of the main application interface demonstrating the layout in case of audio explanation with a visualization panel displaying audio attribution across a waveform.}
    \label{fig:gui_screenshot_audio}
\end{figure}

For audio explanations, the workflow remains identical: the user uploads an audio file and receives attributions aligned with time, visualized alongside the waveform (Figure~\ref{fig:gui_screenshot_audio}).

Two implementation details are critical for usability in the SV setting.
First, the GUI provides an \emph{upfront cost estimate} before launching an explanation, warning the user about the number of required model evaluations and expected latency. Second, the application persists full sessions (configuration, conversation history, and explanations), enabling systematic comparison of runs and ensuring that results can be revisited and exported for analysis.

A complete mapping of GUI features to the functional and non-functional requirements (\textbf{FR-A1}--\textbf{FR-A8}, \textbf{NFRs}) is provided in Appendix~\ref{app:gui_reference}.

\section{Verification of Configurable Package Components}
\label{sec:tests__components}

To ensure the robustness of the package architecture and validate adherence to \textbf{NFR-1} (Usability) and \textbf{NFR-5} (Reliability), we performed an exhaustive integration test covering all implemented configuration options within the explainer. The precise wording of those requirements is given in Appendix~\ref{app:requirements} (Section~\ref{subsec:requirements__nfr}). 

\newpage
This verification ensures that the modular design -- separating embedding extraction, dimensionality reduction, and similarity measurement -- functions correctly across all supported permutations.

We evaluated the package against a complete combinatorial grid of the following parameters:
\begin{itemize}
    \item \textbf{Embeddings:} \textit{Static} (model embeddings), \textit{Contextual} (hidden states), and \textit{External} (via \texttt{intfloat/e5-small-v2});
    \item \textbf{Reducers:} \textit{Mean}, \textit{Max}, \textit{Min}, \textit{Sum}, and \textit{First} token reduction;
    \item \textbf{Similarity Measure:} \textit{Cosine}, \textit{Euclidean}, and \textit{TF-IDF} + Cosine.
\end{itemize}

\begin{figure}[htbp]
    \centering
    \includegraphics[width=1.0\linewidth]{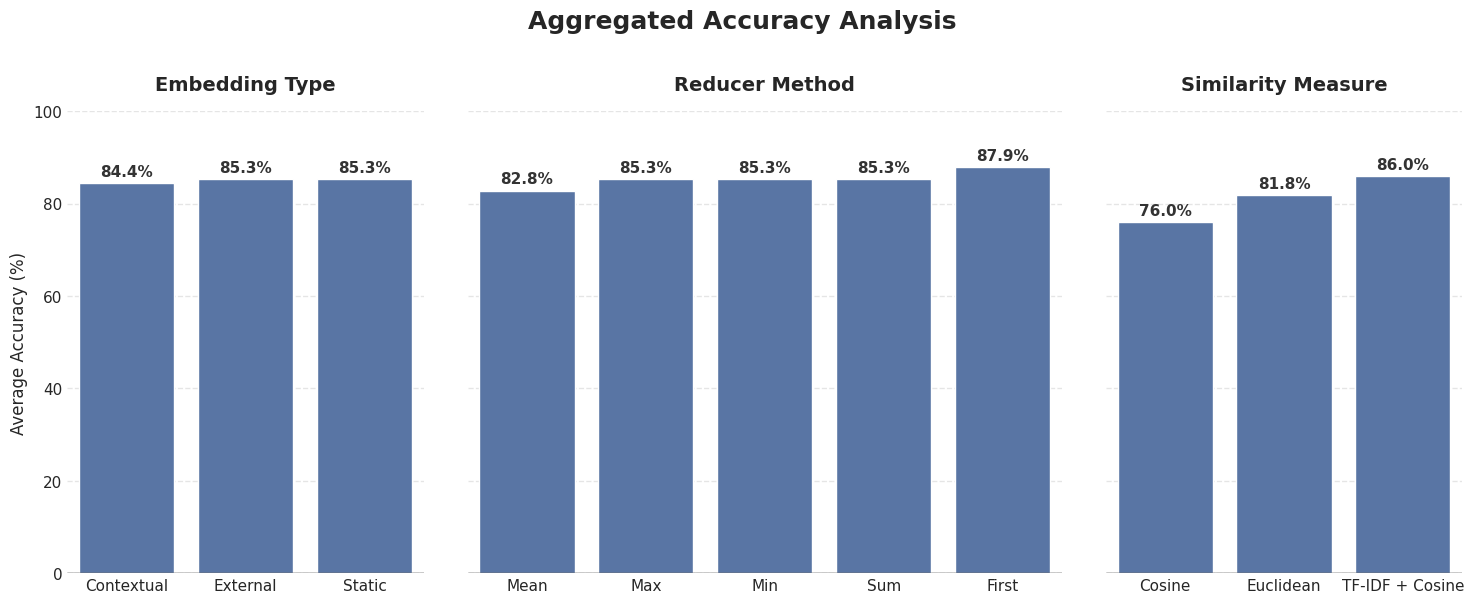}
    \caption{Validation of package stability across all explainer configuration knobs. The panels aggregate accuracy metrics by Embedding type, Reducer, and Similarity measure, confirming that the package correctly processes all permutations of the implemented modules.}
    \label{fig:components_benchmark}
\end{figure}

Figure~\ref{fig:components_benchmark} visualizes the aggregated performance metrics across these configurations. This analysis confirms that the package correctly handles the entire parameter space, demonstrating stability across diverse embedding strategies and reduction techniques.

\section{Shapley Values Approximation Methods Comparison}
\label{subsec:shapley_values__comparison}

This section evaluates the approximation methods (Sections~\ref{subsec:shapley_values__monte_carlo}-\ref{subsec:shapley_values__hierarchical}) to fulfill requirement \textbf{RR-4} (Appendix~\ref{app:requirements}), quantifying the trade-off between quality and cost. We use a tractable subset of \textit{VoiceBench\_\_single\_sentence} (Appendix~\ref{subsec:experiments__datasets__voice_bench}) containing $27$ samples with $9$--$10$ tokens. All estimators and variants (``limited''/``standard'') are benchmarked against ground-truth exact Shapley values (SV) under identical settings.

Throughout this section we use the text model \texttt{microsoft/phi-2}\footnote{\url{https://huggingface.co/microsoft/phi-2}} wrapped in the \texttt{trans-\allowbreak formers} interface. The model is configured to be as deterministic as possible (\texttt{temperature} $0$, \texttt{top\_k} = $1$), so that variance in the results is dominated by the sampling scheme rather than model stochasticity. Sampling budgets are expressed as fractions of the full coalition space, $m \in \{0.05, 0.10, \dots, 0.40\}$. For methods that support it, we additionally consider the minimal configuration in which only first-order omission coalitions are evaluated. Approximation quality is measured via accuracy (compared to the exact baseline) or Mean Absolute Error (MAE) and aggregated across samples as means with empirical $95\%$ confidence intervals if applicable. All experiments in this section employ the external embedding model \texttt{intfloat/e5-small-v2}\footnote{\url{https://huggingface.co/intfloat/e5-small-v2}} with utility function $\mathbf{U_3}$.

\begin{figure}[t!]
    \centering
    \includegraphics[width=\linewidth]{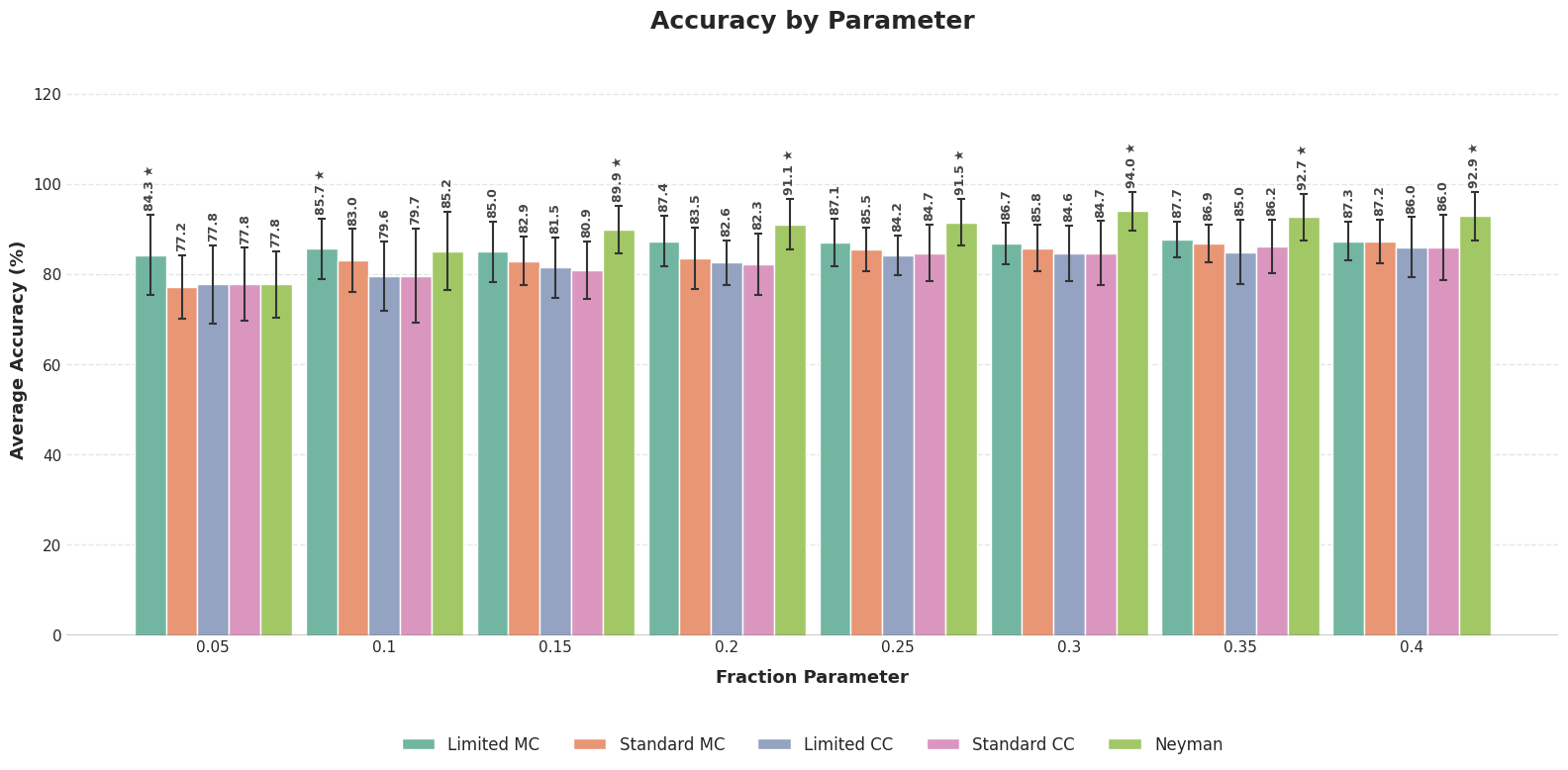}
    \caption{Approximation accuracy of all non-hierarchical methods relative to the exact Shapley values (SV) on the \textit{VoiceBench\_\_single\_sentence} subset. Bars denote mean accuracy; error bars indicate empirical $95\%$ confidence intervals across the 27 samples.}
    \label{fig:methods_comparison}
\end{figure}

Figure~\ref{fig:methods_comparison} summarizes the results for all non-hierarchical estimators. Across sampling fractions, the Neyman-allocation–based method achieves the highest average accuracy once its initial phase is completed. The limited Monte Carlo (MC) estimator, which enforces evaluation of all first-order omission coalitions, also performs competitively and consistently outperforms its standard counterpart that relies purely on random sampling. For the Complementary Contributions (CC) estimator the effect of limiting to first-order omissions is smaller, and its overall accuracy remains slightly below that of MC for comparable budgets. However, CC becomes an effective substrate for Neyman allocation: applying Neyman stratification on top of CC yields a clear and monotonic improvement, with the gap widening as the sampling fraction increases. As expected, larger sampling budgets lead to higher approximation quality, with gains tapering off beyond $m \approx 0.3$.

\begin{figure}[t!]
    \centering
    \includegraphics[width=0.65\linewidth]{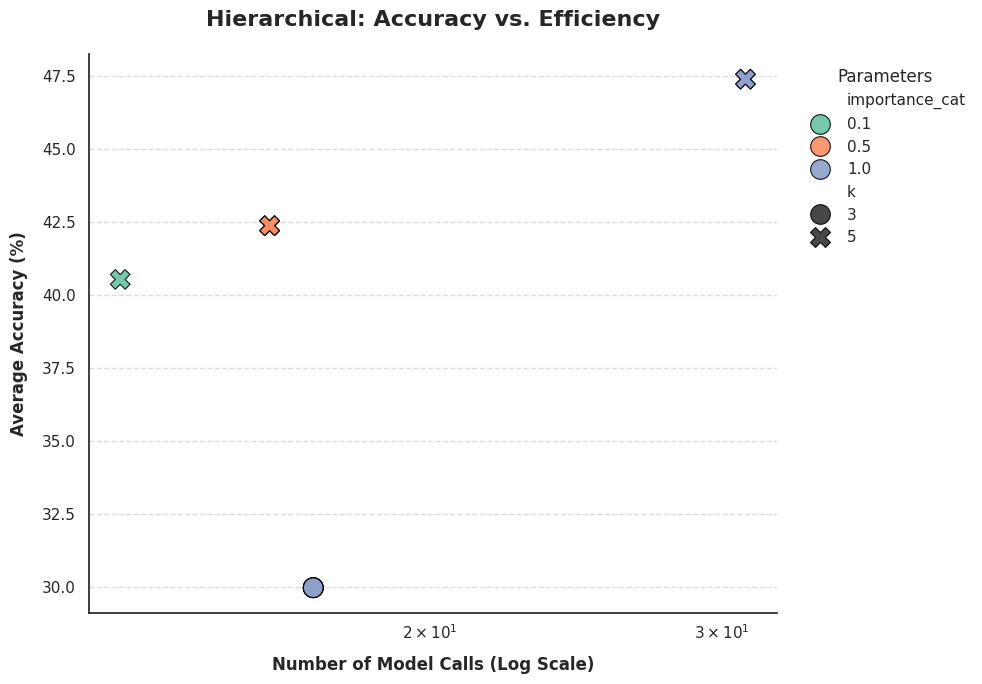}
    \caption{Accuracy–runtime trade-off for hierarchical configurations. The horizontal axis shows runtime in terms of the number of utility-function calls per explanation; points correspond to different combinations of group size $k$ and importance-sampling parameter.}
    \label{fig:methods_comparison_hier}
\end{figure}

The hierarchical estimator exhibits substantially lower accuracy (Figure~\ref{fig:methods_comparison_hier}). Varying the group size $k$ and the importance-sampling parameter produces the expected qualitative trends -- configurations with larger $k$ and stronger importance weighting are more accurate but require more utility evaluations -- yet all tested settings remain below $50\%$ mean accuracy. Given this combination of limited fidelity and non-trivial runtime, the hierarchical estimator is excluded from subsequent experiments. 

We interpret this primarily as a limitation of the concrete formulation studied here rather than of the hierarchical paradigm itself. In particular, bias accumulated at higher levels of the hierarchy propagates downward, leading to cascading errors. We partially address this by introducing a first-layer explainer that estimates group-level SV using a cheaper base method instead of distributing importance uniformly, but preliminary results did not yield sufficient improvements. Designing a more robust hierarchical scheme is therefore left as future work.

\begin{figure}[t!]
    \centering
    \includegraphics[width=0.85\linewidth]{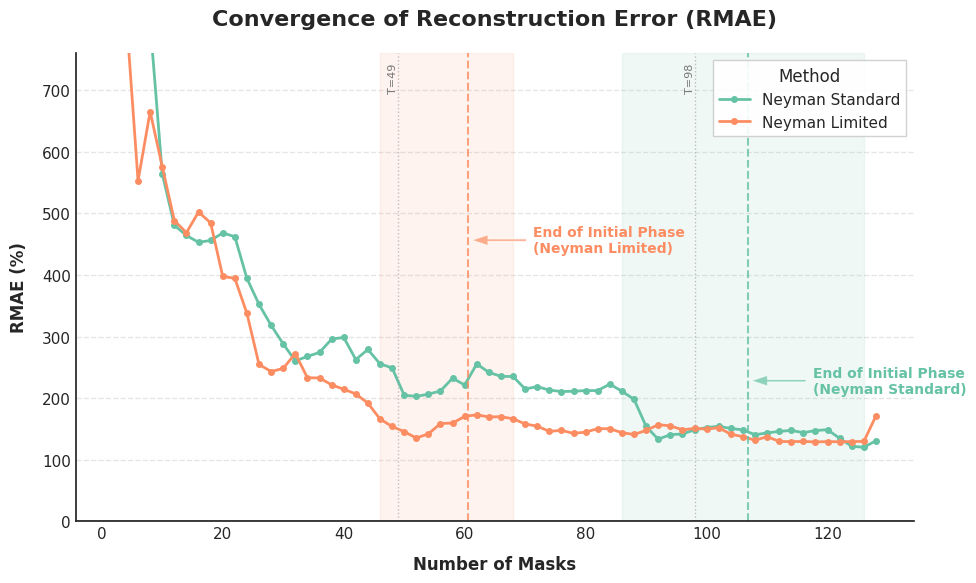}
    \caption{Convergence of standard and limited Neyman estimators on a single explanation. Relative Mean Absolute Error (RMAE) is plotted against the number of sampled masks (excluding the initial exact coalitions). Vertical lines indicate the end of the initial sampling phase ($n^2$ samples); shaded regions denote empirical $95\%$ confidence intervals.}
    \label{fig:neyman_trajectory}
\end{figure}

Finally, we investigate how quickly the Neyman-based estimator converges as a function of the number of sampled masks, with the aim of selecting a budget schedule suitable for longer inputs. Instead of fixing the sampling fraction, we consider a linearly growing budget and recompute the approximation after every pair of sampled masks, obtaining an error trajectory over time. Figure~\ref{fig:neyman_trajectory} presents MAE and relative MAE (RMAE) for both the standard and limited Neyman variants on a representative example.

\newpage
Both error measures decrease rapidly during the initial phase, which always consists of exactly $n^2$ samples; by the end of this phase the average MAE is already around $0.15$, and RMAE has dropped substantially. Additional samples further reduce the error, but with clearly diminishing returns. 

Importantly, the limited Neyman variant proposed in this thesis -- combining Neyman allocation over CC coalitions with the first-order omission constraint from limited MC -- consistently attains lower MAE than the standard formulation of \cite{shapleyapproximations} under the same budget. Based on these observations, and to retain a simple polynomial schedule for longer inputs, we adopt $3 n^2$ samples as a conservative default budget for Neyman-based explainers in all subsequent experiments.

\newpage
\section{SGPA Diagnostics}
\label{subsec:experiments__sgpa__diagnostics}

Since Spectrogram-Guided Phonetic Alignment (SGPA) changes the \emph{player definition} of the cooperative game, it is important to quantify both its operational benefit (feasibility) and the attribution-level shifts it induces. To this end, we perform a controlled comparison on the \textit{VoiceBench\_\_single\_sentence} (refer to Section~\ref{subsec:experiments__datasets}) setting for speech-to-speech modes (SM2S, SF2S -- male and female voices on input, respectively), computing SV under two segmentation schemes: (i) the model's native audio tokenization and (ii) SGPA word-aligned segments. All remaining estimator and inference settings are held constant.

Detailed logs and reproduction notebooks for this experiment are available in the project repository\footnote{\url{https://github.com/Pawlo77/MLLM-Shap/blob/feature/multi-sentence-analysis/experiments/analysis/}}.

\begin{figure}[h!]
    \centering
    \includegraphics[width=0.70\linewidth]{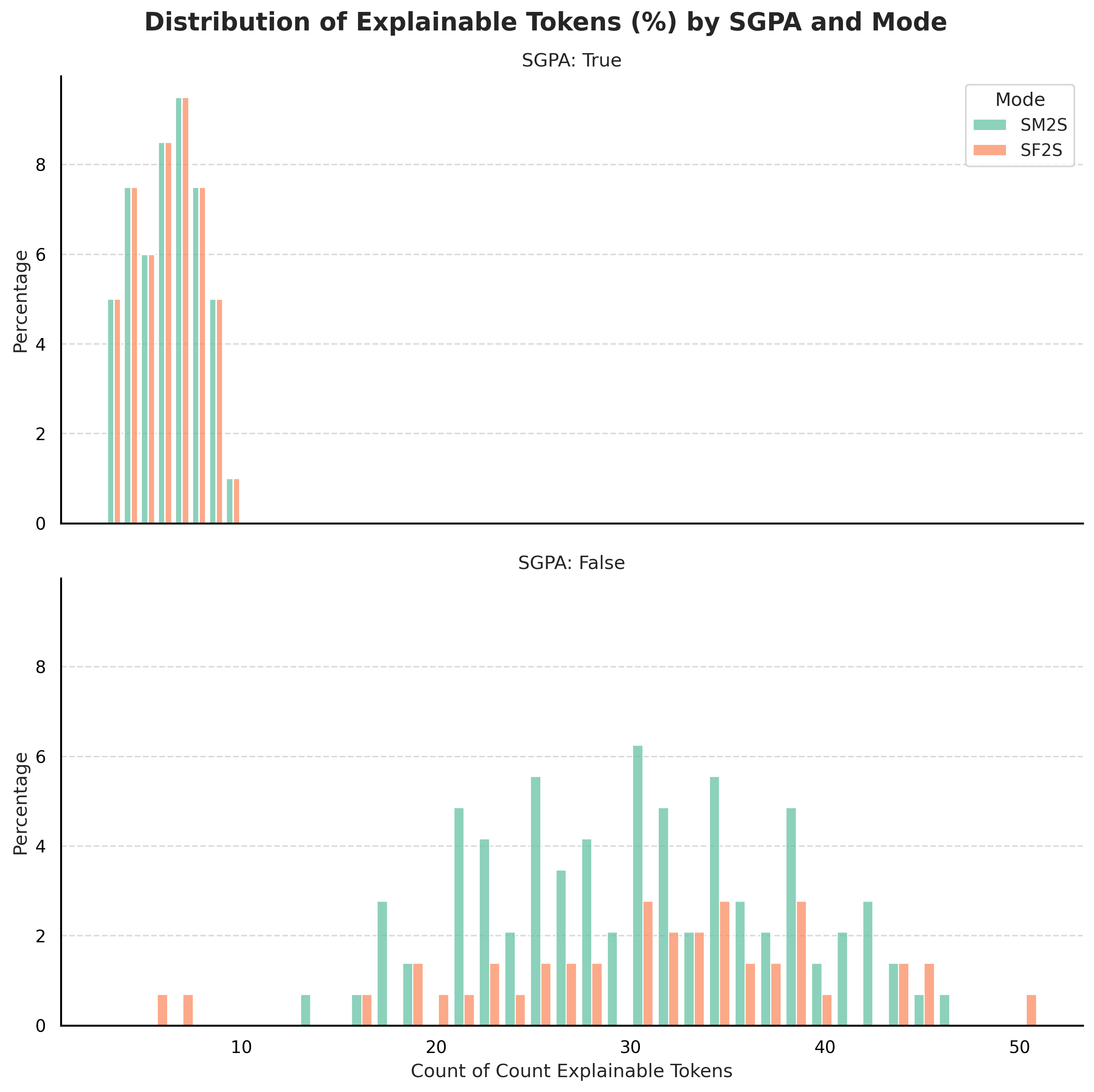}
    \caption{Distribution of explainable tokens (\%) by Spectrogram-Guided Phonetic Alignment (SGPA) setting and mode. With SGPA enabled, explainable lengths concentrate in a narrow, low-count regime; without SGPA, lengths expand substantially and become more dispersed, increasing the practical cost of Shapley values (SV) estimation.}
    \label{fig:experiments__sgpa__token_count_distribution}
\end{figure}

\paragraph{Reduction of the explainable sequence length}
One of the main insights gained from our study is that SGPA compresses the effective game size by operating on aligned word-level segments instead of dense audio token sequences. Figure~\ref{fig:experiments__sgpa__token_count_distribution} illustrates that SGPA yields a compact and stable explainable-length regime, whereas native tokenization induces substantially longer sequences and higher variance, which directly increases coalition complexity and estimator cost.

\paragraph{Compute feasibility and run stability}
Table~\ref{tab:experiments__sgpa__performance} summarizes the resulting cost profile. SGPA reduces the mean number of model calls and wall-clock runtime by over an order of magnitude for both speech-to-speech modes, converting an otherwise long-running configuration into a tractable one. In addition, Table~\ref{tab:experiments__sgpa__neyman_steps} shows that enabling SGPA changes the distribution of Neyman stages reached during estimation, consistent with operating in a smaller-game regime with different initialization and variance characteristics.

\begin{table}[h!]
\centering
\caption{Cost statistics for speech-to-speech Shapley values (SV) estimation with and without Spectrogram-Guided Phonetic Alignment (SGPA) on \textit{VoiceBench\_\_single\_sentence}. Without SGPA, the game size induced by native audio tokenization substantially increases the number of required model calls and runtime.}
\begin{tabular}{llcc}
\toprule
\textbf{SGPA} & \textbf{mode} & \textbf{n\_calls (mean)} & \textbf{runtime (s, mean)} \\
\midrule
True  & SM2S & 59.42  & 67.53 \\
True  & SF2S & 59.32  & 66.08 \\
False & SM2S & 2365.76 & 1696.49 \\
False & SF2S & 2552.14 & 1819.88 \\
\bottomrule
\end{tabular}
\label{tab:experiments__sgpa__performance}
\end{table}

\begin{table}[t!]
\centering
\caption{Proportion of observations reaching Neyman allocation stages (Section~\ref{subsec:shapley_values__neyman}) under Spectrogram-Guided Phonetic Alignment (SGPA) and native tokenization for speech-to-speech runs.}
\begin{tabular}{llcc}
\toprule
\textbf{SGPA} & \textbf{mode} & \textbf{Step 1} & \textbf{Step 2} \\
\midrule
True  & SM2S & 0.37 & 0.63 \\
True  & SF2S & 0.37 & 0.63 \\
False & SM2S & 0.00 & 1.00 \\
False & SF2S & 0.02 & 0.98 \\
\bottomrule
\end{tabular}
\label{tab:experiments__sgpa__neyman_steps}
\end{table}

\begin{figure}[t!]
    \centering
    \includegraphics[width=0.70\linewidth]{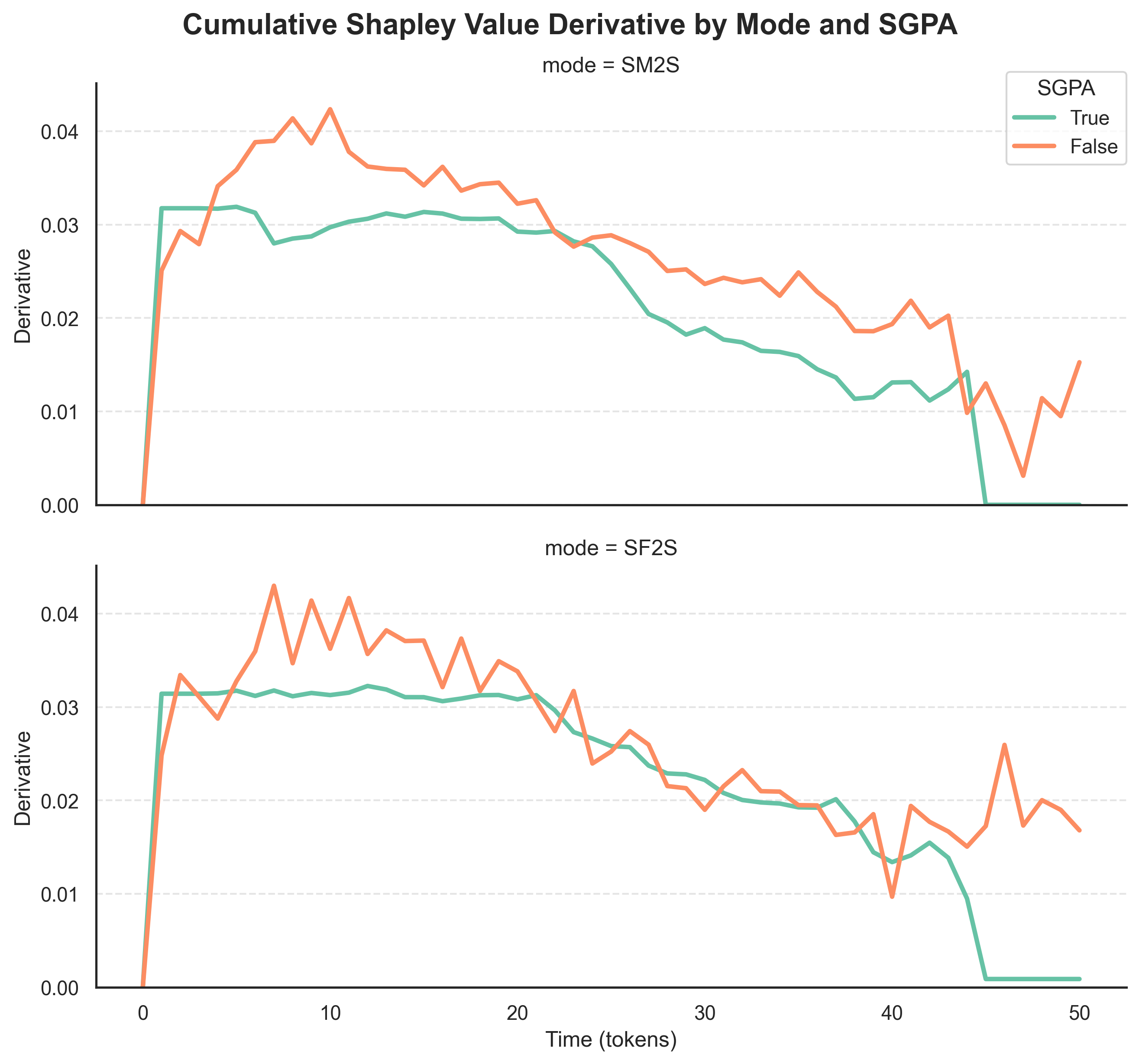}
    \caption{Derivative comparison under Spectrogram-Guided Phonetic Alignment (SGPA) and native tokenization, stratified by mode. SGPA modifies local attribution density (the fine-scale derivative structure) while preserving the broader accumulation profile.}
    \label{fig:experiments__sgpa__derivative_compare}
\end{figure}

\paragraph{Preservation of cumulative trends and changes in local dynamics}
A key methodological question is whether SGPA alters the qualitative attribution trajectory. Figure~\ref{fig:experiments__sgpa__derivative_compare} shows that SGPA primarily alters local allocation density while maintaining the macro-shape of ``early mass'' followed by gradual decay. This behavior is expected: SGPA aggregates fine-grained audio units into semantically interpretable segments, which dampens high-frequency variations attributable to micro-boundary differences under native tokenization.

\begin{figure}[t!]
    \centering
    \includegraphics[width=0.70\linewidth]{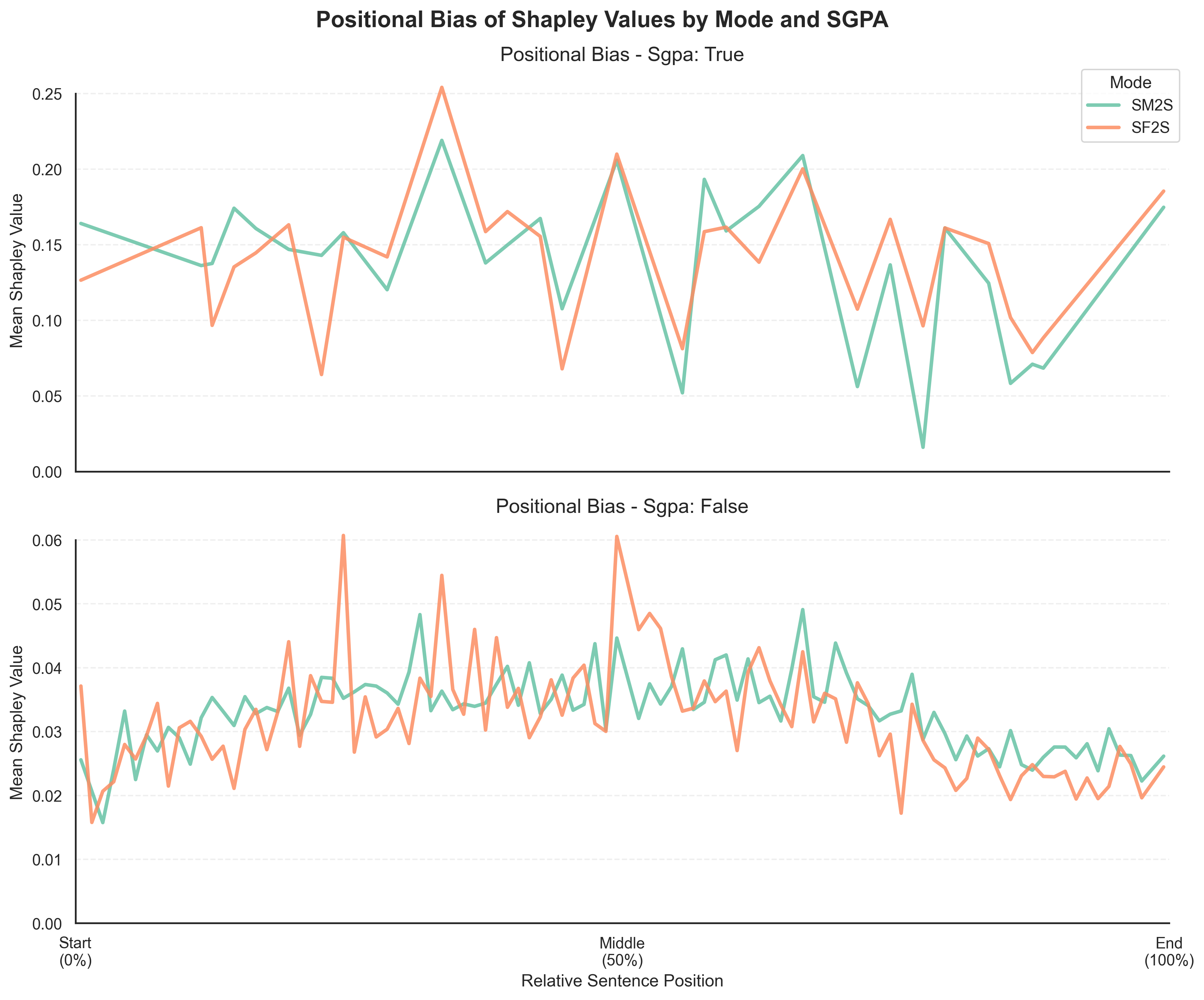}
    \caption{Positional bias of mean Shapley value (SV) by mode under Spectrogram-Guided Phonetic Alignment (SGPA) (top) and native tokenization (bottom). SGPA produces a segment-level positional diagnostic, whereas native tokenization reflects dense audio-unit behavior that is more sensitive to micro-level boundary effects.}
    \label{fig:experiments__sgpa__positional_bias}
\end{figure}

\paragraph{Position-normalized behavior}
Figure~\ref{fig:experiments__sgpa__positional_bias} shows position-normalized mean SV curves (relative sentence position). The SGPA setting yields a coarser but more interpretable positional diagnostic because each point corresponds to an aligned segment rather than a dense audio token. In contrast, native tokenization produces smoother curves but conflates model-internal token boundaries with acoustic boundary artifacts introduced by masking.

\begin{figure}[t!]
    \centering
    \includegraphics[width=0.70\linewidth]{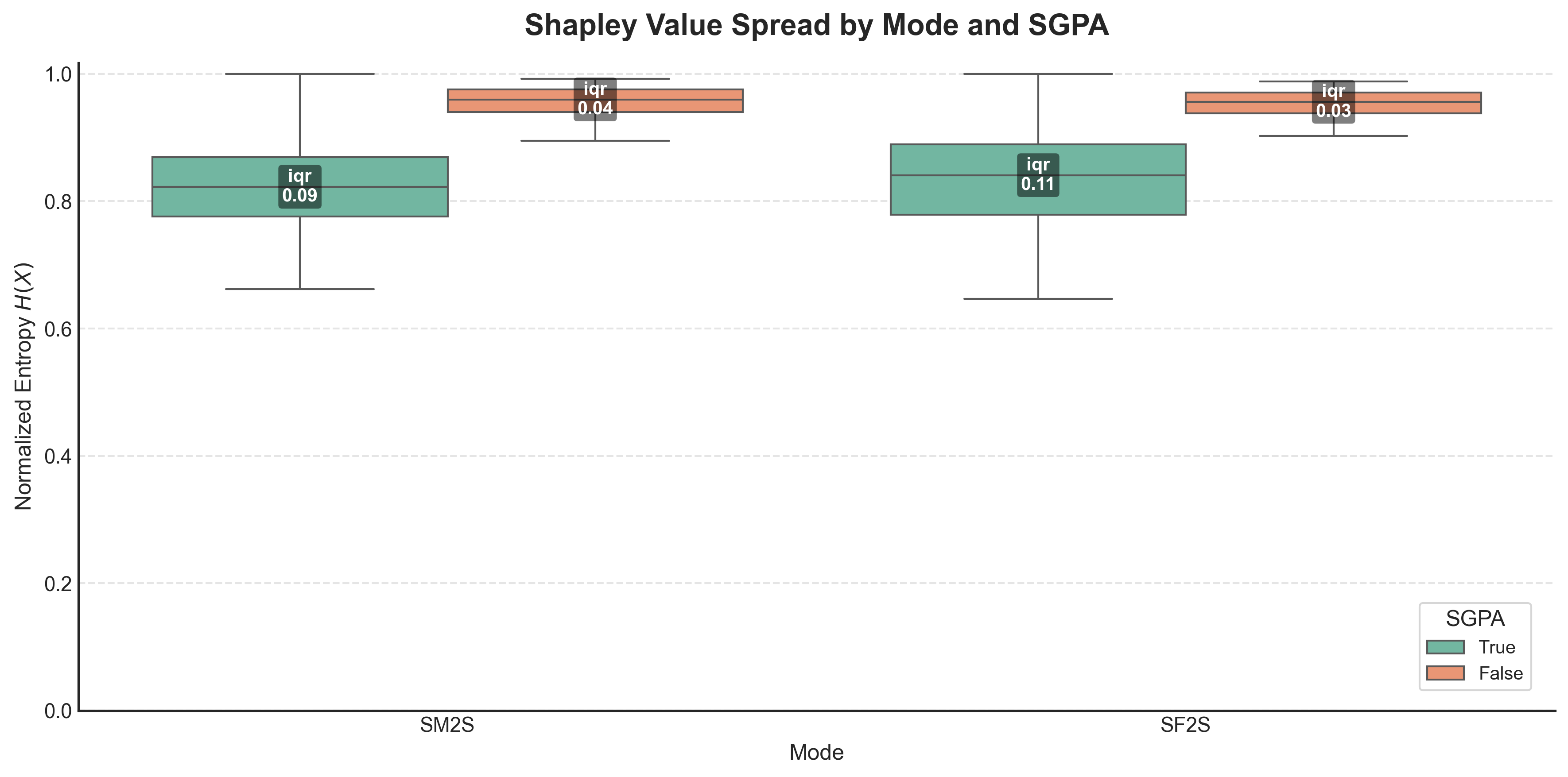}
    \caption{Normalized attribution entropy by mode with and without Spectrogram-Guided Phonetic Alignment (SGPA). SGPA expands the interquartile range of entropy normalized by the square root of the sample token count, reflecting a broader set of concentration regimes across observations under segment-level attribution.}
    \label{fig:experiments__sgpa__entropy}
\end{figure}

\paragraph{Attribution spread and concentration proxies}
Figure~\ref{fig:experiments__sgpa__entropy} summarizes attribution spread via normalized entropy (by the square root of the sample token count, as explained in Chapter~\ref{sec:experiments__methodology}). SGPA preserves the global trajectory of the cumulative SV while expanding the \emph{effective range} of normalized entropy (larger interquartile ranges), indicating that attribution concentration becomes less stable across observations. In practical terms, the segment-level game induced by SGPA tends to reduce the appearance of consistently sharp, highly concentrated attribution patterns relative to the model-native tokenization baseline. This shift is not incidental: it follows from redefining the SV players as externally aligned segments that need not coincide with the model’s internal audio units, thereby redistributing salience over a different partition of the input.

\newpage
For completeness, Table~\ref{tab:experiments__sgpa__ttest} presents paired significance testing across standard per-observation SV summaries and spread proxies, including Gini coefficient, top-$20$-mass, and normalized entropy (by the square root of the sample token count). Nearly all comparisons involving the SGPA condition are statistically significant under significance level $\alpha = 0.05$ with Bonferroni adjustment. This confirms that SGPA is not a neutral transformation; rather, it fundamentally alters attribution statistics by modifying the segmentation of the game.

\begin{table}[t!]
\centering
\caption{Paired significance tests of per-observation Shapley value (SV) summary metrics across speech-to-speech modes, comparing samples with and without Spectrogram-Guided Phonetic Alignment (SGPA). Significance level is $\alpha = 0.05$ with Bonferroni correction. Entropy values are normalized by the square root of the sample token count.}
\begin{tabular}{lcccc}
\toprule
\textbf{Mode} & \textbf{Metric} & \textbf{$p$-value} & \textbf{Cohen's $d$} & \textbf{Significant} \\
\midrule
\multirow{3}{*}{SM2S} & Gini                  & $< 0.01$ & 0.50  & True \\
                       & entropy (normalized)  & $< 0.01$ & -1.37 & True \\
                       & top-$20$-mass         & $< 0.01$ & 0.86  & True \\
\addlinespace
\multirow{3}{*}{SF2S} & Gini                  & 0.01      & 0.21  & False \\
                       & entropy (normalized)  & $< 0.01$ & -0.97 & True \\
                       & top-$20$-mass         & $< 0.01$ & 0.72  & True \\
\bottomrule
\end{tabular}
\label{tab:experiments__sgpa__ttest}
\end{table}

\paragraph{Limitations and segmentation-induced bias}

Spectrogram-Guided Phonetic Alignment (SGPA) is introduced to make Shapley values (SV) estimation feasible and interpretations semantically legible, but it necessarily introduces methodological trade-offs that must be stated explicitly.

\begin{itemize}
    \item \textbf{Attribution depends on the chosen player partition.}
By construction, SV explanations are defined with respect to a particular set of players (input units). SGPA changes the player partition from model-native audio units to word-aligned segments, which alters the cooperative game being solved and, therefore, changes the attribution statistics.

    \item \textbf{Mismatch with model-native audio tokenization.}
End-to-end audio models typically rely on internal audio tokenizations that need not align with word boundaries. SGPA, therefore, evaluates salience on a segmentation that is external to, and generally distinct from, the one implicitly used by the model. This mismatch can smooth or redistribute attribution mass by aggregating multiple internal units into a single segment and can increase prompt-dependent variability when alignment boundaries shift.

    \item \textbf{Model-agnostic segmentation as a comparability mechanism.}
SGPA also provides a practical comparability mechanism for cross-model studies: because segmentation is performed externally (via transcript, Connectionist Temporal Classification (CTC) alignment, and acoustic refinement), it does not require access to a model's internal tokenization scheme. This makes SGPA suitable for comparisons across models with different audio encoders and token granularities, and it moves the analysis toward a black-box regime where the attribution units are defined independently of architecture-specific internals.
\end{itemize}

\chapter{Experiment Methodology}
\label{sec:experiments__methodology}

Before presenting the main experimental comparisons, we briefly outline what constitutes a ``signal'' in our analysis and how to read the results. 
For each prompt, Shapley value (SV) assigns an attribution score to each explainable unit (token or segment), yielding a distribution over the input sequence. 
We report three complementary views of this distribution: (i) \emph{token-level structure} (e.g., whether SV peaks align with simple linguistic proxies), 
(ii) \emph{aggregate concentration and spread} (summary metrics such as mean/max SV, entropy normalized by the square root of the sample token count, and Gini), and 
(iii) \emph{position-resolved dynamics} (cumulative and derivative trajectories after normalization to account for modality-dependent token counts). 
This ordering is intentional: we start with lightweight interpretability diagnostics and then move to hypothesis tests that quantify whether observed differences are systematic across the dataset.

\paragraph{Length effects and metric choice.}
Across languages and modalities, inputs differ in the number of explainable units $n$ (native tokenization and SGPA segmentation induce different supports). This directly affects length-sensitive summaries: for longer sequences, per-unit means tend to decrease, and maxima become less comparable because semantically coherent words may be split into multiple tokenizer units. For this reason, we treat mean/max SV as auxiliary context and use concentration metrics as the primary ``focus'' signals under varying $n$:

\begin{itemize}
    \item \textbf{Top-20\% mass} -- 
        For each observation, we compute absolute attributions and normalize them to a probability mass function:
        \[
        \tilde{s}_i \;=\; \frac{|s_i|}{\sum_{j=1}^{n} |s_j|},
        \]
        sort them in descending order $\tilde{s}_{(1)} \ge \tilde{s}_{(2)} \ge \dots \ge \tilde{s}_{(n)}$, and sum the top $20\%$:
        \[
        \texttt{top-$20$-mass} \;=\; \sum_{i=1}^{\lceil 0.2n \rceil} \tilde{s}_{(i)}
        \]
        Higher values indicate that a small fraction of the input carries a large fraction of the explanatory mass.

    \item \textbf{Gini coefficient} --
        We compute the Gini coefficient on the same normalized weights $\tilde{s}_i$ as a length-robust measure of inequality of attribution allocation (higher means more concentrated):
        \[
        G \;=\; \frac{1}{2n}\sum_{i=1}^{n}\sum_{j=1}^{n} \left|\tilde{s}_i - \tilde{s}_j\right|
        \]

    \item \textbf{Normalized entropy} --
        Entropy quantifies the spread of the normalized attributions $\tilde{s}_i$:
        \[
        H \;=\; -\sum_{i=1}^{n} \tilde{s}_i \log \tilde{s}_i.
        \]
        To account for sequences of different lengths, we divide by the square root of the sequence length $n$:
        \[
        H_{\text{normalized}} \;=\; \frac{H}{\sqrt{n}}.
        \]
        Higher values indicate a more uniform (less concentrated) attribution distribution, while lower values indicate that a few tokens dominate the explanation.
\end{itemize}

\begin{figure}[t!]
    \centering
    \includegraphics[width=0.75\linewidth]{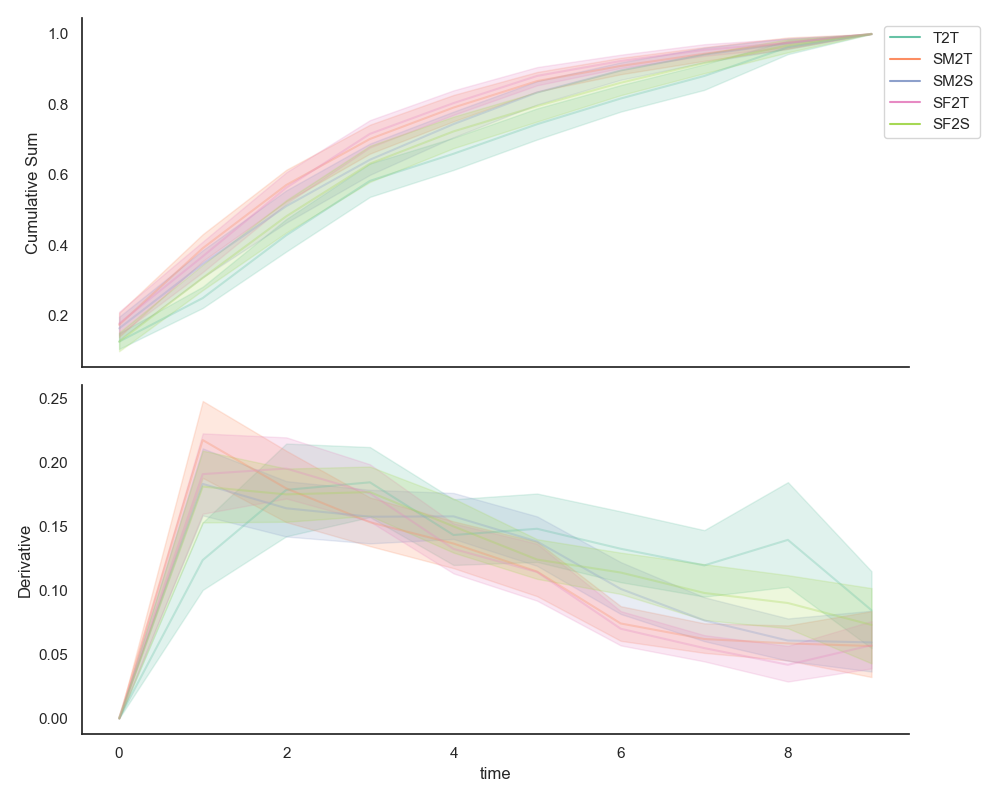}
    \caption{The top chart presents the interpolated cumulative sum of Shapley values (SV) by mode, while the bottom chart shows the corresponding derivative. The time axis represents the cumulative token count. Results are based on the \textit{VoiceBench\_\_single\_sentence} dataset.}
    \label{fig:experiments__multi_modal_context__results__cumsum}
\end{figure}

\paragraph{Position-resolved comparisons.}
As shown in Figure~\ref{fig:experiments__multi_modal_context__run__token_count_distribution}, while token distributions are broadly consistent across modes, minor variations in sequence length between Text-to-Text (T2T) and Speech modes complicate the direct comparison of attribution curve shapes. To resolve this, we apply linear interpolation to normalize these varying lengths onto a unified interval $[0, 1]$. This technique is used exclusively for the cumulative and derivative visualizations (e.g., Figure~\ref{fig:experiments__multi_modal_context__results__cumsum}) to facilitate the comparison of information accumulation dynamics. Crucially, all statistical hypothesis testing is performed on the unaltered, original token counts.

\paragraph{Paired t-test.}
For paired conditions (the same prompts evaluated under two modes, $X$ and~$Y$), we assume the pairwise differences are independent and identically distributed (i.i.d.) and follow a normal distribution. We test the null hypothesis:
\[
H_0:\; \mathbb{E}[X-Y] = 0.
\]
Let $d_i = x_i - y_i$ denote the observed differences for $i=1,\dots,n$. We compute the sample mean difference $\bar{d}$, sample variance $s_d^2$, and standard error $SE_{\bar{d}}$ as follows:
\[
\bar{d}=\frac{1}{n}\sum_{i=1}^{n} d_i,
\qquad
s_d^2=\frac{1}{n-1}\sum_{i=1}^{n}(d_i-\bar{d})^2,
\qquad
SE_{\bar{d}}=\frac{s_d}{\sqrt{n}}.
\]
The test statistic is defined as
\[
T \;=\; \frac{\bar{d}}{SE_{\bar{d}}} \;=\; \frac{\bar{d}}{s_d/\sqrt{n}} \;\overset{H_0}{\sim}\; t_{n-1},
\]
where $t_{n-1}$ denotes the Student's $t$-distribution with $n-1$ degrees of freedom. Moreover, 

\begin{itemize}
    \item Unless stated otherwise, we report two-sided $p$-values calculated for the observed statistic $t_{\text{obs}}$ \parencite{student1908probable}:
        \[
        p = 2 \cdot \min \left( \mathbb{P}(T \ge t_{\text{obs}}), \mathbb{P}(T \le t_{\text{obs}}) \right).
        \]
    \item We report effect size for paired differences using Cohen's standardized mean difference (\textbf{Cohen's $d$}, \cite{cohen1988statistical}.):
         \[
        d_{\text{Cohen}} \;=\; \frac{\bar{d}}{s_d},
        \]

    \item When conducting a family of $m$ hypothesis tests, we apply the Bonferroni correction by adjusting the significance level to $\alpha' = \alpha / m$ \parencite{dunn1961multiple}. This method controls the family-wise error rate (FWER), reducing the likelihood of committing a Type I error (false positive) across the set of simultaneous tests.

    \item We verify the assumptions of the paired parametric t-test on the difference samples $\{d_i\}$ using the Shapiro-Wilk normality test \parencite{shapiro1965analysis}. Under the null hypothesis $H_0$, the sample is assumed to be drawn from a normal distribution. The test statistic is defined as
        \[
        W = \frac{\left( \sum_{i=1}^{\lfloor n/2 \rfloor} a_i \bigl(x_{(n-i+1)} - x_{(i)}\bigr) \right)^2}{\sum_{i=1}^{n} (x_i - \bar{x})^2},
        \]
        where $x_{(i)}$ denotes the $i$-th order statistic and $a_i$ are tabulated constants. When the test indicates violations of normality (e.g., substantial skewness or heavy-tailed behavior), the corresponding parametric results are not interpreted as evidential. These diagnostics are applied uniformly across all reported comparisons, but their detailed outcomes are omitted from the results chapter to avoid unnecessary clutter.
\end{itemize}

\newpage
\paragraph{Linguistic analysis methods.}
For linguistic diagnostics we use \texttt{spaCy}\footnote{\url{https://pypi.org/project/spacy/}} \parencite{honnibal2017spacy2} as a consistent, automated proxy for dependency structure across languages. We run language-specific pipelines for dependency parsing: \texttt{en\_core\_web\_sm} (English), \texttt{es\_core\_news\_sm} (Spanish), and \texttt{fr\_core\_news\_sm} (French). Dependency labels are mapped into the importance tiers listed in Table~\ref{tab:experiments__multi_modal_context__results__spacy_deps}. Because this mapping is heuristic and not a linguistically exhaustive taxonomy, it was sanity-checked with LLM-based review and spot-verified on representative samples. Example token-level features and the resulting importance assignment are illustrated in Table~\ref{tab:experiments__multi_modal_context__results__token_analysis}.

\begin{table}[h!]
\centering
\caption{\texttt{spaCy} Dependency Labels with Importance and Grouping. Importance $1$ represents most important dependencies from linguistic perspective.}
\label{tab:experiments__multi_modal_context__results__spacy_deps}

\begin{minipage}[t]{0.48\textwidth}
\vspace{0pt} 
\centering
\begin{tabular}{p{3.5cm} c l}
\toprule
\textbf{Group} & \textbf{Imp.} & \textbf{Dep} \\
\midrule
\multirow{5}{=}{\makecell[l]{Core meaning \\(highest importance)}} 
 & 1 & ROOT\_DEP \\
 & 1 & nsubj \\
 & 1 & nsubjpass \\
 & 1 & dobj \\
 & 1 & agent \\
\midrule
\multirow{6}{=}{\makecell[l]{Clause structure \\(very important)}} 
 & 2 & ccomp \\
 & 2 & xcomp \\
 & 2 & advcl \\
 & 2 & relcl \\
 & 2 & acl \\
 & 2 & pcomp \\
\midrule
\multirow{5}{=}{\makecell[l]{Nominal / verbal \\ complements}} 
 & 3 & attr \\
 & 3 & acomp \\
 & 3 & dative \\
 & 3 & pobj \\
 & 3 & oprd \\
\bottomrule
\end{tabular}
\end{minipage}%
\hfill
\begin{minipage}[t]{0.48\textwidth}
\vspace{0pt} 
\centering
\begin{tabular}{p{3.5cm} c l}
\toprule
\textbf{Group} & \textbf{Imp.} & \textbf{Dep} \\
\midrule
\multirow{7}{=}{\makecell[l]{Modifiers \\(less critical, \\descriptive)}} 
 & 4 & amod \\
 & 4 & advmod \\
 & 4 & npadvmod \\
 & 4 & nmod \\
 & 4 & compound \\
 & 4 & poss \\
 & 4 & appos \\
\midrule
\multirow{2}{=}{\makecell[l]{Coordination / \\structure helpers}} 
 & 4 & conj \\
 & 5 & cc \\
\midrule
\multirow{6}{=}{\makecell[l]{Function / \\grammar glue \\(lowest importance)}} 
 & 5 & det \\
 & 5 & case \\
 & 5 & aux \\
 & 5 & auxpass \\
 & 5 & mark \\
 & 5 & prep \\
\midrule
Punctuation & 6 & punct \\
\midrule
Generic / unknown & 4 & dep \\
\bottomrule
\end{tabular}
\end{minipage}

\end{table}

\chapter{Experiment Setup}
\label{sec:experiments}\label{sec:experiments_setup}

This chapter documents the experimental setup used to study how Multimodal Large Language Models (MLLMs) allocate explanatory mass across inputs under Shapley value (SV) analysis. The focus here is methodological: we describe the evaluated models and datasets, the audio preparation pipeline, feasibility constraints, and the infrastructure required to run long SV sweeps reproducibly. The empirical findings for multimodal and multilingual behavior are presented separately in Chapter~\ref{sec:experiments_results}.

A practical constraint shapes the remainder of this chapter and the scope of reported results. 
We initially surveyed multiple publicly available speech-capable MLLMs to ensure that the experimental design is not tailored to a single vendor ecosystem. 
However, SV estimation on speech inputs is computationally demanding (Appendix~\ref{app:compute_analysis}). While we did our best to supplement this demand with sufficient compute capabilities (see Appendix~\ref{app:compute_analysis}), ultimately our resources were limited to hardware outlined in Appendix~\ref{app:requirements_inventory}.
As a result, while we document the broader model selection rationale in Section~\ref{subsec:experiments__models}, the full experimental sweep reported in Chapter~\ref{sec:experiments_results} is executed on a single model chosen for feasibility and tooling stability, with the remaining candidates retained as reference points for future extension.

Throughout this work, \emph{modes} refer to input--output modality combinations. We distinguish between text (\textbf{T}) and speech (\textbf{S}), with speech further categorized by the input gender: female (\textbf{F}) or male (\textbf{M}). Under this convention, \textbf{T2T} denotes text-to-text; \textbf{SF2T} and \textbf{SM2T} represent speech-to-text using female and male audio, respectively; and \textbf{SF2S} and \textbf{SM2S} denote speech-to-speech for both genders. Aggregated results are reported as \textbf{S2T} (combining SF2T and SM2T), \textbf{S2S} (combining SF2S and SM2S) and sometimes using notation like \textbf{S2*} (combining all modes with audio input) or analogous notation for other modes of input and output. Finally, \textbf{IAF\_SM2T} and \textbf{ITF\_SM2T} refer to interleaved male audio and text inputs with text outputs, as detailed in Section~\ref{subsec:experiments__multi_modal_context}.

\section{Models overview}
\label{subsec:experiments__models}

The choice of models was, apart from the factors mentioned in Chapters~\ref{sec:introduction} and \ref{sec:business_goal}, further influenced by their performance on \textit{VoiceBench} \parencite{chen2024voicebench}, one of the most widely used benchmarks for audio-to-audio models. Our search was based primarily on the models listed in their official repository. For further details on the benchmark choice, please refer to Section~\ref{subsec:experiments__datasets}.

Additional important factors contributing to our final decision were (model size (for models hosted on-premise) or pricing (for models hosted online). Preference was given to models that are fast, cost-effective, and suitable for scalable systems or low-end devices such as mobile phones. With practical usability in mind, the focus was placed on models likely to appear in real-world scenarios rather than outdated ones.

Based on these requirements, we list the three shortlisted candidates below. The final execution scope is constrained to a single model for feasibility reasons, as discussed at the end of this section and quantified in Appendix~\ref{app:compute_analysis}.

\begin{itemize}

    \item \textit{Alibaba Cloud (Qwen Team)}, \textbf{Qwen2.5-Omni-3B}\footnote{https://huggingface.co/Qwen/Qwen2.5-Omni-3B} \parencite{xu2025qwen25omnitechnicalreport} -- a 3B end-to-end ``omni'' model released in July~2025, designed for natural speech interaction alongside text, image, audio, and video input. The vendor originates from China.

    \item \textit{Baichuan AI}, \textbf{Baichuan-Audio (Instruct)}\footnote{https://huggingface.co/baichuan-inc/Baichuan-Audio-Instruct} \parencite{li2025baichuan} -- a 7B model released at the beginning of 2025. The official training data is not available, but the model is mostly benchmarked in the creators' technical report using English- and Chinese-based datasets. The company originates from Beijing, China.
    
    \item \textit{Liquid AI}, \textbf{LFM2-Audio-1.5B}\footnote{https://huggingface.co/LiquidAI/LFM2-Audio-1.5B} \parencite{LIQUID_LFM2_AUDIO_2025} -- a 1.5B model released at the beginning of October 2025. While the official training data has not been released, the available benchmarking includes only English-dominant \textit{VoiceBench}. The company originates from Cambridge, Massachusetts, USA, so we assume its leading language is English.
    
\end{itemize}

\noindent One can note that \textit{LFM2-Audio-1.5B} is much smaller than the other two models and, as expected, achieves significantly worse results on \textit{VoiceBench}. However, it was the only model of its size we encountered that met our architecture-related criteria while being publicly available, so we use it as the main model in our experiments and development. In fact, we did not find any comparable model of this size, which still used the architecture targeted by our study.

General plan assumed at the beginning of study was to use all three models for multimodality experiments (Section~\ref{subsec:experiments__multi_modal_context}), but only \textit{Qwen2.5-Omni-3B} for multilingual experiments (Section~\ref{subsec:experiments__multi_language_context}), given its explicit multilingual speech support and mature tooling.

Due to computational constraints (refer to Appendix~\ref{app:compute_analysis}), \textbf{we restricted the scope of our experiments exclusively to the \textit{LFM2} model}. We retain the descriptions of the other models in this section to facilitate future validation or expansion of this work. Consequently, all experimental results presented in Chapter~\ref{sec:experiments_results} are based solely on \textit{LFM2}.

\begin{table}[t!]
\renewcommand{\arraystretch}{1.1}
\centering
\caption{VoiceBench results comparison between \textit{Qwen2.5-Omni-3B}, \textit{Baichuan-Audio}, and \textit{LFM2-Audio-1.5B}. Scores for the first two models are taken directly from the official \textit{VoiceBench} repository, whereas the score for \textit{LFM2-Audio} is taken from its creator's official website as of \DTMdate{2025-10-03}.}
\begin{tabular}{lccc}

\toprule

\textbf{Section} & \textbf{Qwen2.5-Omni-3B} & \textbf{Baichuan-Audio} & \textbf{LFM2-Audio-1.5B} \\

\midrule

AlpacaEval  & 3.72   & 4.41   & 3.71   \\
CommonEval  & 3.51   & 4.08   & 3.49   \\
WildVoice   & 3.42   & 3.92   & 3.17   \\
SD-QA       & 44.94  & 45.84  & 30.56  \\
MMSU        & 55.29  & 53.19  & 31.95  \\
OBQA        & 76.26  & 71.65  & 44.40  \\
BBH         & 61.30  & 54.80  & 30.54  \\
IFEval      & 32.90  & 50.31  & 98.85  \\
AdvBench    & 88.46 & 99.42  & 67.33  \\
\hline
\textbf{Overall}     & \textbf{76.91}  & \textbf{69.27}  & \textbf{56.78}  \\

\bottomrule

\end{tabular}
\label{tab:experiments__models__voice_bench}
\end{table}

\section{Datasets overview}
\label{subsec:experiments__datasets}

Our experiments require short conversational prompts that (i) admit both text and speech inputs, (ii) remain low in linguistic complexity to match the capacity of the evaluated model and the computational cost of Shapley values (SV) estimation, and (iii) support sentence-level manipulation for mixed-modality settings.

We evaluate two data sources. First, \textit{VoiceBench}~\parencite{chen2024voicebench} serves as an established benchmark for speech-capable assistants; we derive two English-only subsets from it: \textit{VoiceBench\_\_single\_sentence} (used in experiment \textbf{E1}, as described in Section~\ref{subsec:experiments__multi_modal_context}) and \textit{VoiceBench\_\_multi\_sentence} (used in experiment \textbf{E2}, as described in Section~\ref{subsec:experiments__multi_modal_context}). Second, \textit{Infinity Instruct}~\parencite{li2025infinityinstruct} is used to construct \textit{InfinityInstruct\_\_multi\_lingual}, enabling controlled multilingual evaluation across English, Spanish, and French (used in experiments \textbf{E3}--\textbf{E4}, as described in Section~\ref{subsec:experiments__multi_language_context}).

Table~\ref{tab:experiments__datasets__summary} summarizes the final dataset sizes and basic statistics. All produced datasets are published on Hugging Face\footnote{\url{https://huggingface.co/datasets/Pawlo77/mllm-shap}}. Additional details on dataset construction, sampling, filtering, and representative examples are provided in Appendix~\ref{app:datasets_construction}.

\begin{table}[h!]
\centering
\caption{Summary statistics of datasets used in experiments.}
\begin{tabular}{p{5.3cm} p{0.8cm} p{1.2cm} p{1.7cm} p{1.7cm} p{2.8cm}}
\toprule
\textbf{name} & \textbf{rows} & \textbf{tokens (mean)} & \textbf{characters (mean)} & \textbf{sentences (mean)} & \textbf{Languages} \\
\midrule
VoiceBench\_\_single\_sentence & 100 & 7.19  & 37.87  & 1.0  & English (100\%) \\
VoiceBench\_\_multi\_sentence  & 100 & 27.54 & 139.27 & 2.44 & English (100\%) \\
InfinityInstruct\_\_multi\_lingual & 108 & 21.30 & 121.49 & 1.3 & English, Spanish, French (each 34\%) \\
\bottomrule
\end{tabular}
\label{tab:experiments__datasets__summary}
\end{table}

All audio recordings were artificially generated using \textit{Google Text-to-Speech}\footnote{\url{https://cloud.google.com/text-to-speech}}. The selected model was \textit{Chirp 3 HD}, the latest version recommended for general use according to Google’s documentation as of \DTMdate{2025-10-11}. All calls were made using the default voice settings, and prior to audio encoding, all sentences were stripped on both ends of punctuation and whitespace characters, including \texttt{.,!?:;"'()[]{}-\textbackslash n\textbackslash t}. The results can be seen in Table~\ref{tab:experiments__audio_pp__summary}.

\begin{table}[h!]
\centering
\caption{Summary statistics of datasets in terms of audio: mean duration in seconds for male and female audio encoding within one observation (joined sentences for each observation without any lag in between them) and total audio length.}

\begin{tabular}{p{5cm} p{1cm} p{1.5cm} p{1.5cm} p{2cm}}
\toprule

\textbf{name} & \textbf{male (mean)} & \textbf{female (mean)} & \textbf{total}  \\
\midrule

VoiceBench\_\_single\_sentence & 2.69 & 2.48 & 1,035.5 \\

VoiceBench\_\_multi\_sentence & 9.67 & 9.03 & 1,869.99 \\

InfinityInstruct\_\_multi\_lingual & 8.07 & 7.85 & 4,727.04 \\

\bottomrule
\end{tabular}
\label{tab:experiments__audio_pp__summary}
\end{table}

Since some experimental scenarios require separate sentence-level audio encodings, all sentences were generated individually to eliminate potential bias arising from variations in speech representations caused by punctuation in multi-sentence inputs. For cases where multiple sentences needed to be combined, the resulting audio files were concatenated with a $0.2$-second pause between each segment.

It should also be noted that for the purposes of our experiments, for all audio samples, we use Spectrogram-Guided Phonetic Alignment (SGPA) pre-processing step described in detail in Chapter~\ref{sec:experiments__sgpa}, which is required due to prohibitive cost of audio segmentation (explored in-depth in Appendix~\ref{app:compute_analysis}) when using native segmentation of the models.

\chapter{Experiment Results}
\label{sec:experiments_results}

Having specified the experimental setup, feasibility constraints, and the infrastructure that enables reproducible Shapley values (SV) computation (Chapter~\ref{sec:experiments_setup}), we now report empirical findings. We first show a representative example to ground the experiment results in Section~\ref{subsec:experiments__initial_exploration}. The results are presented in two parts. Section~\ref{subsec:experiments__multi_modal_context} focuses on multimodal behavior, including mode-dependent effects and interleaved input sequencing. Section~\ref{subsec:experiments__multi_language_context} evaluates multilingual prompts and tests whether modality effects and linguistic alignment patterns persist across languages. Methodology used during the experiments was covered in Chapter~\ref{sec:experiments__methodology}.

\section{Initial Exploration}
\label{subsec:experiments__initial_exploration}

To ground the interpretation, we include a minimal example based on the prompt \textit{``Who developed reCAPTCHA''}. Explanatory model response (LFM2) for this question is:

\begin{quote}
    \textit{reCAPTCHA was developed by \textbf{Luis von Ahn} at Carnegie Mellon University and later acquired by \textbf{Google} in 2009. It was created to help verify that users are human while simultaneously digitizing books and improving image recognition.}
\end{quote}

For the run used in Chapter~\ref{sec:experiments_results}, the explainable units (native tokenizer fragments) and their SV values are shown in Table~\ref{tab:experiments__sanity__recaptcha_tokens}.

\begin{table}[htbp]
\centering
\caption{SV allocation for the prompt ``Who developed Recaptcha''. The word \textit{Recaptcha} is split into subword units under model-native tokenization.}
\renewcommand{\arraystretch}{1.1}
\begin{tabular}{r l r}
\toprule
\textbf{ID} & \textbf{Token} & \textbf{SV} \\
\midrule
1 & \texttt{Who}       & 0.1231 \\
2 & \texttt{developed} & 0.3637 \\
3 & \texttt{Rec}       & 0.3862 \\
4 & \texttt{apt}       & 0.1270 \\
5 & \texttt{cha}       & 0.0000 \\
\bottomrule
\end{tabular}
\label{tab:experiments__sanity__recaptcha_tokens}
\end{table}

\newpage
The distribution is sharply concentrated on the predicate (\texttt{developed}) and the entity anchor (\texttt{Rec...}), which is consistent with how the prompt constrains the response: it asks for an origin/creator relation tied to a specific named concept. At the same time, the example highlights a practical interpretability caveat used throughout this thesis: explanations are assigned to \emph{tokenizer units}, so a single human word may be represented by multiple fragments, and attribution can be unevenly split across them. This is one reason we rely on concentration metrics (top-$20$-mass, Gini), position-normalized views, and normalized entropy when drawing cross-condition conclusions at scale. We also use linguistic analysis, details of which were presented in Chapter~\ref{sec:experiments__methodology}. Such analysis performed on our representative prompt is shown in Table~\ref{tab:experiments__multi_modal_context__results__token_analysis}.

\begin{table}[h!]
\centering
\caption{Token-level linguistic features and importance scores for the prompt ``Who developed Recaptcha''. Shapley values (SV) are rounded to 2 meaningful places.}
\begin{tabular}{lcccccc}
\hline
\textbf{token} & \textbf{SV} & \textbf{dep} & \textbf{pos} & \textbf{children} & \textbf{importance} & \textbf{group} \\ \hline
Who & 0.12 & nsubj & PRON & 0 & 1 & core\_structure \\
developed & 0.36 & ROOT & VERB & 2 & 1 & core\_structure \\
Rec & 0.39 & dobj & PROPN & 0 & 1 & core\_structure \\
apt & 0.13 & ROOT & VERB & 2 & 1 & core\_structure \\
cha & 0.00 & dobj & PROPN & 0 & 1 & core\_structure \\ \hline
\end{tabular}
\label{tab:experiments__multi_modal_context__results__token_analysis}
\end{table}

\section{Multimodal Context Understanding}
\label{subsec:experiments__multi_modal_context}

These experiments aim to analyze how input and output modalities affect model focus. We employ our model in the following cases:

\begin{itemize}
    \item \textbf{E1}: Using \textit{VoiceBench\_\_single\_sentence}, we validate whether input modality, output modality, and speaker gender make a difference. We evaluate $6$ scenarios: text-only input / audio-only input (male, female) and text-only output / audio-only output.
    
    \item \textbf{E2}: Using \textit{VoiceBench\_\_multi\_sentence}, we validate whether mixing input modalities makes a difference. We evaluate $4$ scenarios: text-only input / audio-only input / text and audio interleaved (by sentence). Given a sequence of sentences $s_1, s_2, \dots$, we create two cases: in the first, $s_1, s_3, \dots$ are fed as audio and the remaining as text; in the second, the assignment is reversed.
\end{itemize}

All runs use the same system prompt and inference pipeline as in previous experiments; we do not introduce any additional modality-specific instruction beyond changing the dataset input format itself.

\newpage
For reporting convenience in \textbf{E2}, we denote the two interleaved variants as \textbf{ITF\_SM2T} (text-first interleaving) and \textbf{IAF\_SM2T} (audio-first interleaving) in tables and figures as was mentioned in Chapter~\ref{sec:experiments_setup}.

The rest of this section is structured as follows: Section~\ref{subsec:experiments__multi_modal_context__run} presents run statistics, Section~\ref{subsec:experiments__multi_modal_context__results} reports the results, and Section~\ref{subsec:experiments__multi_modal_context__conclusions} summarizes conclusions. Detailed logs and reproduction notebooks for this experiment are available in the project repository\footnote{\url{https://github.com/Pawlo77/MLLM-Shap/blob/feature/multi-sentence-analysis/experiments/analysis/}}.

\subsection{Run statistics and details}
\label{subsec:experiments__multi_modal_context__run}

\begin{figure}[h!]
    \centering
    \includegraphics[width=0.75\linewidth]{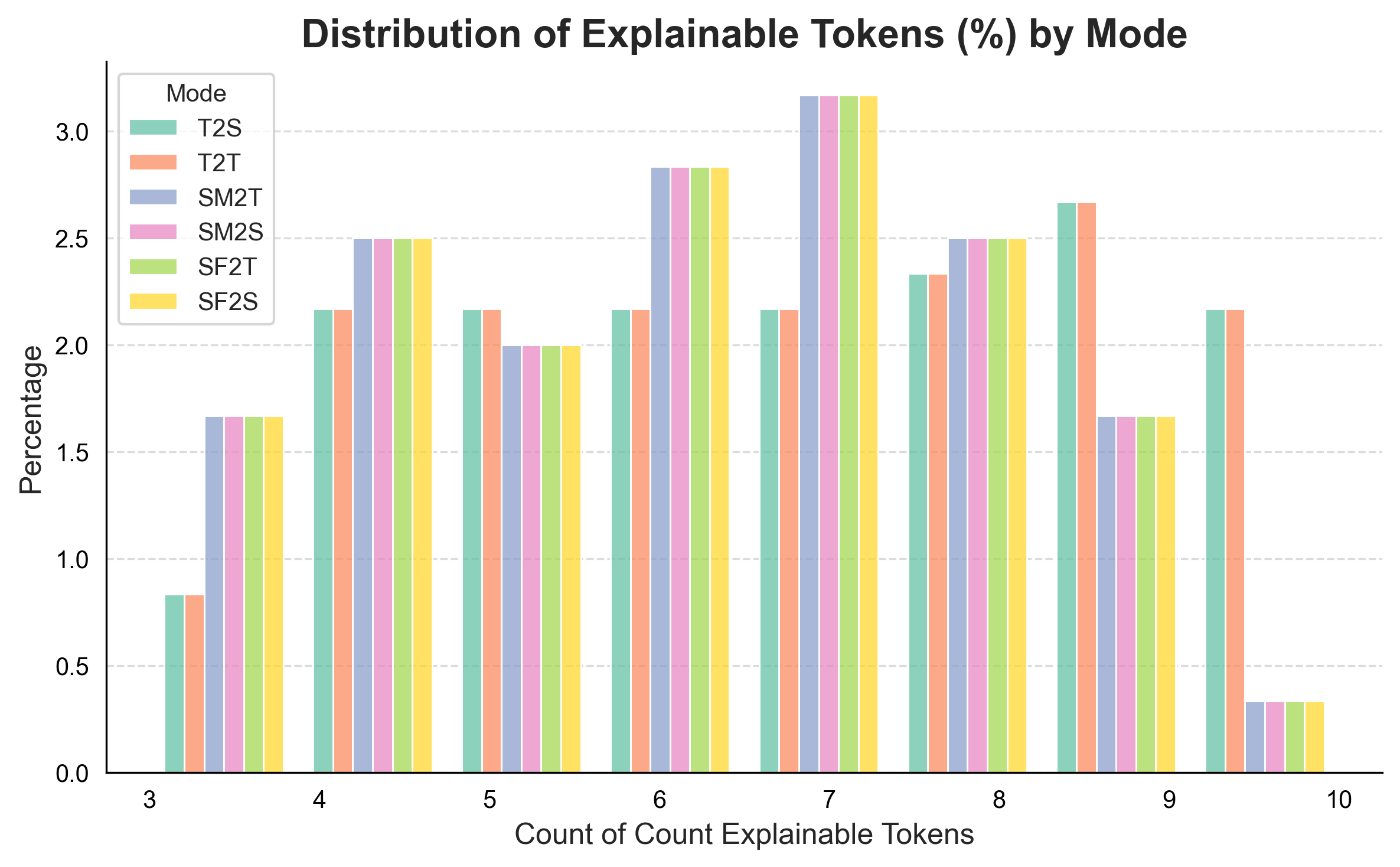}
    \vspace{1em}
    \includegraphics[width=0.75\linewidth]{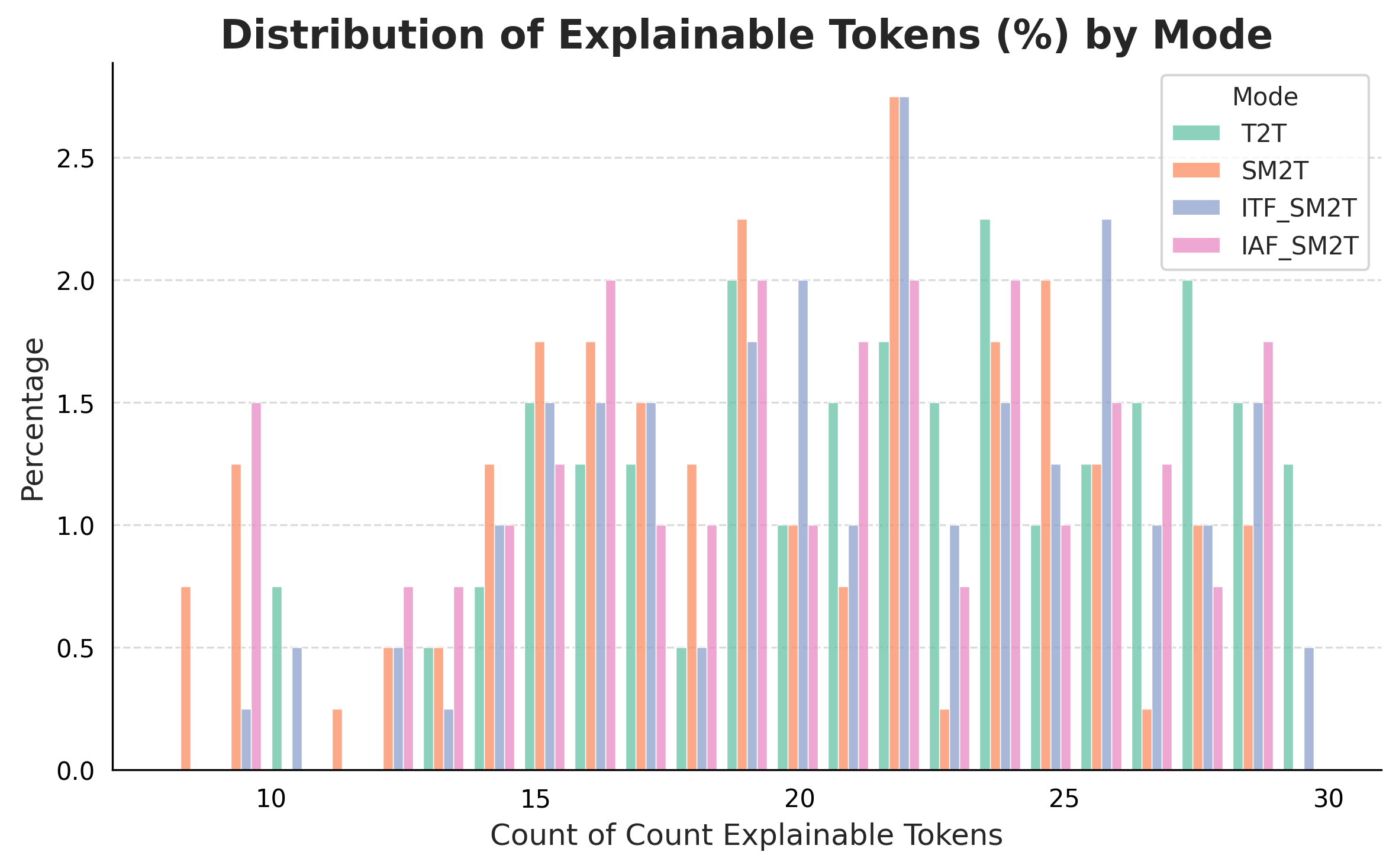}
    \caption{Distribution of explainable tokens per mode. The top chart presents results for \textit{VoiceBench\_\_single\_sentence} and the bottom chart for \textit{VoiceBench\_\_multi\_sentence}.}
    \label{fig:experiments__multi_modal_context__run__token_count_distribution}
\end{figure}

Table~\ref{tab:experiments__multi_modal_context__run__performance_metrics} shows mean duration and number of calls made as well as percentage of observations that reached the second step of Neyman allocation (refer to Section~\ref{subsec:shapley_values__neyman}). We can see that it managed to exit the initialization phase in the majority of cases. In the multi-sentence setting (\textbf{E2}), all evaluated modes consistently reach the second Neyman step and exhibit substantially higher cost (number of calls and runtime), which is expected due to longer, heterogeneous token sequences induced by mixed-modality prompts. Figure~\ref{fig:experiments__multi_modal_context__run__token_count_distribution} illustrates the token distribution across modalities, underscoring that this experiment is confined to short-input sequences only.

\begin{table}[h!]
\centering
\caption{Experiment run statistics including mean runtime performance and proportion of Neyman steps by modality.}
\begin{small}
\begin{tabular}{llcccc}
\hline
& & \multicolumn{2}{c}{\textbf{Performance}} & \multicolumn{2}{c}{\textbf{Neyman Steps}} \\ \cline{3-6} 
\textbf{dataset} & \textbf{mode} & \textbf{n\_calls} & \textbf{runtime (s)} & \textbf{1} & \textbf{2} \\ \hline
\multirow{5}{*}{VoiceBench\_\_single\_sentence} 
    & SF2S & 59.32  & 66.08 & 0.37 & 0.63 \\
    & SF2T & 59.84  & 29.84 & 0.37 & 0.63 \\
    & SM2S & 59.42  & 67.53 & 0.37 & 0.63 \\
    & SM2T & 59.64  & 29.65 & 0.37 & 0.63 \\
    & T2T  & 154.78 & 91.23 & 0.31 & 0.69 \\ \hline
\multirow{4}{*}{VoiceBench\_\_multi\_sentence} 
    & T2T        & 1174.41 & 450.83 & 0.00 & 1.00 \\
    & SM2T       &  924.96 & 364.99 & 0.00 & 1.00 \\
    & ITF\_SM2T  & 1053.04 & 411.49 & 0.00 & 1.00 \\
    & IAF\_SM2T  & 1005.15 & 398.69 & 0.00 & 1.00 \\ \hline
\end{tabular}
\end{small}
\label{tab:experiments__multi_modal_context__run__performance_metrics}
\end{table}

\subsection{Results}
\label{subsec:experiments__multi_modal_context__results}

\paragraph{SV peaks and syntactic centrality.}
No significant correlation was observed between tokens with the highest number of children and those with the highest Shapley values (SV) scores; an intersection occurs in approximately $32\%$ of cases, with negligible variance across different modalities. A similar trend is evident when measuring the alignment between the syntactic root of the input and the highest SV score ($22\%$ on average).
The comparatively poor performance in the T2T mode is likely attributable to its specific tokenization technique, which prioritizes sub-word units over full words more frequently than the speech-based modalities. Detailed scores are provided in Table~\ref{tab:experiments__multi_modal_context__results__combined_analysis}. Low intersection percentages in those cases are not necessarily a negative indicator of SV performance -- this phenomenon might be caused because SV capture more than just the linguistic context.

\begin{table}[h!]
\centering
\caption{Comparison of token importance metrics based on the \textit{VoiceBench\_\_single\_sentence} dataset. The intersection column represents samples where the token with the highest Shapley value (SV) score also has the most children, while Root Match indicates alignment between the highest SV token and the syntactic root. All values are rounded to $2$ meaningful places.}
\begin{tabular}{lcc}
\hline
\textbf{mode} & \textbf{Intersection ratio (SV/Children)} & \textbf{Root Match ratio (SV/Root)} \\ \hline
SF2S & 0.26 & 0.20 \\
SF2T & 0.37 & 0.27 \\
SM2S & 0.35 & 0.24 \\
SM2T & 0.34 & 0.25 \\ \hline
T2T  & 0.29 & 0.14 \\
S2S  & 0.36 & 0.26 \\
S2T  & 0.39 & 0.29 \\ \hline
\textbf{Average (all)} & \textbf{0.32} & \textbf{0.22} \\ \hline
\end{tabular}
\label{tab:experiments__multi_modal_context__results__combined_analysis}
\end{table}

\begin{figure}[t!]
    \centering
    \includegraphics[width=0.75\linewidth]{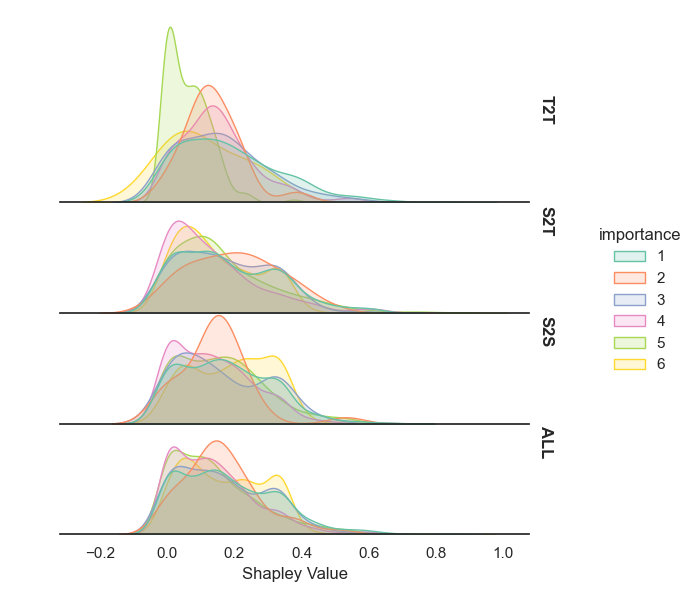}
    
    \vspace{1em}

    \includegraphics[width=0.8\linewidth]{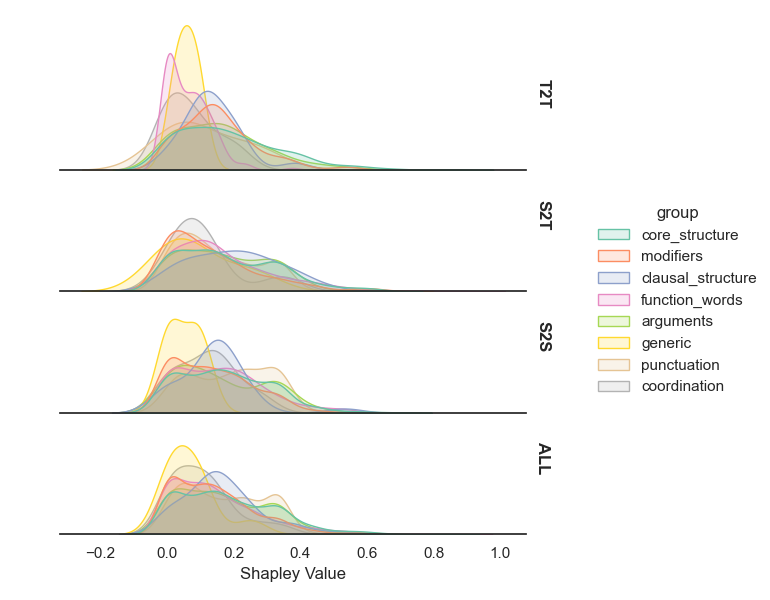}
    
    \caption{Distribution of Shapley values (SV) by mode and importance (Top) or Group (Bottom). While S2S and S2T differs slightly in both cases, more obvious difference can be seen between those modes and the T2T baseline. For each of the charts, scale of y-axis for all modes is the same, they showcase percentages not absolute counts.}
    \label{fig:experiments__multi_modal_context__results__spacy_deps}
\end{figure}

\paragraph{Distributional differences across modalities.}
Distributional analysis of Figure~\ref{fig:experiments__multi_modal_context__results__spacy_deps} shows a contrast between modalities. T2T distributions exhibit significantly lower intra-group variance, particularly in high-importance tiers and the generic category, where density peaks are highly localized. In contrast, S2T and S2S models display heteroscedasticity relative to T2T, with broader, flatter kernels. This increased normalized entropy suggests that speech-input models distribute attribution across a wider range of values.

\paragraph{Significance testing across mode pairs.}
The t-test analysis shows that differences in Shapley values (SV) between the text-only input mode (T2T) and speech-based input modes (S2T, S2S) are consistently significant for the normalized entropy metric, indicating that replacing textual input with speech leads to systematic changes in attribution distributions. In contrast, comparisons based on the Gini coefficient and top-$20$-mass metrics either violated normality assumptions or yielded non-significant $p$-values, suggesting that these measures are less sensitive to differences between text-based and speech-based inputs within the examined sample.

Overall, while SV distributions change markedly when transitioning from purely text input to any speech-involved mode -- as captured by the entropy metric -- the choice of output modality (text vs.\ speech) within the speech-integrated modes does not yield consistently distinguishable effects. Large $p$-values reflect a lack of evidence for statistically significant differences and may arise either from the absence of a true effect or from limited statistical power to detect small effects. Given that significant effects are observed in other comparisons, insufficient power is a plausible explanation in some cases. Detailed results are reported in Table~\ref{tab:experiments__multi_modal_context__results__ttest}.

\begin{table}[t!]
\centering
\caption{Results of paired t-tests comparing Shapley values (SV) across different input–output modes. Entropy values are normalized by the square root of the sample token count. Statistical significance was assessed at $\alpha = 0.05$ using Bonferroni correction for multiple comparisons. Entries marked as \texttt{NaN} correspond to tests for which normality assumptions were not satisfied.}
\begin{tabular}{ll c c c c}
\toprule
\textbf{Metric} & \textbf{Mode\_0} & \textbf{Mode\_1} & \textbf{$p$-value} & \textbf{Significant} & \textbf{Cohen's $d_z$} \\
\midrule
\multirow{6}{*}{entropy} 
 & S2T & S2S & $<0.01$ & True  & -0.49 \\
 & T2S & S2S & $<0.01$ & True  & -0.54 \\
 & T2S & S2T & $<0.01$ & True  & -0.39 \\
 & T2T & S2S & $<0.01$ & True  & -0.65 \\
 & T2T & S2T & $<0.01$ & True  & -0.45 \\
 & T2T & T2S & 0.97    & False & 0.00 \\
\addlinespace
\multirow{6}{*}{Gini} 
 & S2T & S2S & \texttt{NaN} & \texttt{NaN}  & \texttt{NaN} \\
 & T2S & S2S & 0.06    & False & 0.19 \\
 & T2S & S2T & 0.71    & False & -0.04 \\
 & T2T & S2S & $<0.01$ & False & 0.29 \\
 & T2T & S2T & 0.71    & False & 0.04 \\
 & T2T & T2S & \texttt{NaN}    & \texttt{NaN} & \texttt{NaN} \\
\addlinespace
\multirow{6}{*}{top-$20$-mass} 
 & S2T & S2S & \texttt{NaN} & \texttt{NaN}  & \texttt{NaN} \\
 & T2S & S2S & \texttt{NaN} & \texttt{NaN} & \texttt{NaN} \\
 & T2S & S2T & 0.89    & False & 0.01 \\
 & T2T & S2S & $<0.01$ & True  & 0.41 \\
 & T2T & S2T & \texttt{NaN}    & \texttt{NaN} & \texttt{NaN} \\
 & T2T & T2S & \texttt{NaN}    & \texttt{NaN} & \texttt{NaN} \\
\bottomrule
\end{tabular}
\label{tab:experiments__multi_modal_context__results__ttest}
\end{table}

\paragraph{Mixed-modality sequencing effects in E2.}
Using \textit{VoiceBench\_\_multi\_sentence}, we analyze the sparsity of attribution distributions across text-only (T2T), audio-only (SM2T), and interleaved variants (ITF\_SM2T, IAF\_SM2T). Across all evaluated metrics (Gini coefficient, top-$20$-mass, and normalized entropy), most pairwise comparisons either violated normality assumptions or produced non-significant $p$-values. These results provide no evidence for systematic differences between modalities in this setting and may reflect either the absence of a true effect or insufficient statistical power to detect small effects.

\begin{figure}[t!]
    \centering
    \includegraphics[width=0.60\linewidth]{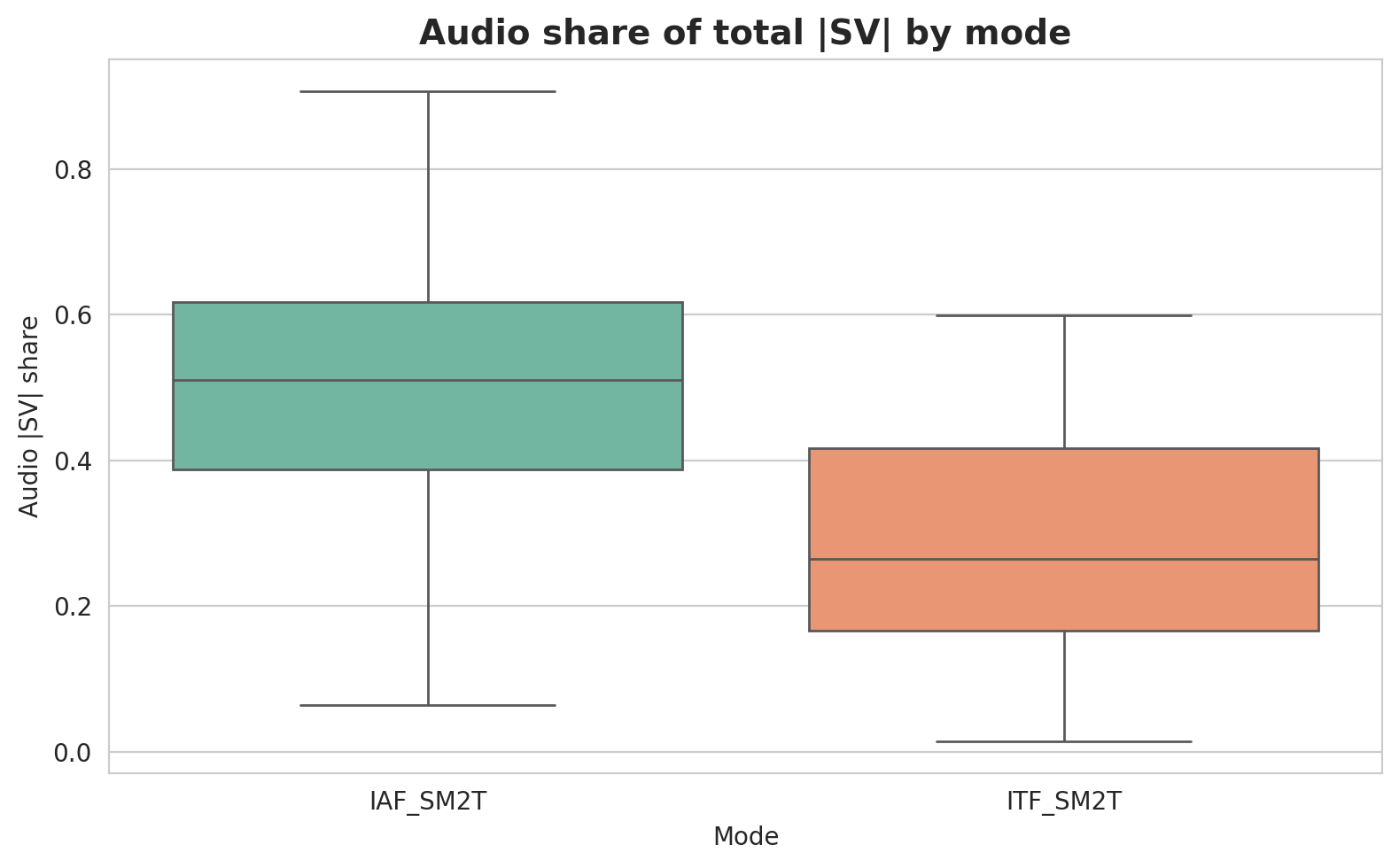}
    
    \vspace{1em}

    \includegraphics[width=0.60\linewidth]{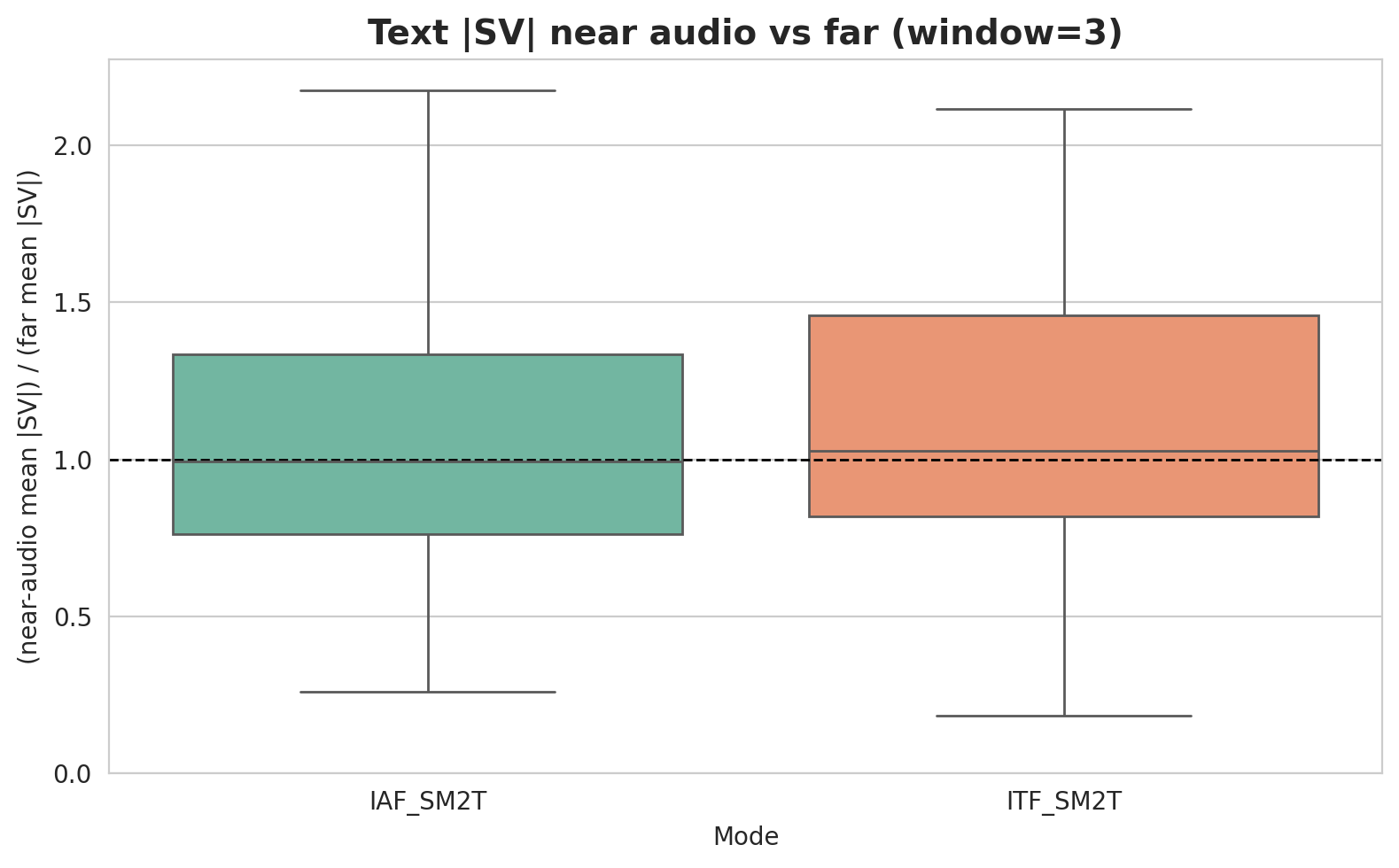}
    
    \caption{E2 diagnostics for \textit{VoiceBench\_\_multi\_sentence}. (Top) Audio share of total attribution mass ($|SV|$) by mode; interleaving patterns differ, indicating modality-order asymmetry. (Bottom) Ratio of mean $|SV|$ on text tokens near audio segments (within a small neighborhood) versus far from audio segments; ratios near $1$ suggest weak but non-zero local cross-modal coupling.}
    \label{fig:experiments__multi_modal_context__results__multi_sentence_audio_text}
\end{figure}

\newpage
\paragraph{Interleaving asymmetry and local cross-modal coupling.}
To ground the interleaving effect in a modality-aligned diagnostic, Figure~\ref{fig:experiments__multi_modal_context__results__multi_sentence_audio_text} reports two simple proxies derived from \textbf{E2} prompts. First, the audio share of total attribution mass differs systematically between interleaving patterns: audio-first interleaving assigns a substantially larger fraction of total $|SV|$ to audio segments than text-first interleaving, suggesting that early modality can act as a stronger attribution anchor under otherwise comparable content. Second, the ratio of mean $|SV|$ on text tokens near audio segments versus text tokens far from audio segments is close to $1$ on average, indicating only a mild proximity effect.

\subsection{Conclusions}
\label{subsec:experiments__multi_modal_context__conclusions}

Our analysis highlights distinct distributional shifts in token attribution across modalities, though we caution that statistical evidence remains preliminary:

\begin{itemize} 

    \item \textbf{Syntactic Independence:} Model attribution mechanisms extend beyond purely linguistic hierarchies, showing limited alignment with syntactic roots or high-degree nodes. 

    \newpage
    \item \textbf{Entropy Shifts:} Speech-integrated inputs (S2T, S2S) yield higher entropy distributions, implying a broader spread of information utilization compared to the localized peaks of text-only (T2T) models. 
    
    \item \textbf{Interleaved Asymmetry:} Mixed-modality scenarios exhibit an anchoring effect where attribution mass favors the initial modality, while local cross-modal coupling appears weak. 
    
    \item \textbf{Statistical Limitations:} Non-significant results in comparative metrics (Gini, top-$20$-mass) prevent definitive conclusions; larger datasets are required to distinguish systematic modality effects from stochastic variance or insufficient power. 
\end{itemize}

\section{Multilingual Context Understanding}
\label{subsec:experiments__multi_language_context}

These experiments analyze how the input language and input modality affect model attribution patterns. We use the \textit{InfinityInstruct\_\_multi\_lingual} dataset (Section~\ref{subsec:experiments__datasets}) and evaluate each prompt in three languages (English -- \texttt{en}, Spanish -- \texttt{es} and French -- \texttt{fr}). For each entry, we run the following scenarios:

\begin{itemize}
    \item \textbf{E3}: \textbf{Text baseline (T2T)} -- text-only input and text output.
    \item \textbf{E4}: \textbf{Speech-conditioned (SF2T, SM2T)} -- audio-only input (female / male speaker) and text output.
\end{itemize}

All runs use the same system prompt and inference pipeline as in previous experiments; we do not introduce any additional language-specific instruction beyond changing the dataset prompt content itself. For reporting convenience, we additionally refer to S2T as the aggregate of the two speech-conditioned variants (SF2T and SM2T).

Section~\ref{subsec:experiments__multi_language_context__run} presents run statistics, Section~\ref{subsec:experiments__multi_language_context__results} reports the results, and Section~\ref{subsec:experiments__multi_language_context__conclusions} summarizes conclusions. Detailed logs and reproduction notebooks for this experiment are available in the project repository\footnote{\url{https://github.com/Pawlo77/MLLM-Shap/blob/feature/multi-sentence-analysis/experiments/analysis/}}.

\subsection{Run statistics and details}
\label{subsec:experiments__multi_language_context__run}

To enable paired comparisons within each language (T2T vs.\ SF2T vs.\ SM2T), we retain only prompts that were successfully processed in all evaluated modes and retain their contextual meaning in all considered languages. This yields a paired subset of $93$ prompts in total (i.e., $31$ per language).

Table~\ref{tab:experiments__multi_language_context__run__performance_metrics} summarizes the mean runtime, number of model calls per observation, and the proportion of observations that reached each Neyman allocation stage (Section~\ref{subsec:shapley_values__neyman}).

\begin{table}[h!]
\centering
\caption{Multilingual experiment statistics including mean runtime performance and proportion of Neyman steps by modality (paired subset, $n=93$ per mode).}
\begin{small}
\begin{tabular}{llcccc}
\hline
& & \multicolumn{2}{c}{\textbf{Performance}} & \multicolumn{2}{c}{\textbf{Neyman Steps}} \\ \cline{3-6}
\textbf{dataset} & \textbf{mode} & \textbf{n\_calls} & \textbf{runtime (s)} & \textbf{1} & \textbf{2} \\ \hline
\multirow{3}{*}{InfinityInstruct\_\_multi\_lingual}
    & SF2T & 234.84 & 105.16 & 0.02 & 0.98 \\
    & SM2T & 235.25 & 104.45 & 0.02 & 0.98 \\
    & T2T  & 605.61 & 204.72 & 0.01 & 0.99 \\ \hline
\end{tabular}
\end{small}
\label{tab:experiments__multi_language_context__run__performance_metrics}
\end{table}

\begin{figure}[h!]
    \centering
    \includegraphics[width=\linewidth]{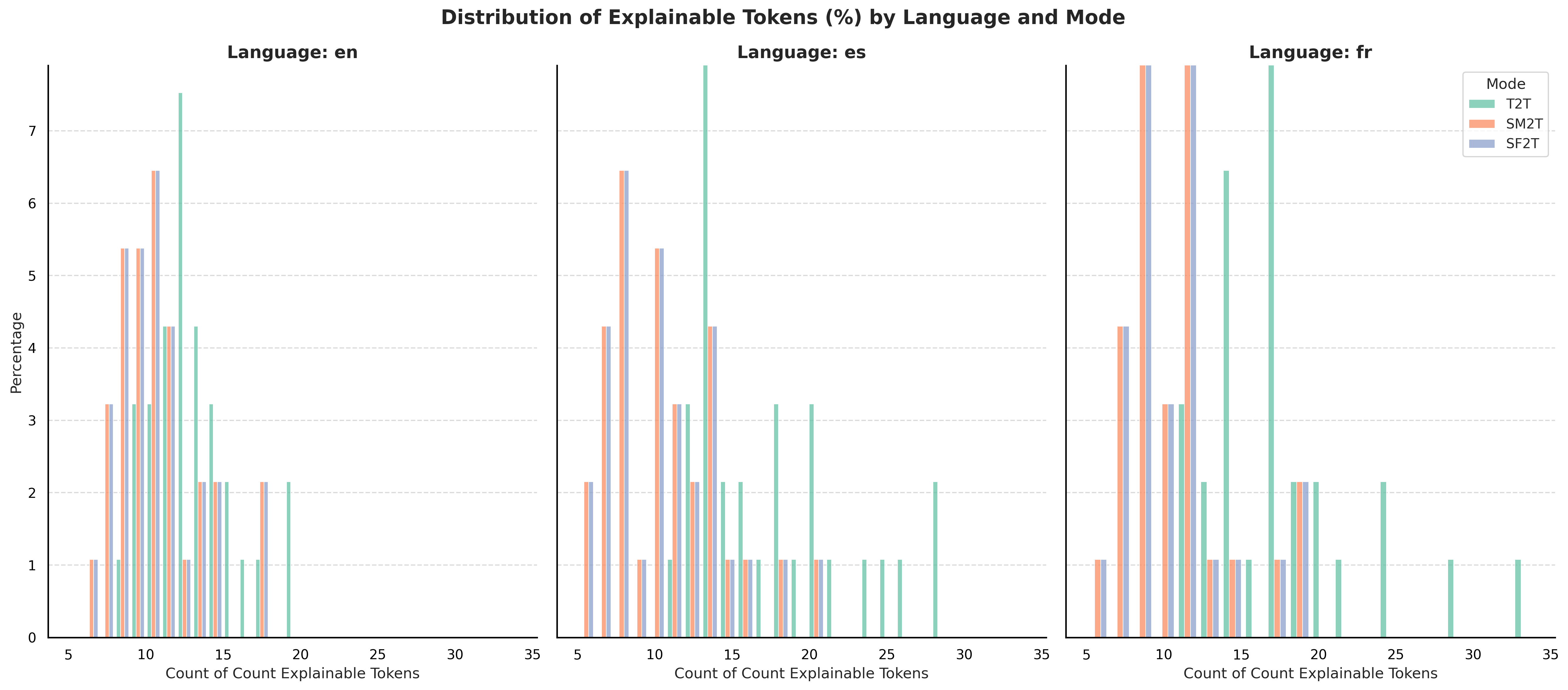}
    \caption{Distribution of explainable tokens (\%) by language and mode for \textit{InfinityInstruct\_\_multi\_lingual}.}
    \label{fig:experiments__multi_language_context__run__token_count_distribution}
\end{figure}

Figure~\ref{fig:experiments__multi_language_context__run__token_count_distribution} shows the distribution of explainable token counts by language and mode. As in the multimodal study (Section~\ref{subsec:experiments__multi_modal_context__run}), differences in token-count distributions motivate normalized comparisons when contrasting attribution trajectories across modalities and languages.

\subsection{Results}
\label{subsec:experiments__multi_language_context__results}

\paragraph{SV peaks rarely coincide with simple syntactic centrality proxies.}
We first test whether the token with the maximum Shapley value (SV) coincides with the token that has the largest dependency fan-out (most syntactic children). Table~\ref{tab:experiments__multi_language_context__results__sv_children_match} reports intersection rates by language and mode. The overall rates remain low, indicating that SV peaks are not well-explained by a single syntactic centrality heuristic (similarly to the multimodal findings in Section~\ref{subsec:experiments__multi_modal_context__results}).

\begin{table}[h!]
\centering
\caption{Shapley values (SV) / children intersection rate (fraction of prompts where the max-SV token also has the maximum dependency fan-out), reported by language and mode. The average is computed over the three base modes (SF2T, SM2T, T2T).}
\begin{tabular}{lccc}
\hline
\textbf{mode} & \textbf{EN} & \textbf{ES} & \textbf{FR} \\ \hline
SF2T & 0.16 & 0.13 & 0.10 \\
SM2T & 0.13 & 0.10 & 0.16 \\
T2T  & 0.19 & 0.10 & 0.06 \\ \hline
\textbf{Average (all)} & \textbf{0.16} & \textbf{0.11} & \textbf{0.11} \\ \hline
\end{tabular}
\label{tab:experiments__multi_language_context__results__sv_children_match}
\end{table}

\begin{figure}[h!]
    \centering
    \includegraphics[width=0.90\linewidth]{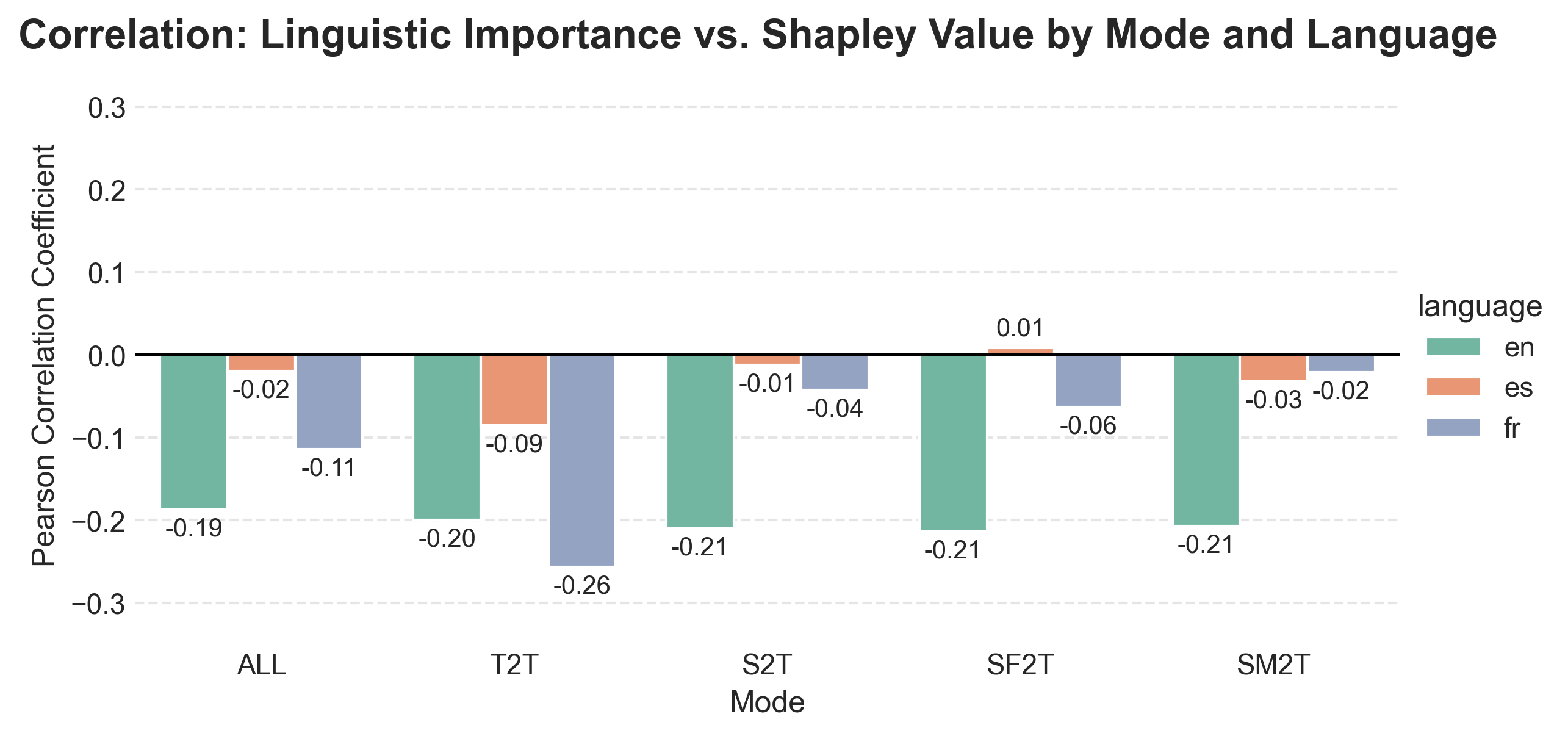}
    \caption{Correlation between linguistic importance and Shapley values (SV) by mode and language.}
    \label{fig:experiments__multi_language_context__results__importance_corr_barplot}
\end{figure}

\paragraph{Language modulates how strongly linguistic importance aligns with attribution magnitude.}
Using the same dependency-to-importance mapping defined in Table~\ref{tab:experiments__multi_modal_context__results__spacy_deps}, we compute the Pearson correlation between the importance index (lower is more linguistically central) and SV. Figure~\ref{fig:experiments__multi_language_context__results__importance_corr_barplot} shows that English exhibits the most consistently negative relationship (indicating that tokens with lower indices, i.e., higher centrality, receive higher SV), whereas Spanish is close to zero (weak alignment), and French shows the strongest effect in T2T.

\begin{figure}[t!]
    \centering
    \includegraphics[width=0.92\linewidth]{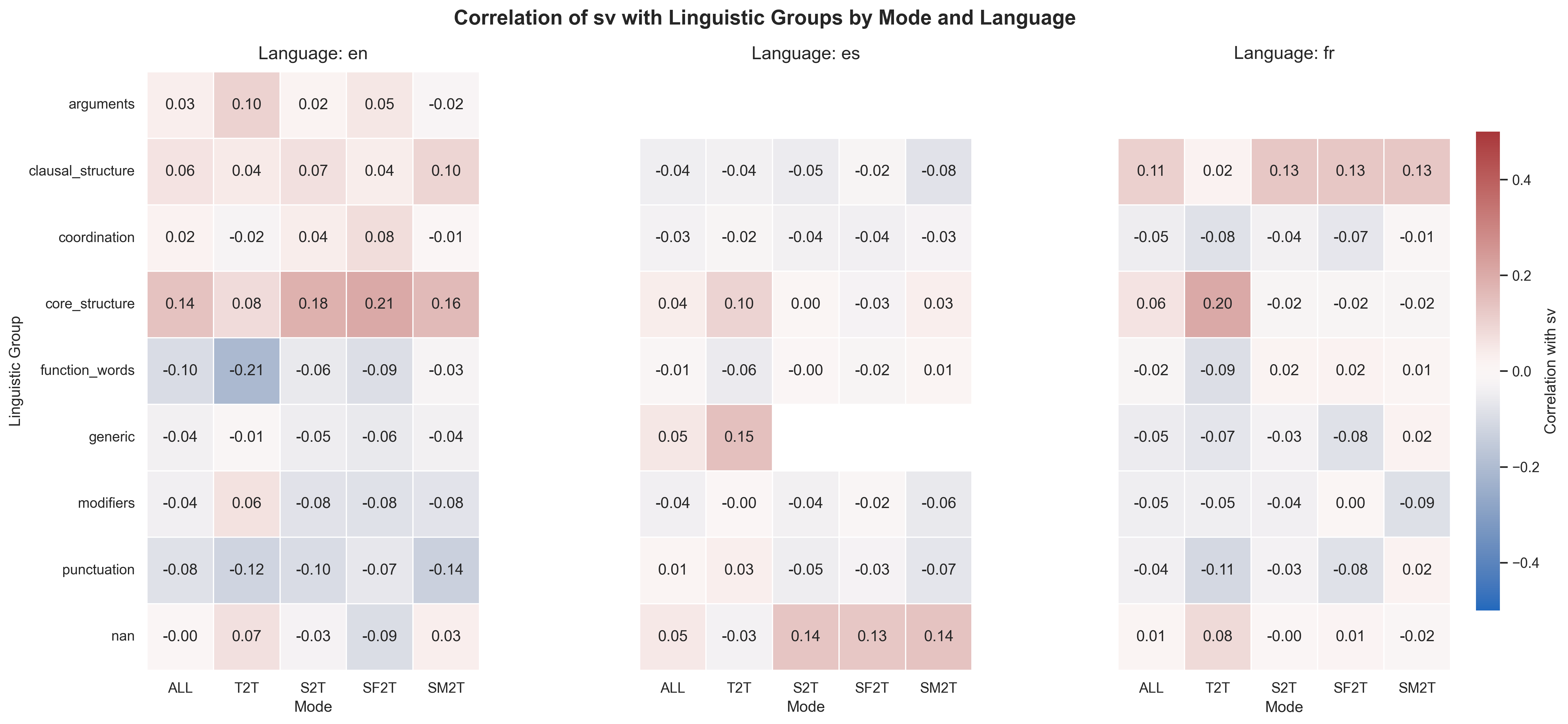}
    \caption{Correlation of Shapley values (SV) with linguistic groups by mode and language (group definitions follow Table~\ref{tab:experiments__multi_modal_context__results__spacy_deps}).}
    \label{fig:experiments__multi_language_context__results__group_corr_heatmap}
\end{figure}

\paragraph{Attribution correlates with linguistic groups in a language-specific way.}
Figure~\ref{fig:experiments__multi_language_context__results__group_corr_heatmap} reports correlations between SV and the linguistic groups (Table~\ref{tab:experiments__multi_modal_context__results__spacy_deps}) across modes and languages. Two qualitative patterns stand out:
(i) English shows a stable positive association of SV with \texttt{core\_structure} and a negative association with \texttt{function\_words} and \texttt{punctuation}, consistent with attributions emphasizing content-bearing tokens;
(ii) French exhibits stronger positive association for \texttt{clausal\_structure} in several modes, suggesting comparatively higher sensitivity to clause-level scaffolding under the same analysis pipeline.
Spanish remains closer to zero in most cells, which is consistent with the weaker importance -- SV alignment seen in Figure~\ref{fig:experiments__multi_language_context__results__importance_corr_barplot}.

\begin{figure}[t!]
    \centering
    \includegraphics[width=0.8\linewidth]{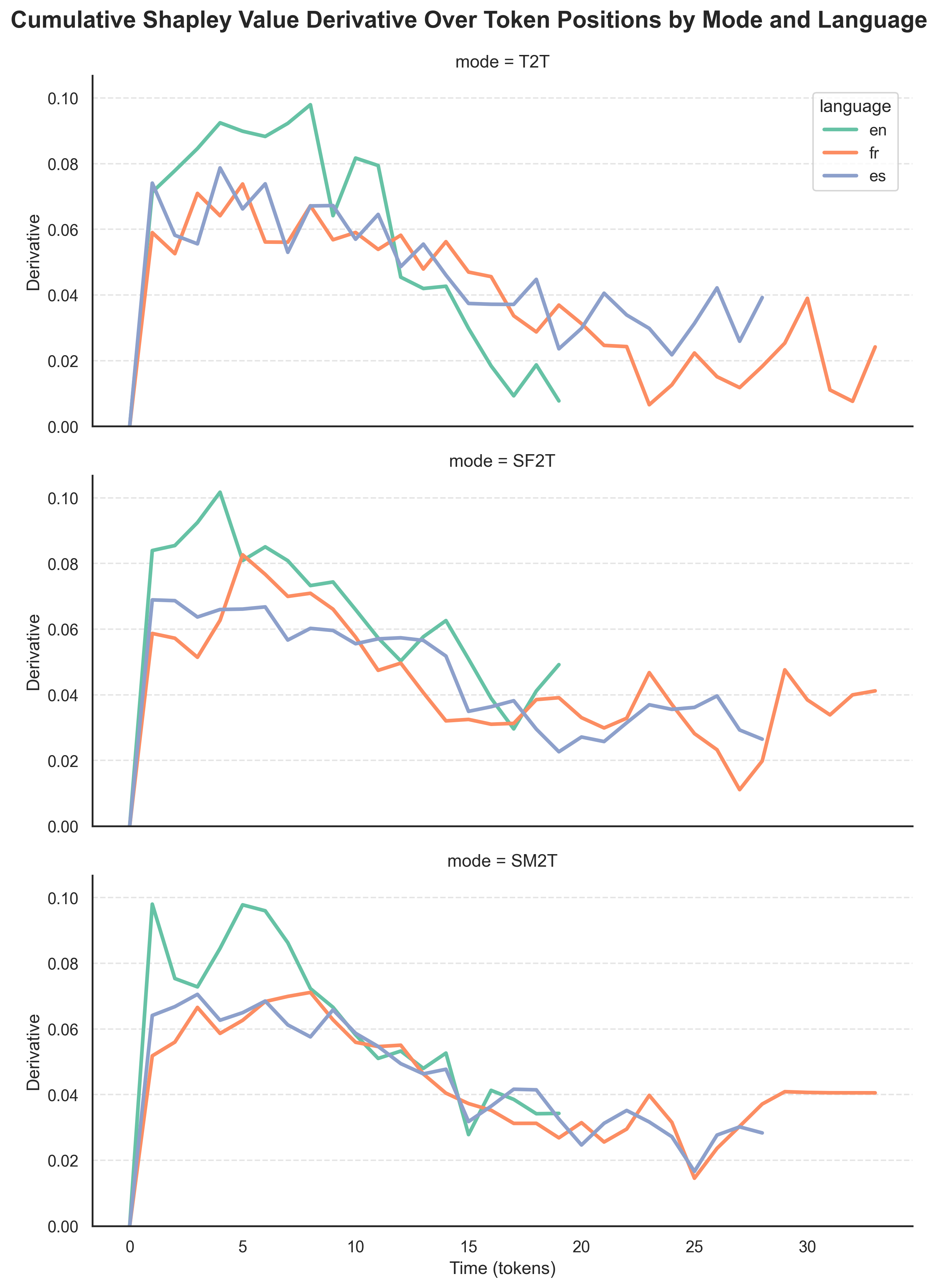}
    \caption{Derivative of interpolated cumulative Shapley values (SV) by language and mode for \textit{InfinityInstruct\_\_multi\_lingual}. The time axis corresponds to token position after normalization/interpolation.}
    \label{fig:experiments__multi_language_context__results__cumsum}
\end{figure}

\paragraph{Normalized attribution trajectories reveal where information is accumulated.}
Figure~\ref{fig:experiments__multi_language_context__results__cumsum} shows derivative-of-cumulative SV trajectories by mode and language (position-interpolated), highlighting how attribution density is distributed across the token sequence despite heterogeneous token counts. Across languages, the derivative is typically largest early and then decreases, suggesting that a substantial portion of ''decision-relevant'' mass is assigned to earlier segments of the prompt.

\begin{figure}[t!]
    \centering
    \includegraphics[width=0.75\linewidth]{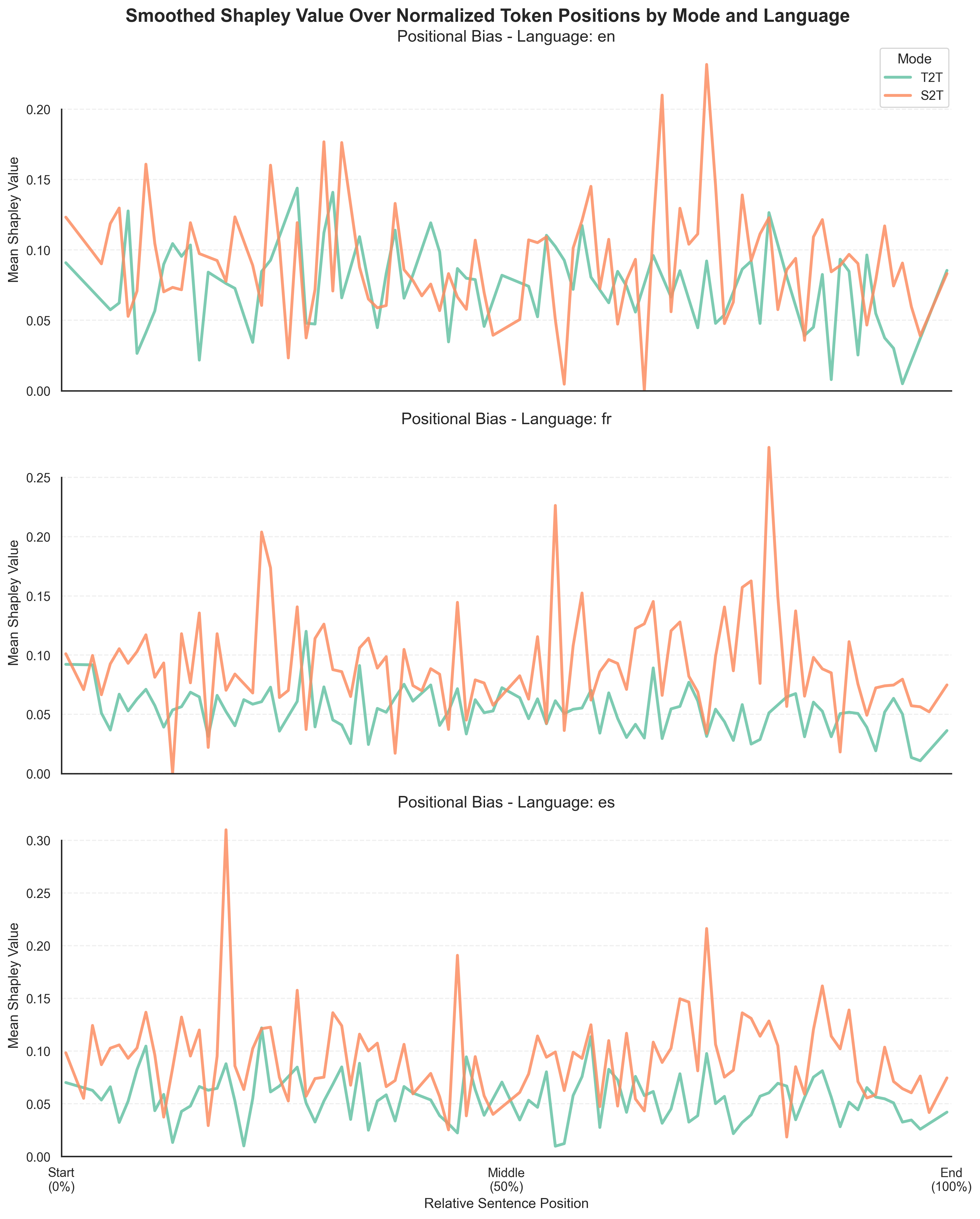}
    \caption{Smoothed Shapley values (SV) over normalized token positions by language and mode (position-normalized analysis).}
    \label{fig:experiments__multi_language_context__results__positional_bias}
\end{figure}

\paragraph{Position-normalized SV indicates higher volatility under speech input rather than a fixed ''early/late'' bias.}
Figure~\ref{fig:experiments__multi_language_context__results__positional_bias} reports smoothed mean SV over normalized token positions. Compared to T2T, S2T is consistently more variable (with occasional spikes) in all three languages, but the spikes do not concentrate in a single narrow region (pure ''start bias'' or ''end bias'').

\paragraph{Statistical context of global metrics.} Across all evaluated metrics (Gini coefficient, top-$20$-mass, and normalized entropy), the tests consistently yielded large $p$-values and small Cohen’s $d$ effect sizes, indicating either the absence of meaningful differences between the compared groups or insufficient statistical power to detect small effects. This suggests that while the \textit{targets} of attribution shift based on linguistic structure (as seen in the correlations in Figure~\ref{fig:experiments__multi_language_context__results__group_corr_heatmap}), the \textit{overall sparsity and concentration} of the model's attention mechanism remain relatively invariant across the tested languages.

\subsection{Conclusions}
\label{subsec:experiments__multi_language_context__conclusions}

Our analysis suggests the following qualitative patterns, though we emphasize that statistical tests across global metrics (Gini, entropy) yielded no significant differences:

\begin{itemize} 
    \item \textbf{Syntactic Independence:} Syntactic centrality (dependency fan-out) is a poor proxy for model attention; the token with the highest linguistic importance rarely coincides with the peak Shapley value. 

    \newpage
    \item \textbf{Language Modulation:} English attributions align most clearly with content and core structure, whereas French shows higher sensitivity to clausal scaffolding, and Spanish exhibits weak alignment with standard linguistic proxies. 
    
    \item \textbf{Modality Volatility:} Speech inputs (SF2T/SM2T) induce higher local attribution volatility compared to text (T2T), suggesting audio conditioning sensitizes the model to specific segments rather than adhering to a smooth positional bias. 
    
    \item \textbf{Early Accumulation:} Attribution density generally peaks early (instruction framing) before decreasing, regardless of the language. 
    
    \item \textbf{Statistical Limitations:} The absence of statistically significant differences suggests these variations may be stochastic or require greater power to distinguish from noise. 
    
\end{itemize}

\chapter{Summary}
\label{sec:summary}

This thesis addressed a practical and currently underexplored problem: explaining \emph{where} Multimodal Large Language Models (MLLMs) place their attribution mass when inputs and outputs span both text and speech. While Shapley values (SV) provide a principled, model-agnostic framework for attributing a model's output to parts of its input, applying SV to audio-conditioned systems is not a direct extension of the text-only setting. Audio introduces substantially longer token sequences, different discretization regimes, and masking operations that can easily become computationally infeasible or methodologically misleading. The core contribution of this work is therefore not only a set of empirical observations about modality and language, but an end-to-end, reproducible methodology that makes SV-based analysis of text–audio MLLMs practically tractable.

From a systems perspective, we implemented an SV computation framework for multi-turn MLLM conversations and integrated multiple estimation strategies to balance fidelity and cost. A central engineering outcome is a reproducible execution pipeline with explicit configuration, pinned data revisions, checkpointed runs, and artifact-based result storage, designed to support long-running sweeps under constrained compute. This infrastructure is complemented by feasibility analyses that quantify the magnitude of the cost problem in the native audio-token regime and motivate the need for principled reductions of the explainable sequence length.

Methodologically, we proposed Spectrogram-Guided Phonetic Alignment (SGPA) to reconcile the mismatch between word-level semantics and dense audio tokenizations. SGPA maps transcripts to acoustically stable, word-aligned segments using Connectionist Temporal Classification (CTC)-based alignment followed by spectrogram-guided boundary refinement, enabling SV games to be defined over interpretable units while reducing coalition complexity by more than an order of magnitude. Importantly, we also measured the consequences of this transformation: SGPA is not attribution-neutral, as it changes the player partition of the cooperative game. However, it converts an otherwise impractical setting into a controllable and auditable approximation layer, making cross-modal comparisons feasible in the first place.

\newpage
Empirically, our experiments on \textit{LFM2-Audio-1.5B} suggest that input modality is a significant driver of attribution volatility and distribution shape. Relative to the text-only baseline, speech-conditioned inputs consistently shift SV distributions and induce more locally volatile attribution profiles, indicating that the model's decision logic may be reweighted when information arrives through audio encoders. In contrast, once the input modality is fixed, the output modality appears to have limited impact on attribution statistics, suggesting that the primary mechanism of change occurs at ingestion rather than at response rendering. In the mixed-modality, multi-sentence setting, the ordering of interleaved audio and text is itself a measurable experimental factor: interleaving patterns yield statistically distinct attribution summaries, implying that multimodal prompt composition is not merely a presentation detail but can influence how the model anchors and propagates contextual cues.

The multilingual experiments further indicate that modality-driven reweighting generalizes beyond English, though linguistic diagnostics are not equally transferable across languages. Correlations between SV and syntactic-importance proxies differ markedly between English, Spanish, and French, and group-level signatures vary accordingly. Across both experimental settings, simple syntactic centrality heuristics (such as dependency fan-out or root alignment) explain only a small fraction of SV peaks, reinforcing that SV highlights model-centric sensitivities reflecting semantics, tokenization effects, and internal representations rather than syntax alone.

These findings, while based on a single model architecture, support a methodological conclusion: SV-based explainability can be extended to text–audio MLLMs in a way that remains both computationally feasible and analytically meaningful, provided that segmentation, normalization, and comparability constraints are treated as first-class methodological concerns. The work provides both a practical framework and preliminary evidence that modality and language shape attribution behavior, with potential implications for the evaluation and design of voice-enabled systems. In particular, our results suggest that comparisons across modes should be interpreted through normalized and position-aware diagnostics, and that mixed-modality prompts merit treatment as distinct experimental conditions rather than as a single ``hybrid'' category. Replication across additional model architectures remains necessary to establish the generality of these patterns.

\chapter{Future Work}
\label{sec:future_work}

This work establishes a methodological foundation for multimodal Shapley value (SV) computation, but compute constraints limited the empirical scope to a single model. Future work naturally divides into four directions: \textbf{(S)} methodological strengthening, \textbf{(E)} expanded and validated experiments, \textbf{(B)} business-oriented applications, and \textbf{(R)} research extensions.

\subsection*{(S) Methodological Strengthening}

\begin{itemize}
\item \textbf{Utility function flexibility:}  
Extend support beyond the utility functions in Section~\ref{subsec:shapley_values__utility_func} to enable task-specific similarity measures, semantic alignment metrics, or domain-adapted importance scoring.

\item \textbf{Granularity control:}  
Implement word-level and phrase-level aggregation as first-class segmentation options alongside the current token-level and Spectrogram-Guided Phonetic Alignment (SGPA) sentence-aligned modes, reducing computational cost while improving human interpretability.

\item \textbf{API integration:}  
Add native support for cloud-hosted and proprietary Multi Modal Large Language Models (MLLM) endpoints, enabling attribution analysis without local model deployment.
\end{itemize}

\subsection*{(E) Experimental Validation and Extension}

\begin{itemize}
\item \textbf{Cross-model replication:}  
Replicate the experimental protocol (Sections~\ref{subsec:experiments__multi_modal_context}--\ref{subsec:experiments__multi_language_context}) on Qwen2.5-Omni and Baichuan-Audio to determine whether observed modality and language effects generalize beyond LFM2 or reflect architecture-specific behavior.

\item \textbf{Human alignment study:}  
Validate SV rankings against human judgments of token importance via controlled annotation experiments, measuring inter-rater agreement and correlation with attribution scores.

\item \textbf{Linguistic analysis deepening:}  
Collaborate with domain experts to interpret language-specific attribution patterns (e.g., why Spanish shows weak SV-importance correlation in Figure~\ref{fig:experiments__multi_language_context__results__importance_corr_barplot}), grounding findings in morphosyntactic theory.
\end{itemize}

\subsection*{(B) Business Applications}

\begin{itemize}
\item \textbf{Prompt engineering toolkit:}  
Develop SV-guided prompt optimization tools that automatically identify redundant tokens, suggest high-impact rephrasing, and quantify sensitivity to input variations -- enabling data-driven prompt refinement beyond manual A/B testing.

\item \textbf{SV-informed RAG:}  
Integrate attribution tracking into retrieval-augmented generation pipelines, where historical SV scores guide dynamic re-ranking of context sources. Systems would learn which retrieved passages consistently contribute high marginal utility, creating self-optimizing retrieval strategies.
\end{itemize}

\subsection*{(R) Research Directions}

\begin{itemize}
\item \textbf{Hierarchical explainer refinement:}  
Investigate why \texttt{HierarchicalExplainer} underperforms (Figure~\ref{fig:methods_comparison_hier}) despite theoretical promise. Test adaptive grouping strategies, importance-weighted sampling schedules, and bias correction at aggregation boundaries.

\item \textbf{Efficient approximations:}  
Explore variance-reduction techniques beyond Neyman allocation, such as control variates, antithetic sampling, or neural surrogate models for coalition evaluation, targeting $10\times$ speedup on sequences exceeding 100 tokens.

\item \textbf{Relational attribution:}  
Extend the SV framework to capture inter-token dependencies explicitly (e.g., via Shapley interaction indices), enabling attribution over token pairs or syntactic constituents rather than assuming feature independence.

\item \textbf{Implementation optimizations:}  
Address known limitations in \texttt{ComplementaryNeyman\allowbreak ShapExplainer} (restricted chat structures) and \texttt{HierarchicalExplainer} (first-level group handling), and reduce redundant utility evaluations through adaptive $m_{\text{init}}$ scheduling per matrix position.
\end{itemize}

\chapter*{Work Separation}
    \label{app:work_separation}
    
    This chapter outlines the division of responsibilities between the two authors. Leveraging their distinct academic backgrounds, tasks were allocated as follows:
    
    \begin{enumerate}
        \item \textbf{Data Science} – comprehensive literature review and identification of business applications; development of the core package for Shapley value (SV) computation, including approximation methodologies and business-oriented functionalities; experimental design; selection of datasets and models; and analysis of experimental results.
        \item \textbf{Computer Science} – development of the Graphical User Interface (GUI); implementation of a distributed workflow for experiment execution; and technical analysis of the obtained results; optimization and verification of the package implementation.
    \end{enumerate}
    
    \noindent Cross-functional tasks, such as repository management, code maintenance, and the drafting of this manuscript, were shared between both authors. Table~\ref{tab:work_separation} presents the detailed separation of work.
    
    \newcolumntype{L}[1]{>{\raggedright\arraybackslash}p{#1}}
    
    \begin{longtable}{| L{2.5cm} | L{6cm} | L{6cm} |}
    \caption{The contribution of the authors of the thesis}
    \label{tab:work_separation} \\
    
    \hline
    \textbf{Role} & 
    \textbf{Paweł Pozorski [327304]} & 
    \textbf{Jakub Muszyński [327246]} \\
    \hline
    \endfirsthead
    
    \multicolumn{3}{c}{\footnotesize\textit{...continued from previous page}} \\
    \hline
    \textbf{Role} & 
    \textbf{Paweł Pozorski [327304]} & 
    \textbf{Jakub Muszyński [327246]} \\
    \hline
    \endhead
    
    \hline
    \multicolumn{3}{r}{\footnotesize\textit{Continued on next page...}} \\
    \endfoot
    
    \hline
    \endlastfoot
    
    
    \textbf{Writing - original draft} & 
    Both Abstracts and Terminology (co-authored), \nameref{sec:introduction}, Work Separation, \nameref{sec:literature}, \nameref{sec:shapley_values} (without \nameref{subsec:shapley_values__comparison}), \nameref{sec:package}, \nameref{subsec:experiments__models}, \nameref{subsec:experiments__datasets} (including \nameref{app:datasets_construction}), \nameref{subsec:experiments__multi_modal_context}, \nameref{sec:experiments_results} (co-authored), \nameref{sec:summary} (co-authored), \nameref{sec:future_work} (co-authored). & Terminology (co-authored), \nameref{sec:business_goal}, \nameref{sec:requirements}, \nameref{app:risk_assessment}, 
    \nameref{subsec:shapley_values__comparison}, \nameref{sec:experiments__sgpa}, \nameref{sec:experiments_setup}, 
    \nameref{subsec:experiments__multi_language_context}, \nameref{sec:gui}, \nameref{sec:experiments_results} (co-authored),
    \nameref{sec:summary} (co-authored), \nameref{sec:future_work} (co-authored) \\ 
    \hline
    
    \textbf{Writing - review \& editing} & 
    Review of sections drafted by co-author. Validation of the Bibliography, Terminology and paper's compliance with formal regulations. Support with section \nameref{subsec:experiments__multi_language_context}. Charts, methodology and conclusions validation. & 
    Review of sections drafted by co-author. Final edits, formatting of the paper (organization), and linguistic corrections. Support with section \nameref{subsec:experiments__multi_modal_context}. Rewrites based on changed scope of the paper. \\ 
    \hline
    
    \textbf{Software} & 
    Core package for Shapley value (SV) computation; Implementation of Exact SV, Monte Carlo, Complementary Contributions (CC), Neyman allocation, and Hierarchical algorithms; Utility functions and token filtering strategies; Caching system for model responses; Audio segmentation using SGPA; Connector for LFM2 model; Unit tests. & 
    Development environment setup; Development of Graphical User Interface (GUI); Implementation of distributed workflow for experiments; Connectors for remaining models; Performance optimizations and system integration; Fixing memory leaks and bottlenecks for cluster deployment. \\ 
    \hline
    
    \textbf{Conceptua-\allowbreak lization} & 
    Idea for multimodal SV implementation; Definition of experiment objectives; Methodology for dataset and model selection. & 
    Strategy for integrating explainability scores into chatbot interfaces; Architecture of the distributed experiment pipeline. \\ 
    \hline
    
    \textbf{Investigation} & 
    Research on SV for Large Language Models (LLMs) and Multimodal LLMs (MLLMs); Investigation of approximation methods for SV; Assessment of hardware requirements and usage costs. Investigation of possible grant to PL-Grid (Appendix~\ref{app:compute_analysis}). & 
    Research on tooling ecosystems for pipelines; Definition of risk assessment protocols; Investigation of Eden Cluster integration; Assessment of compute requirements for experiments; Research on SGPA pre-processing method.\\ 
    \hline
    
    \textbf{Methodology} & 
    Detailed experiment plan (including package tests and linguistic analysis); Approximation methods and computational budget planning. & 
    Experimental protocols; Computational resource estimation methods; Strategies for efficient audio handling. \\ 
    \hline
    
    \textbf{Formal Analysis and Validation} & 
    Analysis of results for single-sentence experiments; Validation of the package using manually created example notebooks; Exploratory Data Analysis (EDA) of datasets; Model performance analysis on benchmarks. & 
    Analysis of results for multi-lingual experiments; Validation of pipeline reliability and GUI responsiveness; Verification and correction of implementation primitives and approaches in the package. \\ 
    \hline
    
    \textbf{Data Curation} & 
    Preparation of VoiceBench and Infinity-Instruct datasets; Selection of target languages and Text-To-Speech (TTS) models. & 
    Management of experiments results, integration of created datasets to experimentation platform. Dataset restriction and modification.\\ 
    \hline
    
    \textbf{Project Administration} & 
    Setup of project repository, code quality checks (pre-commit), and package management systems. Management of core package development. & 
    Establishment of development environment (\texttt{uv}); Management of experiment execution pipeline and GUI development lifecycle; Management of experiments orchestration. \\ 
    \hline
    
    \textbf{Resources} & 
    Management of API keys and credit allocation for external models; Operational support during experiments. & 
    Provisioning of computational resources (Eden Cluster); Allocation of execution time and hardware monitoring. \\ 
    \hline
    
    \textbf{Visualization} & 
    Distributions visualization for datasets EDA and experiments results. & 
    GUI design, visualization of experimental flows and experiments results. \\ 
    \hline
    
    \textbf{Funding Acquisition} & 
    NO & 
    NO \\
    
    \end{longtable}

\chapter*{Terminology}

This chapter defines the key terminology and concepts used throughout this work.

\begin{description}

    \item[A/B Testing] An empirical evaluation method that compares two or more system variants (e.g., prompts, models, or configurations) by exposing them to separate user groups or input samples to measure relative performance.

    \item[Attribution] Quantitative assignment of contribution from input units (e.g., tokens, segments, turns) to a model’s output in a local explanation setting.

    \item[Black-box XAI] Explainability approach that treats a model as an oracle, allowing only input–output access without gradients, attention maps, or internal states.

    \item[Cascaded Architecture] A sequential system structure in which speech input is first transcribed to text (via ASR -- Automatic Speech Recognition), processed linguistically, and then synthesized back into speech (via TTS). This introduces latency and potential information loss.

    \item[Chain-of-Thought (CoT) Prompting] A prompting technique that encourages step-by-step reasoning in LLMs by explicitly eliciting intermediate reasoning steps, improving performance on complex or multi-step problems.

    \item[Coalition] A subset of input features used in SV computation; here, coalitions may consist of text tokens/sentences, audio segments, and dialogue turns.

    \item[Context Window] The maximum effective length of input (tokens or equivalent units) a model can condition on.

    \item[Contextual Understanding] The capacity of a model to incorporate information from previous dialogue turns or multimodal inputs to produce coherent, contextually aligned responses.

    \item[Cosine Similarity] A measure of similarity between two vectors, calculated as the cosine of the angle between them; commonly used to compare semantic similarity between token, sentence, or document embeddings in NLP and model explanation contexts. Refer to Section~\ref{subsec:shapley_values__utility_func} for details.

    \item[Direct Audio-to-Audio Model] A model that processes and generates speech directly in the audio domain -- often via latent audio tokens -- without intermediate text conversion, enabling lower latency and richer paralinguistic representation.

    \item[Embedding] A numerical representation of data (text, image, etc.) in a continuous vector space for model processing.

    \item[Explainable Artificial Intelligence (XAI)] A subfield of AI focused on developing systems whose actions and outputs can be understood by humans, ensuring that model behavior -- such as retrieval ranking or generation logic -- can be audited and interpreted.

    \item[Faithfulness] The degree to which an explanation reflects the true causal influence of inputs on the model’s output (as opposed to merely appearing plausible).

    \item[Few-Shot Prompting] A prompting strategy that provides the model with a limited number of input–output examples within the prompt to induce task-specific behavior without parameter updates.

    \item[Granularity] The unit at which attribution is computed or visualized (e.g., token vs. sentence for text; frame/segment vs. sentence for audio; per-turn for dialogue).

    \item[Large Language Model (LLM)] A transformer-based deep neural network trained on large-scale text corpora to perform natural language understanding and generation tasks such as reasoning, summarization, or dialogue.

    \item[Local vs. Global Explanations] Local explanations target a specific input/output instance (this work); global explanations aim to summarize model behavior over a distribution.

    \item[Mean Absolute Error (MAE)] A metric that quantifies the average magnitude of errors in a set of predictions, without considering their direction. It is calculated as the arithmetic average of the absolute differences between predicted values ($\hat{y}_i$) and observed values ($y_i$):
    \[
    \text{MAE} = \frac{1}{n} \sum_{i=1}^{n} |y_i - \hat{y}_i|
    \]

    \item[Modality] A distinct input or output channel through which information is represented or communicated (e.g., text, audio, image). In this work, modalities refer specifically to text and speech.

    \item[Monte Carlo SV Approximation] A stochastic estimation method for SV using random sampling of feature coalitions, reducing the exponential computational cost of exact computation while maintaining approximation accuracy (conditioned on sample size). Refer to Section~\ref{subsec:shapley_values__monte_carlo} for details.

    \item[Multi-turn Conversation] An interactive dialogue consisting of sequential user–model exchanges where each turn contributes contextual information to subsequent responses, requiring temporal coherence and memory management.

    \item[Multilinguality] Ability to process or generate content in multiple natural languages.

    \item[Multimodal Large Language Model (MLLM)] A model capable of interaction over multiple data modalities (such as text, image, audio, and video), allowing unified reasoning over heterogeneous data.

    \item[Multimodal Reasoning] The ability of a model to integrate and reason jointly over data from different modalities.

    \item[Natural Language Processing (NLP)] A subfield of artificial intelligence focused on enabling machines to understand, interpret, and generate human language. NLP encompasses a range of tasks such as text classification, translation, summarization, question answering, and dialogue generation, combining computational linguistics with machine learning to process natural language data effectively.

    \item[Neyman Allocation] A contribution distribution principle applied to approximate SV by accounting for complementary effects among input features or modalities, offering a cost-efficient explainability approach. Refer to Section~\ref{subsec:shapley_values__neyman} for details.

    \item[Plausibility] The degree to which an explanation appears reasonable to humans; not necessarily equivalent to faithfulness.

    \item[Prompt] The complete input provided to the model in a single inference call, which can include multiple modalities and conversational turns.

    \item[Prompt Engineering] The process of crafting, optimizing, or structuring prompts to guide model behavior, improve output reliability, or induce desired reasoning patterns.

    \item[RAG (Retrieval-Augmented Generation)] A hybrid generation method where an external retriever fetches relevant documents or passages, which are then integrated into the model’s input to enhance factuality and domain coverage.

    \item[Relative Mean Absolute Error (RMAE)] A normalized variation of MAE that expresses the error as a fraction of the magnitude of the actual values, allowing for comparison across datasets with different scales:
    \[
    \text{RMAE} = \frac{\sum_{i=1}^{n} |y_i - \hat{y}_i|}{\sum_{i=1}^{n} |y_i|}
    \]

    \item[Sampling Budget] The number of model evaluations allocated to an approximation method (e.g., number of permutations or stratified samples).

    \item[Sentence-level Alignment] Mapping between text sentences and their corresponding audio spans to enable comparable attribution units across modalities.

    \item[Shapley value (SV)] A cooperative game-theoretic measure that assigns each feature the average marginal contribution to an outcome across all coalitions; used here for local, model-agnostic attribution. Refer to Section~\ref{sec:shapley_values} for details.

    \item[Speech-to-Speech (S2S) System] A speech translation or transformation model converting spoken input directly into spoken output. Early systems employed cascaded pipelines (ASR $\to$ text $\to$ TTS), while modern approaches favor direct audio-to-audio mappings.

    \item[Stratified Sampling] Partitioning features (or permutations) into strata and sampling within each to reduce estimator variance.

    \item[Supervised Fine-Tuning (SFT)] The process of taking a pre-trained machine learning model and adapting it to a specific task using labeled data.

    \item[Text-to-Speech (TTS)] Synthesis used to generate controlled, noise-free audio from textual prompts.

    \item[Token] The smallest unit of input data processed by an LLM (word, subword, or character).

    \item[Transformers] A neural architecture based on self-attention mechanisms, enabling parallelized sequence processing and long-range dependency modeling. It underlies nearly all state-of-the-art LLMs and MLLMs.

    \item[Turn] One side’s contribution in a dialogue (human or model); per-turn attribution aggregates importance over all units within that turn. In this work turns can be also created by the system (system prompts that differ from user messages or steering tokens, interpreted differently by some models).

\end{description}

\newpage
\begin{singlespace}
    \setlength{\bibitemsep}{7pt}
    \printbibliography
\end{singlespace}

\newpage \ \null\thispagestyle{empty}\newpage
\listoffigures

\listoftables

\newpage \ \null\thispagestyle{empty}\newpage

\begin{appendices}
    \listofappendices
    \newpage \ \null\thispagestyle{empty}\newpage

    \chapter{Comprehensive Requirements}
    \label{app:requirements}
    
    This section presents more extensive conditions and expectations for both the research component (Chapter~\ref{sec:experiments}) and the front-facing application (Chapter~\ref{sec:gui}) compared to the brief and narrative description in Chapter~\ref{sec:requirements}. Information is grouped into \textit{constraints} (Section~\ref{subsec:requirements__constraints}) and concrete \textit{functional} as well as \textit{non-functional} requirements (Sections~\ref{subsec:requirements__app} --  \ref{subsec:requirements__research}).
    
    \section{Constraints}
    \label{subsec:requirements__constraints}
    
    Key constraints on the development of this work are discussed in Table~\ref{tab:requirements__constraints__tab}.
    
    \begin{table}[htbp]
    \renewcommand{\arraystretch}{1.1}
    \centering
    \caption{Key constraints governing model evaluation, data handling, and methodological scope. Abbreviations: API (Application Programming Interface), GUI (Graphical User Interface), and TTS (Text-To-Speech).}
    \begin{tabular}{p{2.5cm} p{11.5cm}}
    
    \toprule
    
    \textbf{Constraint} & \textbf{Description} \\
    
    \midrule
    
    Cost and latency & API rate limits and throughput constraints bound the number of feasible model evaluations per session. \\
    
    Hardware & Attributions must complete on commodity workstations; GPU acceleration is optional but not required for correctness. \\
    
    Privacy & No storage of user-provided content without explicit action; session artifacts are local by default. \\
    
    Methodological & Feature units for attribution must be well-defined and consistent across modalities (e.g., text tokens or sentences; audio frames or sentence-aligned segments). \\
    
    Evaluation scope & Datasets are limited to general-purpose, non-specialist domains (Section~\ref{subsec:experiments__datasets}), avoiding heavy mathematical notation or code to reduce confounds in TTS and modeling. \\
    
    \bottomrule
    
    \end{tabular}
    \label{tab:requirements__constraints__tab}
    \end{table}
    
    \section{Functional requirements: front-facing application}
    \label{subsec:requirements__app}
    
    \begin{description}
    
    \item[FR-A1 (Model loading).] Provide a Graphical User Interface (GUI) workflow to select and load models compatible with \texttt{transformers} from local files or public hubs; allow basic configuration (e.g., device, precision).
    
    \item[FR-A2 (Modalities).] Accept text input and audio input (file upload); support sentence-level alignment where available.
    
    \item[FR-A3 (SV methods).] Expose exact Shapley values (SV) for short inputs and sampling-based approximations (e.g., permutation MC and stratified allocation via Neyman method) for longer inputs; surface key hyperparameters (sample budget, granularity, random seed).
    
    \item[FR-A4 (Granularity).] Allow attribution units at token level for text and at segment or sentence level for audio; enable per-turn aggregation for dialogues.
    
    \item[FR-A5 (Visualization).] Render attribution overlays for text (e.g., heat or bar maps) and audio (time-aligned intensity); provide per-turn summaries and modality-comparative views.
    
    \item[FR-A6 (Cost/latency estimation).] Display an estimated number of model evaluations and an approximate time budget based on the chosen method and configuration.
    
    \item[FR-A7 (Session history).] Persist configurations and resulting attributions locally for later inspection and comparability.
    
    \item[FR-A8 (Export).] Enable export of figures and machine-readable attribution data for inclusion in experiments.
    
    \end{description}
    
    \section{Non-functional requirements}
    \label{subsec:requirements__nfr}
    
    \begin{description}
    \item[NFR-1 (Usability).] Configuration and visualization shall be operable without code; defaults shall be sensible and documented. Application Programming Interface (API) should be well documented, include straightforward usage examples, and allow for explanation with minimal coding overhead.
    
    \item[NFR-2 (Performance).] Interface updates shall remain responsive during attribution; long-running computations shall provide progress and be cancellable.
    
    \item[NFR-3 (Reproducibility).] All runs shall log seeds, software versions, and method parameters; reloading a session shall reproduce results given the same model.
    
    \item[NFR-4 (Portability).] The application shall run on a standard Linux environment; no proprietary dependencies are strictly required for core functionality.
    
    \item[NFR-5 (Reliability).] Failures in model calls or audio/text parsing shall be surfaced with actionable messages; partial results shall not be silently discarded.
    
    \item[NFR-6 (Security and privacy).] No external transmission of user inputs or attributions occurs without explicit user action.
    
    \end{description}
    
    \section{Research requirements}
    \label{subsec:requirements__research}
    
    \begin{description}
    
    \item[RR-1 (Black-box setting).] Methods shall not require gradients, attention maps, or internal states; only input--output access is assumed.
    
    \item[RR-2 (Feature units).] Attribution units shall be defined consistently across modalities and documented for each experiment.
    
    \item[RR-3 (Estimators).] Exact SV shall be applied on tractable inputs; sampling-based estimators shall be used otherwise, with configurable budgets and seeds.
    
    \item[RR-4 (Evaluation protocol).] For a subset where exact SV is tractable, approximation quality shall be assessed via absolute error and rank correlation against exact SV; elsewhere, stability across seeds and budgets shall be reported.
    
    \item[RR-5 (Modalities and languages).] Experiments shall cover text-only, audio-only, and mixed-modality inputs, and multilingual conditions as defined in Section~\ref{subsec:experiments__datasets}.
    
    \item[RR-6 (Reporting).] Each experiment shall report: (i) attribution distributions and visualizations, (ii) estimator configuration and evaluation cost (number of model evaluations), and (iii) observed runtime context (e.g., batch size assumptions).
    
    \item[RR-7 (Artifacts).] All raw experimental results must be made publicly available. Any analysis presented should be fully reproducible from this data, utilizing code that is documented and easily accessible to future researchers.

\end{description}

    \newpage \ \null\thispagestyle{empty}\newpage
    \chapter{Inventory of resources}
    \label{app:requirements_inventory}

    This appendix inventories the resources utilized in the experiments described in Chapter \ref{sec:experiments}. Detailed specifications are provided in Table \ref{tab:requirements_inventory}.
    
    \begin{table}[htbp]
    \renewcommand{\arraystretch}{1.1}
    \centering
    \caption{Inventory of resources for experiments and application development. Abbreviations: MLLM (Multimodal Large Language Model), TTS (Text-To-Speech), and GUI (Graphical User Interface). \textit{EN}, \textit{ES}, and \textit{FR} denote English, Spanish, and French, respectively.}
    \begin{tabular}{m{1.8cm} m{4cm} m{8cm}}
    
    \toprule
    
    \textbf{Category} & \textbf{Item} & \textbf{Purpose/Notes} \\
    
    \midrule
    
    Compute     & Linux workstation              & Primary environment for local development, inference, and attribution runs (experiments, application, and package development). \\
    
    Compute     & Consumer-grade Nvidia RTX 4080 & Optional GPU acceleration for experiments and/or application tasks. \\
    
    Software    & Python stack                   & Provides access to \texttt{transformers}, audio I/O, preprocessing, and visualization tools. \\
    
    Datasets    & \textit{VoiceBench} subsets    & English single- and multi-sentence prompts paired with controlled audio. \\
    
    Datasets    & \textit{Infinity Instruct} subsets & Multilingual (EN/ES/FR) multi-turn conversations. \\
    
    Models    & \textit{Kimi Audio} & MLLM used for experiments, state-of-the-art for models under 10B \\
    
    Models    & \textit{Baichuan-Audio} & MLLM used for experiments (origin: China) \\
    
    Models    & \textit{LFM2-Audio-1.5B} & MLLM used for experiments (origin: USA) \\
    
    Audio       & Google TTS engine                     & Generates consistent, noise-free speech for controlled audio inputs. \\
    
    Application & GUI frontend                   & Supports interactive configuration, visualization, and session management. \\
    
    \bottomrule
    
    \end{tabular}
    \label{tab:requirements_inventory}
    \end{table}

    \newpage \ \null\thispagestyle{empty}\newpage
    \chapter{Risk Assessment}
    \label{app:risk_assessment}
    
    The development of a multimodal explainability framework introduces distinct challenges that span theoretical validity, computational feasibility, and software reliability. Unlike standard text-only Large Language Models (LLMs), the integration of the audio modalities and the requirement for multilingual support significantly increase the complexity of the attribution problem. This section provides a comprehensive analysis of these risks, categorizing them into internal methodological constraints and external dependencies.
    
    The risk assessment serves two primary functions. First, it makes explicit the design trade-offs required to adapt the theoretically expensive Shapley value (SV) framework to high-dimensional multimodal data. Second, it establishes a protocol for handling the inherent non-determinism and computational costs associated with sampling-based XAI methods. By identifying these factors early, we define the operational boundaries of the project -- determining which limitations are acceptable constraints and which require active mitigation to ensure the validity of the experimental results.
    
    We begin with a high-level strategic overview using a SWOT (Strengths, Weaknesses, Opportunities, Threats) analysis in Table~\ref{tab:risk_assessment__risk_overview__swot}. This framework highlights the project's unique value proposition -- its model-agnostic and reproducible design -- while acknowledging the unavoidable trade-offs regarding computational latency and the scope of supported modalities.
    
    \begin{table}[t!]
        \centering
        \small
        \caption{SWOT analysis of the project. Abbreviations: GUI (Graphical User Interface), SV (Shapley value), RAG (Retrieval-Augmented Generation). \textit{EN}, \textit{ES}, and \textit{FR} denote English, Spanish, and French, respectively.}
        \label{tab:risk_assessment__risk_overview__swot}
        \begin{tabularx}{\textwidth}{@{}XX@{}}
            \toprule
            \textbf{Strengths} & \textbf{Weaknesses} \\
            \midrule
            Model-agnostic formulation compatible with modern \texttt{transformers} architectures; operates without access to gradients or internal states. & High computational cost and latency on long inputs; exact SV computation is mathematically intractable for coalitions larger than trivial sizes. \\
            \addlinespace
            Unified support for text and audio modalities, enabling multi-turn attribution and sentence-level alignment for cross-modal comparison. & Inherent approximation error and variance in sampling-based estimators, leading to sensitivity to random seeds and sampling budgets. \\
            \addlinespace
            Designed for reproducibility: fixed random seeds, comprehensive config logging, deterministic preprocessing, and persisted session artifacts. & Dependence on specific feature segmentation logic (e.g., tokenizers, silence-based audio segmentation) impacts cross-run comparability. \\
            \addlinespace
            Interactive GUI that bridges research and deployment, offering local session history for complete auditability of experiments. & Scope limited to text and audio modalities (excluding image/video) and a specific subset of languages (EN, ES, FR). \\
            \midrule
            \textbf{Opportunities} & \textbf{Threats} \\
            \midrule
            Growing demand for transparent AI in multimodal systems, aligning with emerging regulatory requirements for model explainability. & External API changes, model deprecations, or licensing updates that could jeopardize data access or reproducibility. \\
            \addlinespace
            Potential to extend the framework to additional tasks (e.g., RAG verification, safety audits) or modalities. & Distribution shifts or hidden biases in the underlying datasets that could lead to misleading attribution results. \\
            \addlinespace
            Community adoption via the open-source package and curated datasets, enabling external validation of the methodology. & Non-determinism across different versions of dependent libraries (e.g., PyTorch, Transformers) reducing result comparability over time. \\
            \bottomrule
        \end{tabularx}
    \end{table}
    
    To quantify the identified risks, we employ a standard assessment matrix evaluating \textit{Likelihood} (L) and \textit{Impact} (I). The criteria for these ratings are defined in Table~\ref{tab:requirements__rating__tab}. This formalization allows us to prioritize mitigation efforts, focusing on risks that threaten the core validity of the thesis (High Impact) or those that are virtually guaranteed to manifest (High Likelihood).
    
    \begin{table}[t!]
        \centering
        \small
        \caption{Risk Assessment Framework: Likelihood and Impact Definitions}
        \label{tab:requirements__rating__tab}
        \begin{tabular}{@{}llll@{}}
            \toprule
            \textbf{Metric} & \textbf{Level} & \textbf{Description} & \textbf{Criteria / Probability} \\
            \midrule
            \multirow{3}{*}{\textbf{Likelihood (L)}}
                & High (H) & Almost certain; a known challenge & $>75\%$ \\
                & Medium (M) & Possible; a known unknown & $25-75\%$ \\
                & Low (L) & Unlikely, but possible & $<25\%$ \\
            \midrule
            \multirow{3}{*}{\textbf{Impact (I)}}
                & High (H) & Invalidates core conclusions & Fundamental methodology change \\
                & Medium (M) & Requires significant rework & Substantially weakens conclusions \\
                & Low (L) & Minor inconvenience & Minor scope limitation \\
            \bottomrule
        \end{tabular}
    \end{table}
    
    Finally, we individually evaluate each technical and operational risk, using a risk register (Table~\ref{tab:risk_register_detailed}). For each entry, we provide the rationale behind the risk rating and the specific technical mitigation strategy planned in the software or methodology.
    
    \clearpage
    {
    \small
    \setlength{\tabcolsep}{5pt}
    \setcounter{LTchunksize}{10}
    \begin{longtable}{@{} p{0.3cm} p{4cm} p{0.2cm} p{0.2cm} p{4.6cm} p{4.6cm} @{}}
        \caption{Project Risk Register and Analysis. SV denotes the Shapley value, API - Application Programming Interface, while EN, ES, and FR stand for English, Spanish, and French, respectively.}
        \label{tab:risk_register_detailed} \\
        
        \toprule
        \textbf{ID} & \textbf{Risk Description} & \textbf{L} & \textbf{I} & \textbf{Rationale} & \textbf{Mitigation} \\
        \midrule
        \endfirsthead
        
        \caption[]{Project Risk Register and Analysis (continued)} \\
        \toprule
        \textbf{ID} & \textbf{Risk Description} & \textbf{L} & \textbf{I} & \textbf{Rationale} & \textbf{Mitigation} \\
        \midrule
        \endhead
        
        \bottomrule
        \multicolumn{6}{r}{\footnotesize\textit{Continued on next page...}} \\
        \endfoot
        
        \bottomrule
        \endlastfoot
    
        R1 & \textbf{Computational Complexity} \newline The computational cost of Shapley values scales as $\mathcal{O}(2^n)$, rendering exact computation intractable for inputs with more than a few dozen features. & H & H & \textbf{L:} This is a fundamental mathematical property of the Shapley method. \textbf{I:} Without effective approximation, the methodology cannot be applied to real-world multimodal inputs, invalidating the project's practical contributions. & We exclusively use sampling-based estimators (Monte Carlo, Stratified Sampling with Neyman allocation) for non-trivial inputs. The system enforces strict computational budget caps, implements request batching, and utilizes result caching to maximize throughput. \\
        \addlinespace
        
        R2 & \textbf{Estimator Variance} \newline Sampling-based approximations introduce stochastic variance, meaning results may fluctuate between runs with different random seeds. & H & M & \textbf{L:} Variance is inherent to all Monte Carlo methods. \textbf{I:} High variance can obscure genuine signal, making it difficult to draw firm conclusions about model behavior or compare different modalities. & We benchmark estimators against exact SVs on small subsets to quantify error. All experimental results report the random seed and sample budget used. We use variance-reduction techniques (stratification) to minimize error for a fixed compute budget. \\
        \addlinespace
        
        R3 & \textbf{Granularity Sensitivity} \newline Attribution scores are highly sensitive to how inputs are segmented (e.g., words vs. tokens, sentences vs. audio chunks). & H & M & \textbf{L:} Multimodal models lack a unified ''atomic'' unit of information. \textbf{I:} Inconsistent segmentation prevents valid comparison between text and audio contributions. & We enforce a strict segmentation policy: text is aligned to sentences or tokens, and audio is segmented by silence or fixed time windows. This policy is logged with every experiment, and conclusions are explicitly scoped to the chosen granularity. \\
        \addlinespace
        
        R4 & \textbf{External Dependency Stability} \newline Reliance on rapidly evolving libraries and external model APIs introduces the risk of breaking changes or result non-determinism. & M & H & \textbf{L:} The AI ecosystem is volatile; updates frequently deprecate features. \textbf{I:} An update could render the codebase non-functional or fundamentally alter model outputs, destroying reproducibility. & We pin all dependency versions in \texttt{requirements.txt} and Docker containers. Where permissible, model weights and tokenizer files are cached locally to insulate the project from remote repository changes. \\
        \addlinespace
        
        R5 & \textbf{Data Quality \& Synthesis Artifacts} \newline The use of Text-to-Speech (TTS) for generating audio inputs may introduce artifacts that the model interprets as signal, leading to spurious attributions. & M & M & \textbf{L:} Synthetic data is never a perfect proxy for natural speech. \textbf{I:} The model might attribute importance to TTS artifacts (e.g., unnatural pauses) rather than semantic content, skewing the analysis. & We use high-quality TTS engines and stratify prompts by difficulty. We employ ''negative controls'' -- randomized or shuffled inputs -- to detect if the model is over-interpreting noise or artifacts in the audio channel. \\
        \addlinespace
        
        R6 & \textbf{Scope \& Generalization} \newline There is a risk that results derived from specific languages (EN, ES, FR) and modalities (Text/Audio) will be inappropriately generalized to other contexts. & H & L & \textbf{L:} The scope is intentionally narrow to ensure depth. \textbf{I:} While external misinterpretation is possible, it does not invalidate the internal logic of the thesis if the scope is clearly communicated. & The thesis explicitly delineates the scope in the Abstract and Introduction. We distinguish between methodological findings (which generalize) and specific behavioral observations (which are specific to the tested models). \\
        \addlinespace
        
        R7 & \textbf{Hardware Constraints} \newline Reliance on limited GPU resources (single commodity card) restricts the scale of experiments and increases the risk of job failures. & H & M & \textbf{L:} Long-running inference jobs are prone to timeouts or memory errors. \textbf{I:} This limits the total number of models and samples that can be analyzed, reducing the statistical power of the broad survey. & The software implements robust error handling, intermediate checkpointing, and graceful degradation. We log hardware specifications with results to provide context for performance metrics. \\
    
    \end{longtable}
    }

\chapter{Package reference}
\label{app:package_reference}

For the scope of this appendix, names starting with an uppercase letter refer to classes, methods within those classes are referred to as \texttt{ClassName:method\_name}, and any lowercase dot-separated strings refer to module paths within the package. \textit{name} refers to parameters class/method from within the scope accepts.

\section{Key objects}
\label{sec:package__func}

\begin{description}
    \item[\texttt{ModelConfig}] (\texttt{connectors.config}) -- default configuration for model inference. 
        \begin{itemize}
            \item \textit{text\_temperature} -- defaults to 0.0;
            \item \textit{text\_top\_k} -- defaults to 1;            
            \item \textit{audio\_temperature} -- defaults to 0.0;
            \item \textit{audio\_top\_k} -- defaults to 1;            
        \end{itemize}

    \item[\texttt{connectors.filters}] – provides utilities for filtering out special phrases during explainability. For example, \texttt{ExcludePunctuationTokensFilter} removes basic punctuation tokens such as ``.'', ``,'' , ``!'', ``?'', ``;'' and ``:''. 

    \item[\texttt{connectors.enums}] Crucial enums for tracking chat history and setting explainability mode:
        \begin{itemize}
            \item \texttt{ModalityFlag} -- flag tracking modality of given token within chat history (\textit{TEXT} / \textit{AUDIO})
            \item \texttt{Role} -- flag indicating the speaker associated with the corresponding token (\textit{USER} / \textit{SYSTEM} / \textit{ASSISTANT})
            \item \texttt{SystemRolesSetup} -- flag indicating what tokens should be explained (\textit{NONE} -- all of them / \textit{SYSTEM} -- all besides one spoken by \textit{Role.SYSTEM} / \textit{SYSTEM\_ASSISTANT} -- just one spoken by \textit{Role.USER})
            \item \texttt{ModelHistoryTrackingMode} -- controls which modality model is expected to generate (\textit{TEXT} / \textit{AUDIO} / \textit{TEXT\_AUDIO})
        \end{itemize}

    \item[\texttt{BaseMllmModel}] (\texttt{connectors.base.model}) -- the base class for all model connectors. It accepts \textit{device} and \textit{history\_tracking\_mode} as constructor parameters. It exposes two crucial public methods:
        \begin{itemize}
            \item \texttt{get\_new\_chat} -- returns a new chat instance associated with this connector.
            \item \texttt{generate} – generates a model response based on the provided \textit{chat}, 
            \textit{max\_new\_tokens}, \textit{model\_config}, and \textit{keep\_history} 
            (defaults to \texttt{False}; when set to \texttt{True}, the generated response is stored in the \textit{chat} instance). Returns \texttt{ModelResponse} instance.
        \end{itemize}
    
    \item[\texttt{ChatEntry}] (\texttt{connectors.base.chat\_entry}) -- represents a single exported entry from the conversation. It provides a \texttt{display} method for a clean, notebook-friendly visualization. Each entry contains the following fields:
    \begin{itemize}
        \item \textit{content\_type} -- a value from the \texttt{ModalityFlag} enum,
        \item \textit{roles} -- a list of speaker roles corresponding to this entry (from the \texttt{Role} enum),
        \item \textit{content} -- encoded conversation tokens (as either \texttt{str} or \texttt{bytes}),
        \item \textit{shap\_values} -- Shapley values (SV) computed for the corresponding tokens (or \texttt{NaN} if the tokens were not included in the explanation).
    \end{itemize}

    \item[\texttt{BaseMllmChat}] (\texttt{connectors.base.chat}) -- the base class representing a chat session. It accepts the following parameters upon creation: \textit{device}, \textit{empty\_turn\_sequences}, \textit{token\_filter}, and \textit{system\_roles\_setup}, where the last two default to \texttt{None}. Its key methods include:
        \begin{itemize}
            \item \texttt{new\_turn} -- creates a new turn in the conversation for a given \textit{speaker} (\texttt{Role}),
            \item \texttt{end\_turn} -- finalizes the most recent turn,
            \item \texttt{add\_text} -- appends a text message to the current turn,
            \item \texttt{add\_audio} -- appends an audio message to the current turn,
            \item \texttt{get\_conversation} -- returns the full conversation as a list of turns; 
                  each turn contains a list of messages, each represented by a \texttt{ChatEntry}.
        \end{itemize}

    \item[\texttt{ModelResponse}] (\texttt{connectors.base.model\_response}) - stores \texttt{chat} with model response added, as well as raw tokens returned by it (\textit{generated\_text\_tokens}, \textit{generated\_audio\_tokens}, \textit{generated\_modality\_flag}).

    \item[\texttt{connectors.liquid}] -- an exemplary connector for the \textbf{LFM2-Audio-1.5B} model (refer to Section~\ref{subsec:experiments__models}). It provides two classes: \texttt{LiquidAudioChat}, inheriting from \texttt{BaseMllmChat}, and \texttt{LiquidAudio}, inheriting from \texttt{BaseMllmModel}.

    \item[Embeddings calculation] The package supports embedding computation using three approaches: static embeddings (where model output tokens are passed through the model's embedding layer, \texttt{BaseMllmModel:get\_static\_embeddings} from \texttt{connectors.base.model}); contextual embeddings (\texttt{BaseMllmModel:get\_contextual\_embeddings} - using the model’s final layer representations); and external embeddings (operating on the detokenized model output) via the interface \texttt{BaseExternalEmbedding} from \texttt{shap.base.embeddings}. Note that the first two methods do not comply with the \textbf{RR-1} (Black-box setting, refer to Appendix~\ref{app:requirements}) requirement.

    \item[Utility functions] Within the package, these are referred to as similarity functions, since they all rely on cosine similarity (see Section~\ref{subsec:shapley_values__utility_func}). \textbf{U1} is implemented using contextual embeddings and \texttt{CosineSimilarity} from \texttt{shap.similarity}; \textbf{U2} supports static and external embeddings using the same similarity class; \textbf{U3} uses \texttt{TfIdfCosineSimilarity} from \texttt{shap.similarity}.

    \item[Embeddings reduction] Since the model returns a sequence of tokens, any embedder produces a sequence of embeddings. As the utility functions require a single vector, these embeddings must be reduced to one. All implementations are available in \texttt{shap.embeddings}: 
        \begin{itemize}
            \item \texttt{MeanReducer}: takes the mean of all vectors along the first dimension. Default;
            \item \texttt{MaxReducer}: takes the maximum;
            \item \texttt{MinReducer}: takes the minimum;
            \item \texttt{SumReducer}: takes the sum;
            \item \texttt{FirstReducer}: returns the first vector, corresponding to the embedding of the first generated token.
        \end{itemize}

    \item[Normalization] Normalization is not required, but SV can be normalized inside the explainer. Three options are available in \texttt{shap.normalizers}: 
        \begin{itemize}
            \item \texttt{AbsSumNormalizer}: $\frac{\overline{SV}}{\sum_i |\overline{SV}_i|}$;
            \item \texttt{PowerShiftNormalizer}: $(\overline{SV} - \min_i \overline{SV}_i)^p$, where $p$ is a hyperparameter (default $1.0$);
            \item \texttt{MinMaxNormalizer}: $\frac{S}{\sum_i S}$, where $S = \frac{\overline{SV} - m}{M - m}$, $M = \max_i \overline{SV}_i$, $m = \min_i \overline{SV}_i$.
        \end{itemize}
    The first two methods return a vector of zeros if that was the input (to avoid division by zero), while the last method returns the original input if $M - m = 0$.

    \item[\texttt{BaseShapApproximation}] (\texttt{shap.base.approx}) - extends \texttt{BaseShapExplainer} and serves as the base class for Shapley value approximation algorithms. It accepts two additional parameters upon creation: \textit{num\_samples} -- the total sampling budget, i.e., the number of coalitions to evaluate; and \textit{fraction} -- the fraction of all possible coalitions to use as the sampling budget.

    \item[\texttt{BaseShapExplainer}] (\texttt{shap.base.shap\_explainer}) -- base class for all explainers. Accepts:
        \begin{itemize}
            \item \textit{mode} -- enum \texttt{Mode} from \texttt{shap.enums}, indicating which embedding calculation to use (contextual or static). Defaults to contextual.
            \item \textit{embedding\_model} -- instance of a class calculating external embeddings. Overrides \textit{mode}. Defaults to \texttt{None}.
            \item \textit{embedding\_reducer} -- instance of a class responsible for reducing embeddings. Defaults to \texttt{None}.
            \item \textit{similarity\_measure} -- instance of a class responsible for calculating the utility function. Defaults to \texttt{CosineSimilarity}.
            \item \textit{normalizer} -- instance of a class responsible for SHAP value normalization. If no normalization is desired, \texttt{IdentityNormalizer} is available in the same location as the rest. Defaults to \texttt{PowerShiftNormalizer}.
            \item \textit{allow\_mask\_duplicates} -- whether to force unique mask generation. Defaults to \texttt{True}.
        \end{itemize}
    It is a callable class that accepts as input:
        \begin{itemize}
            \item \textit{model} -- instance of the model for inference;
            \item \textit{source\_chat} -- instance of chat without the model response turn;
            \item \textit{response} -- model response to explain;
            \item \textit{progress\_bar} -- boolean indicating whether to display a progress bar. Defaults to \texttt{True}.
            \item \textit{verbose} -- whether to save \textit{history}. Defaults to \texttt{False}, as it might be space-consuming;
            \item \textit{n\_generator\_jobs} -- number of threads for parallel model inference. Each thread will receive the same instance of the model.
        \end{itemize}

    \item[\texttt{Explainer}] (\texttt{shap.compact}) -- compact interface that accepts:
        \begin{itemize}
            \item \textit{shap\_explainer} -- an instance of an explainer class that inherits from \texttt{BaseShapExplainer} (provided upon creation);
            \item \textit{chat} -- an instance of the chat to be explained (provided on call);
            \item \textit{generation\_kwargs} -- optional parameters used to configure model inference (provided on call).
        \end{itemize}

    Returns an instance of \texttt{ExplainerResult} (\texttt{shap.explainer\_result}) with the following attributes:
        \begin{itemize}
            \item \textit{full\_chat} -- an instance of the input chat filled with cache, calculated SV, and model responses;
            \item \textit{source\_chat} -- the original input chat instance;
            \item \textit{history} -- estimator history, if applicable; it is a list of tuples \texttt{(mask, mask\_hash, masked\_chat, model\_response)}, where \texttt{masked\_chat} is a copy of \texttt{input\_chat} with \texttt{mask} applied, and \texttt{model\_response} is the output generated from that chat. Positions in \texttt{masked\_chat} can be \texttt{None}, indicating the entry was retrieved from cache;
            \item \textit{total\_n\_calls} -- total number of calls made to the utility function (not the model). This includes entries from cache and represents the number of considered coalitions.
        \end{itemize}

    \item[\texttt{PreciseShapExplainer}] (\texttt{shap.precise}) -- implements precise SV formula from Section~\ref{sec:shapley_values}. 

    \item[MC algorithms] -- proposed in Section~\ref{subsec:shapley_values__monte_carlo}. Implemented by \texttt{LimitedMcShapExplainer} and \texttt{StandardMcShapExplainer} from (\texttt{shap.monte\_carlo}), where the former uses first-order omission coalitions, in contrast to the latter.

    \item[\texttt{ComplementaryNeymanShapExplainer}] (\texttt{shap.neyman}) implements the Neyman approximation formula from Section~\ref{subsec:shapley_values__neyman}. It accepts additional parameters \textit{initial\_num\_samples} and \textit{initial\_fraction}, which are calculated similarly to the non-initial ones in \texttt{BaseShapApproximation}. If neither parameter is provided, the default formula $\max\big(2, \left\lfloor \frac{\text{num\_samples}}{2 n^2} \right\rfloor \big)$ is used. This implementation does not support \textit{allow\_mask\_duplicates} nor \textit{n\_generator\_jobs}; they are fixed to True and $1$ respectively. It differs from the theoretical formula by generating masks of fixed size with a fixed player included during the initial step, as in practice, without this constraint, a much larger initial sampling budget was required to fulfill the criterion $M \ge m_{\text{init}}$. Original behavior can be forced by using \texttt{StandardComplementaryNeymanShapExplainer}. If the entire budget is spent during the initial phase, the explainer falls back to the behavior of \texttt{ComplementaryShapExplainer}.

    \item[\texttt{ComplementaryShapExplainer}] (\texttt{shap.complementary}) - implements Complementary Contributions (CC) formula from Section~\ref{subsec:shapley_values__complementary}. Same as in MC algorithms, it has limited formula where first $2n$ masks $S_1, N \setminus S_1, S_2, N \setminus S_2, \dots$ are drawn as follows: $S_i$ -- mask of $\lfloor n / 3 \rfloor$ players, where player $z_i$ is always present. Original behavior can be forced by using \texttt{StandardComplementaryShapExplainer}.
    
    \item[\texttt{HierarchicalExplainer}] (\texttt{shap.hierarchical.explainer}) - implements the algorithm described in Section~\ref{subsec:shapley_values__hierarchical}. In addition to the standard input accepted by \texttt{Explainer}, it supports the following parameters upon creation:
        \begin{itemize}
            \item \textit{mode} -- an enum value from \texttt{Mode} (\texttt{shap.hierarchical.mode}). It implements Algorithm~\ref{alg:shapley_values__hierarchical__alg} directly only in the \texttt{TEXT} mode. Otherwise, the number of groups on the first level may be larger - they are formed with respect to modalities (\texttt{MULTI\_MODAL}) or both modalities and roles (\texttt{MULTI\_MODAL\_MULTI\_USER}). From a practical standpoint, it is not meaningful to merge turns or modalities within a single group, as they represent different contexts. Therefore, the third mode is provided, while the second mode is the default for multimodal chats.
            \item \textit{k} -- the $k_{\text{max}}$ parameter from the algorithm.
            \item \textit{use\_importance\_sampling} - a boolean flag. If set to \texttt{True}, group importance (calculated top-down, as illustrated in Figure~\ref{fig:shapley_values__hierarchical__example}) will be used as the \textit{fraction} value for the approximation algorithm. It is strongly recommended to use this option together with \texttt{MinMaxNormalizer}. Defaults to \texttt{False}.
            \item \textit{importance\_sampling\_min\_fraction} -- used when importance sampling is enabled, ensuring that the sampling fraction does not become too small.
            \item \textit{first\_layer\_explainer} (\texttt{BaseShapExplainer}) -- if provided, this explainer is used to approximate the first-level group weights by summing the Shapley values of the tokens within each group.
        \end{itemize}

    \item[\texttt{SpectrogramGuidedAligner}] (\texttt{connectors.base.audio}) -- the core implementation of the SGPA algorithm. It performs word-level alignment by combining CTC-based character paths from \texttt{Wav2Vec2} with local spectral cue refinement (energy and flux) to identify acoustically stable boundaries.

    \item[\texttt{AudioSegment}] (\texttt{connectors.base.audio}) -- represents a word-aligned audio unit created by the aligner. It stores timing metadata, sample indices, and (optionally) raw bytes, serving as the discrete ``player'' for audio modality in Shapley value games.
\end{description}

\section{Alignment with requirements}
\label{sec:package__align}

A detailed description of each requirement is available in Appendix~\ref{app:requirements}. Sample output of the package is presented in Table~\ref{tab:package__align__example}.

\begin{description}
    \item[FR-A2: Modalities] The \texttt{connector} enables communication between explainers and multimodal models, supporting both text and audio for inputs and outputs.

    \item[NFR-1: Usability] -- As shown in Section~\ref{sec:package__docs}, the package provides comprehensive, industry-standard documentation. It includes usage examples, release notes, and detailed code references with easy access to the source code. Its straightforward and modular structure exposes key classes at the top level, allowing users to start with just a few lines of code. Each functionality requires minimal setup -- only a chat history must be created, after which any chosen explainer can be applied -- enabling effortless integration into existing projects. The package is designed in way allowing for seamless development of new connectors (interaction with new models) and supports popular \texttt{transformers} interface.
    
    \item[NFR-2: Performance] – The package operates on masks, introducing negligible overhead. With multiple approximation algorithms available (see Section~\ref{sec:package__func}), users can select the most efficient Shapley values (SV) approximation method according to their specific performance or business requirements. All operations can be performed on GPU.
    
    \item[NFR-3: Reproducibility] -- By default, all supported models perform generation using parameters defined in the \texttt{ModelConfig} class. This enforces deterministic behavior by setting the \textit{temperature} to~$0$ and \textit{top\_k} to~$1$ for all applicable modalities, ensuring reproducible results.
    
    \item[NFR-4: Portability] -- The package is installable via \texttt{pip} on all major platforms. It relies solely on the \texttt{torch} library~\parencite{paszke2019pytorch}, which supports all major operating systems and GPU providers, ensuring full cross-platform compatibility.
    
    \item[NFR-5: Reliability] -- All methods validate their input data and internal assumptions to ensure correctness. If a validation step fails, an informative error message is raised. Additionally, the package includes configurable logging functionality at multiple levels, enabling transparent tracing and debugging.
    
    \item[NFR-6: Security and Privacy] -- All interactions with external models occur only through explicit user action. No data is stored or transmitted beyond the local runtime memory, ensuring that user data remains private and secure.

    \item[RR-1: Black-box setting] Integration with the model is handled exclusively through the connector class, which is responsible only for translating model inputs and outputs.
    
    \item[RR-2: Feature units] The package tracks all input units to ensure consistency, including their modalities and the roles responsible for them. This information is used to create informed splits without breaking their logical structure. Each output provides access to the \texttt{response\_chat} object, enabling tracking of these data, estimation history, and final results.
    
    \item[RR-3: Estimators] The package provides approximation algorithms designed to efficiently handle long input sequences.
\end{description}

\section{User-side documentation}
\label{sec:package__docs}

The official documentation is available on GitHub Pages\footnote{\url{https://pawlo77.github.io/MLLM-Shap/}}. It is automatically generated from Python documentation strings (docstrings) using \texttt{Sphinx} \parencite{sphinx}, supplemented with custom content written in \texttt{.rst} files. The documentation provides usage examples (with links to additional notebooks in the main repository), detailed descriptions of key features and modules, a list of supported models and interfaces, installation instructions, release notes, and a guide to the Graphical User Interface (GUI) client. It also includes selected results from Section~\ref{subsec:shapley_values__comparison} and comprehensive package interface documentation, enabling users to easily explore the source code.

\begin{figure}[t!]
    \centering
    \includegraphics[width=0.75\linewidth]{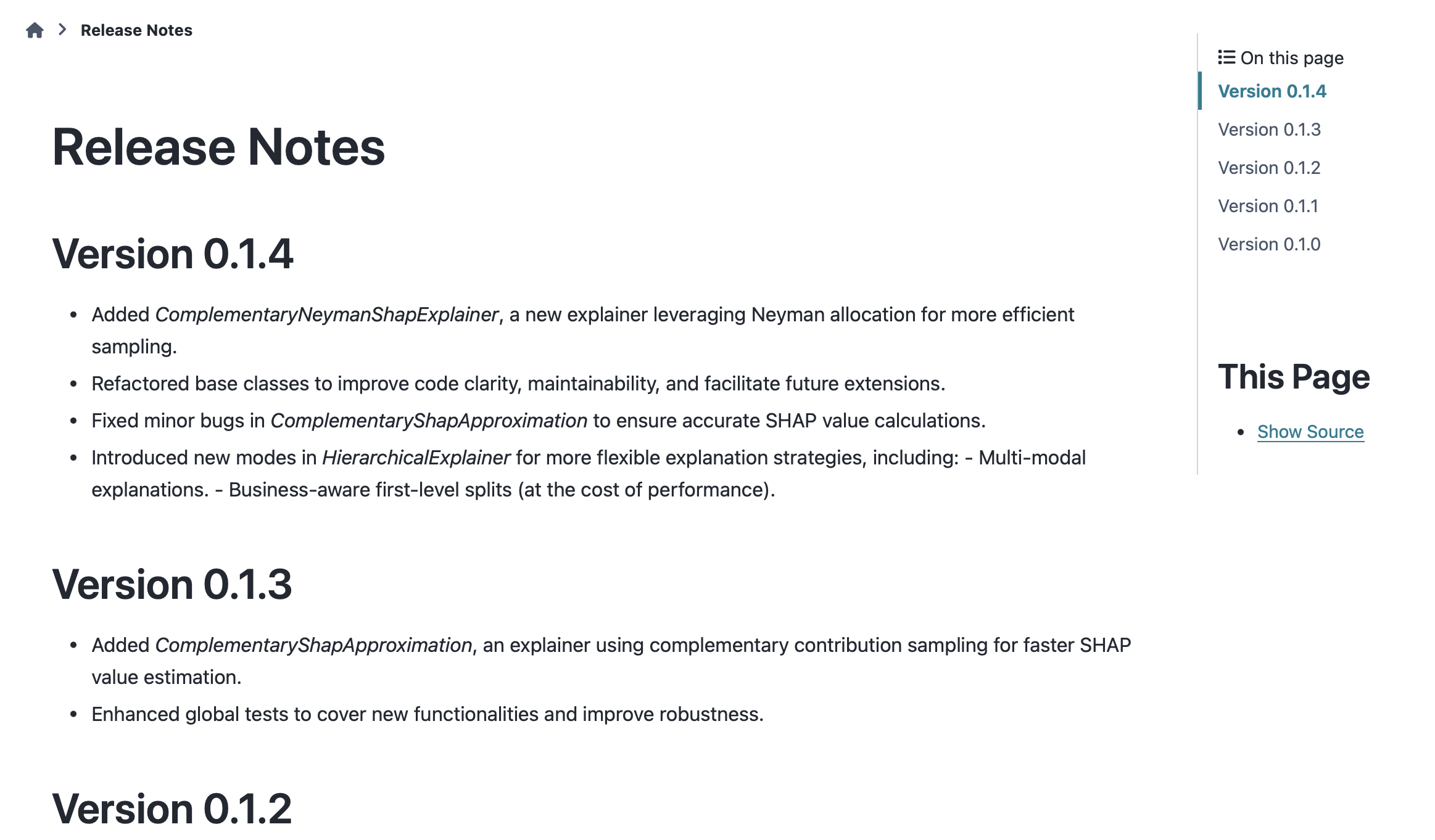}
    \caption{Release notes, allowing users to track introduced changes.}
    \label{fig:package__docs__release_notes}
\end{figure}

\begin{figure}[t!]
    \centering
    \includegraphics[width=0.8\linewidth]{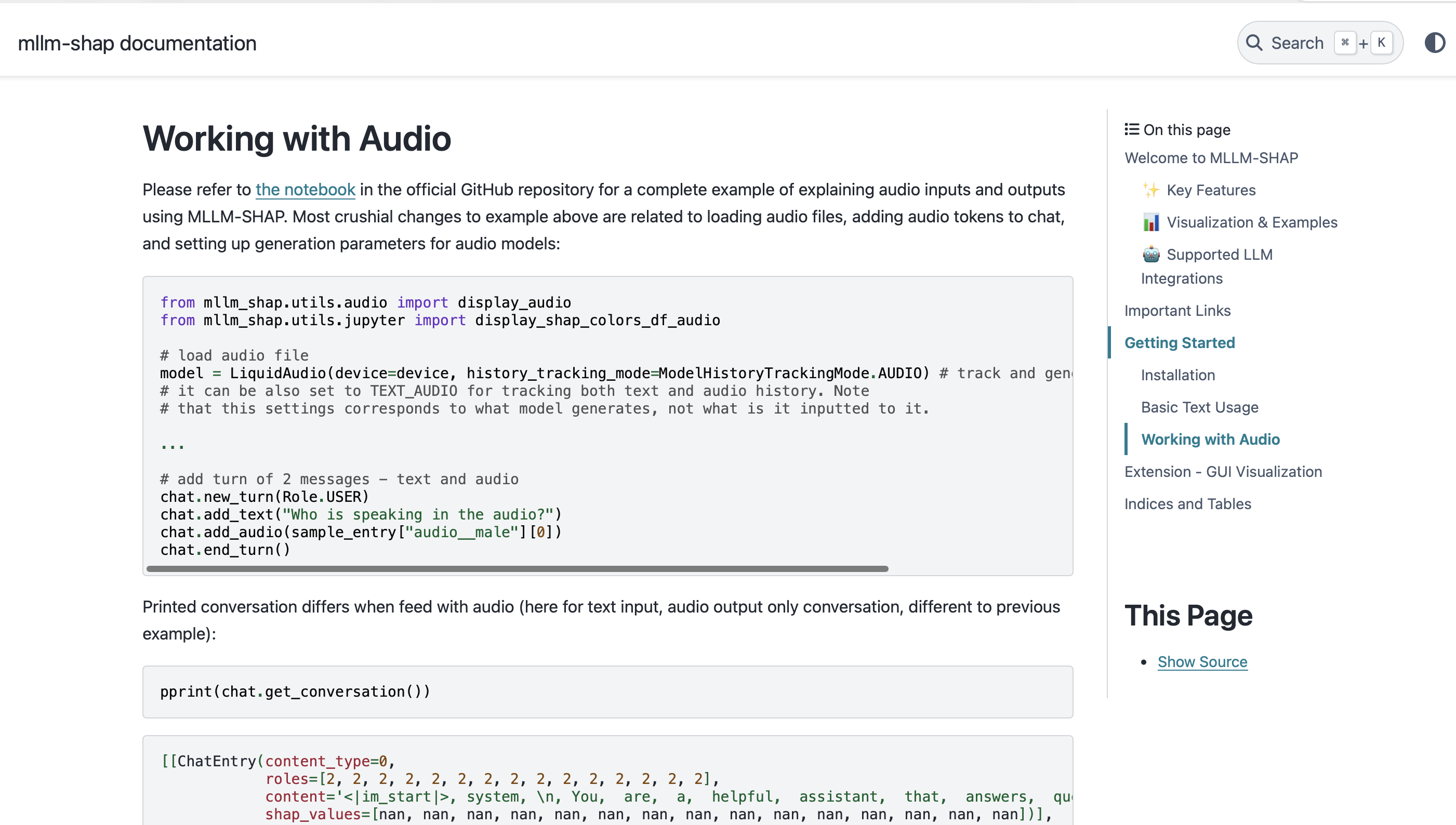}
    \caption{The documentation page includes illustrative examples with code, explanatory comments, and expected outputs, helping users quickly and clearly understand how to use the API.}
    \label{fig:fig:package__docs__examples}
\end{figure}

\begin{figure}[t!]
    \centering
    \adjustbox{valign=t}{\includegraphics[width=0.49\linewidth]{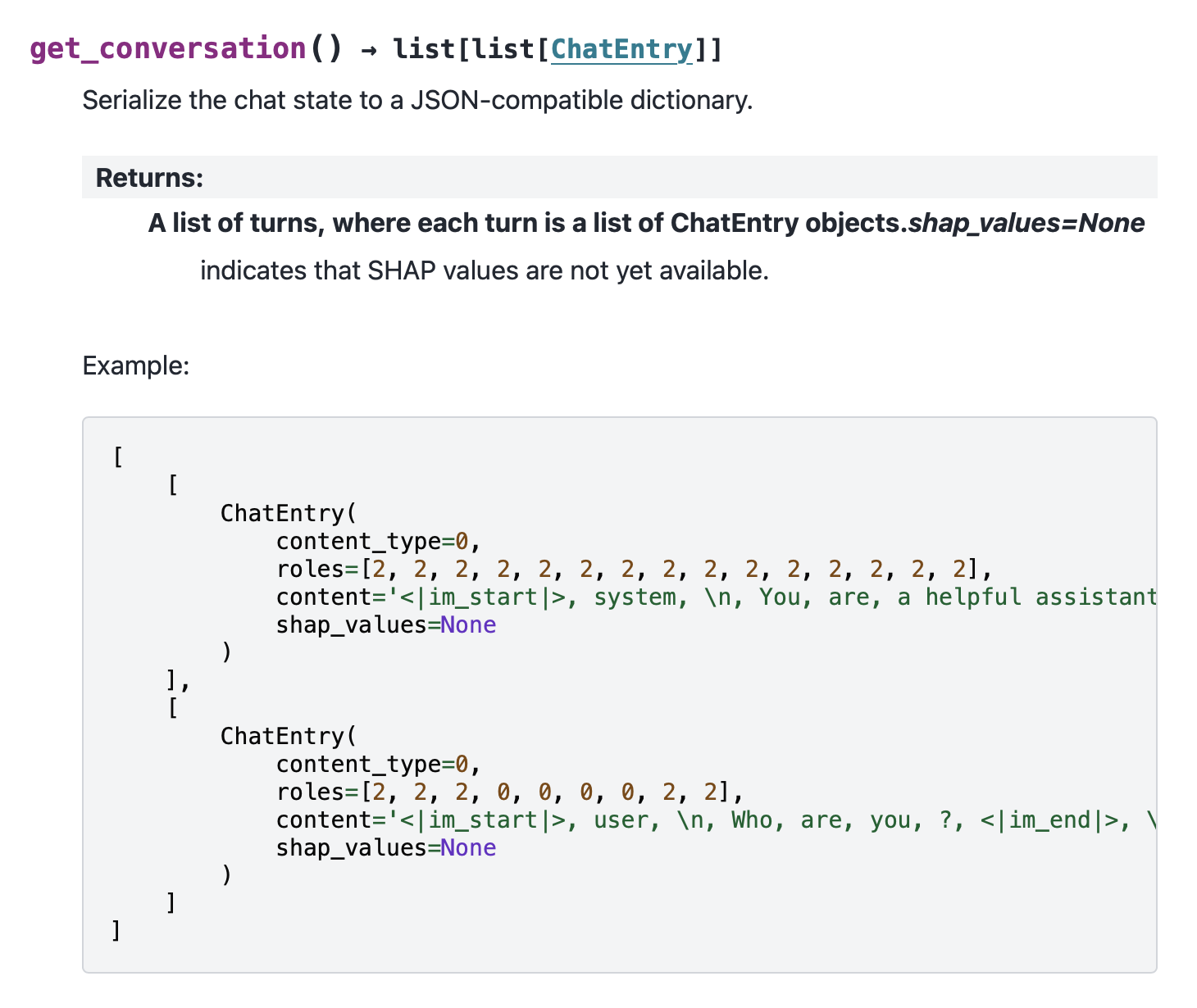}}
    \hfill
    \adjustbox{valign=t}{\includegraphics[width=0.49\linewidth]{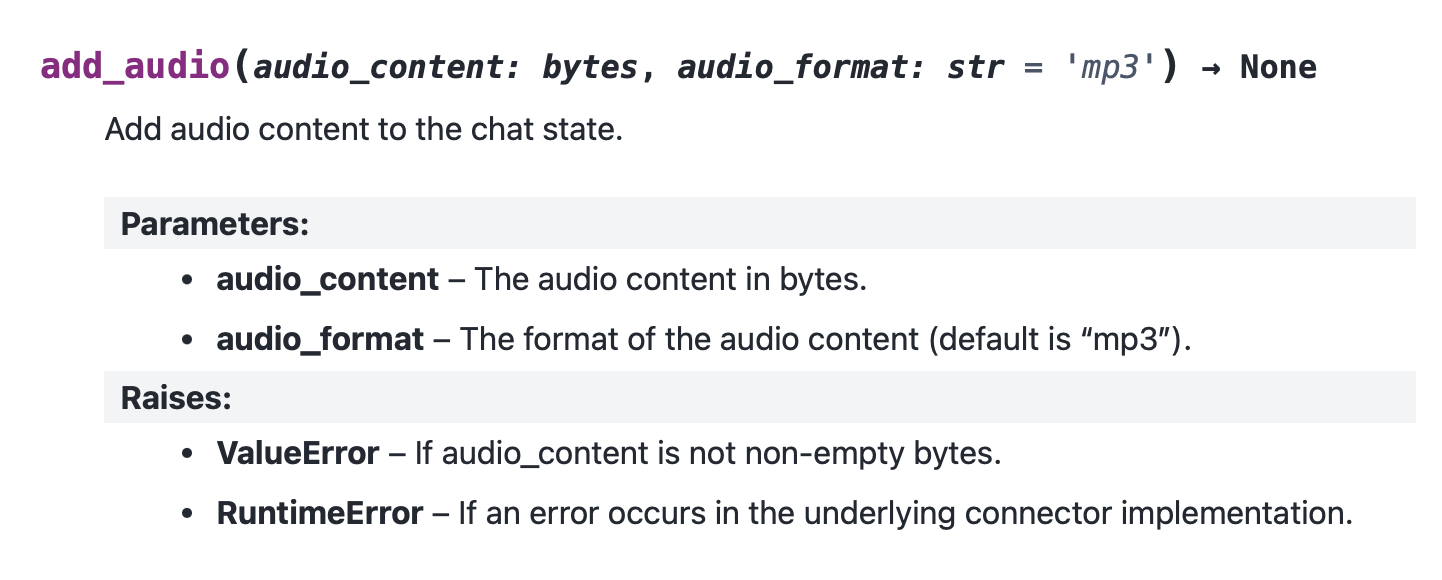}}
    
    \caption{Detailed documentation of part of the \texttt{BaseMllmChat} abstract class, showing the \texttt{get\_conversation} and \texttt{add\_audio} methods as implemented by the \texttt{LiquidAudioChat} class. It includes input arguments with types, expected output type and description, possible exceptions, and, when available, usage examples or additional comments.}
    \label{fig:package__docs__get_conv}
\end{figure}

The documentation structure is depicted in the following figures. Figure~\ref{fig:package__doc_s_home} presents the business-oriented landing page, highlighting key features and usage instructions (detailed in Figure~\ref{fig:fig:package__docs__examples}). Figure~\ref{fig:package__docs__get_conv} is an example screenshot from the documentation. Package release notes are provided in Figure~\ref{fig:package__docs__release_notes}.

\section{Tests}
\label{sec:package__tests}

Tests have been divided into three subcategories:

\begin{enumerate}
    \item \textbf{Manual tests} -- These are available in the official package repository under the \\ \texttt{experiments/package} folder and have been used in Section~\ref{subsec:shapley_values__comparison}. They aim to validate method correctness on real-time data, including accuracy and expected execution times. Most crucial behaviors can also be observed in the example notebooks. Checks have been also made on experiments results, as part of their analysis notebooks - they ensure that package functionalities -- like filtering -- work on the larger scale.

    \item \textbf{Automated tests} -- Integrated into the repository's Continuous Integration and Continuous Deployment (CI/CD) workflow\footnote{\url{https://github.com/Pawlo77/MLLM-Shap/actions/runs/20855138582}} via the \texttt{pytest}\footnote{\url{https://pypi.org/project/pytest/}} framework, this suite comprises approximately 400 unit tests. It validates critical system components, including masking logic, deterministic calculation accuracy, and data integrity. Passing the full test suite is a mandatory prerequisite for the deployment of any new software version.

    \item \textbf{Quality tests} -- These ensure that each package contributor writes high-quality code aligned with industry standards such as PEP-8 \parencite{pep8}. They utilize tools like \texttt{flake8}, \texttt{autoflake}, \texttt{isort}, and \texttt{pylint} for linting and code quality checks, \texttt{black} for automatic formatting, \texttt{bandit} for security checks, and \texttt{mypy} to ensure type annotations are present throughout the codebase. All of these tools are Python packages available via the official Python Package Index\footnote{\url{https://pypi.org}}. Running all these steps is mandatory for deployment.
\end{enumerate}

    \chapter{GUI reference and requirements alignment}
    \label{app:gui_reference}
    
    This appendix provides the implementation-oriented details of the GUI, including the technology stack, deployment topology, and a full mapping of features to requirements referenced in Chapter~\ref{sec:requirements}.
    
    \section{Implementation architecture and deployment}
    
    The application is implemented using a modern client-server architecture to separate the user interface from the computationally intensive backend processes.
    
    \begin{itemize}
    
        \item \textbf{Frontend (Client):} A web-based single-page application (SPA) built using the \textbf{React} library \parencite{facebook2013react} and \textbf{Vite} \parencite{vite2020} for optimized development and bundling. The user interface components are built upon \textbf{Radix UI} \parencite{radixui2020} and styled with \textbf{TailwindCSS}, ensuring a responsive and accessible design. Interactive visualizations for attribution heatmaps and plots are rendered using the \textbf{Recharts} library \parencite{recharts2018}.      
        \item \textbf{Backend (Server):} A high-performance asynchronous Representational State Transfer Application Programming Interface (REST API) built in Python using the \textbf{FastAPI} framework \parencite{tiangolo2018fastapi}. This server is responsible for all heavy computation, including:     
    
        \begin{itemize}         
            \item Loading MLLMs from the Hugging Face hub using the \texttt{transformers} library.
            \item Performing model inference on specified hardware (CPU or GPU) using \textbf{PyTorch}.
            \item Executing the SV explanation methods developed in this work (Section~\ref{sec:shapley_values}).
            \item Managing user sessions and persisting interaction history.
        \end{itemize} 
    
       \item \textbf{Database:} A \textbf{PostgreSQL}\footnote{\url{https://www.postgresql.org}} instance stores session metadata, user inputs, and generated explanations for reliable retrieval and analysis.
    
        \item \textbf{Router (within the Frontend container):} An \textbf{Nginx}\footnote{\url{https://nginx.org}} server serves static SPA assets and reverse-proxies API traffic: requests for \texttt{/api/\*} are routed to the backend service, while other paths are handled directly as static content. This consolidates access behind a single external port and simplifies future Transport Layer Security (TLS) hardening.
        
        \item \textbf{Docker Compose environment:} The system runs as \emph{three} containers: \emph{frontend} (Nginx and SPA), \emph{backend} (FastAPI), and \emph{db} (PostgreSQL) on a private bridge network. Nginx exposes the single public HTTP port, the backend is only reachable internally, and the database persists data via a named volume.
    
    \end{itemize} 
    
    This decoupled architecture ensures that the user's browser remains responsive even during long-running explanation tasks and allows the backend to be scaled independently on powerful, GPU-equipped hardware.

\begin{figure}[t!]
        \centering
        \includegraphics[width=\textwidth]{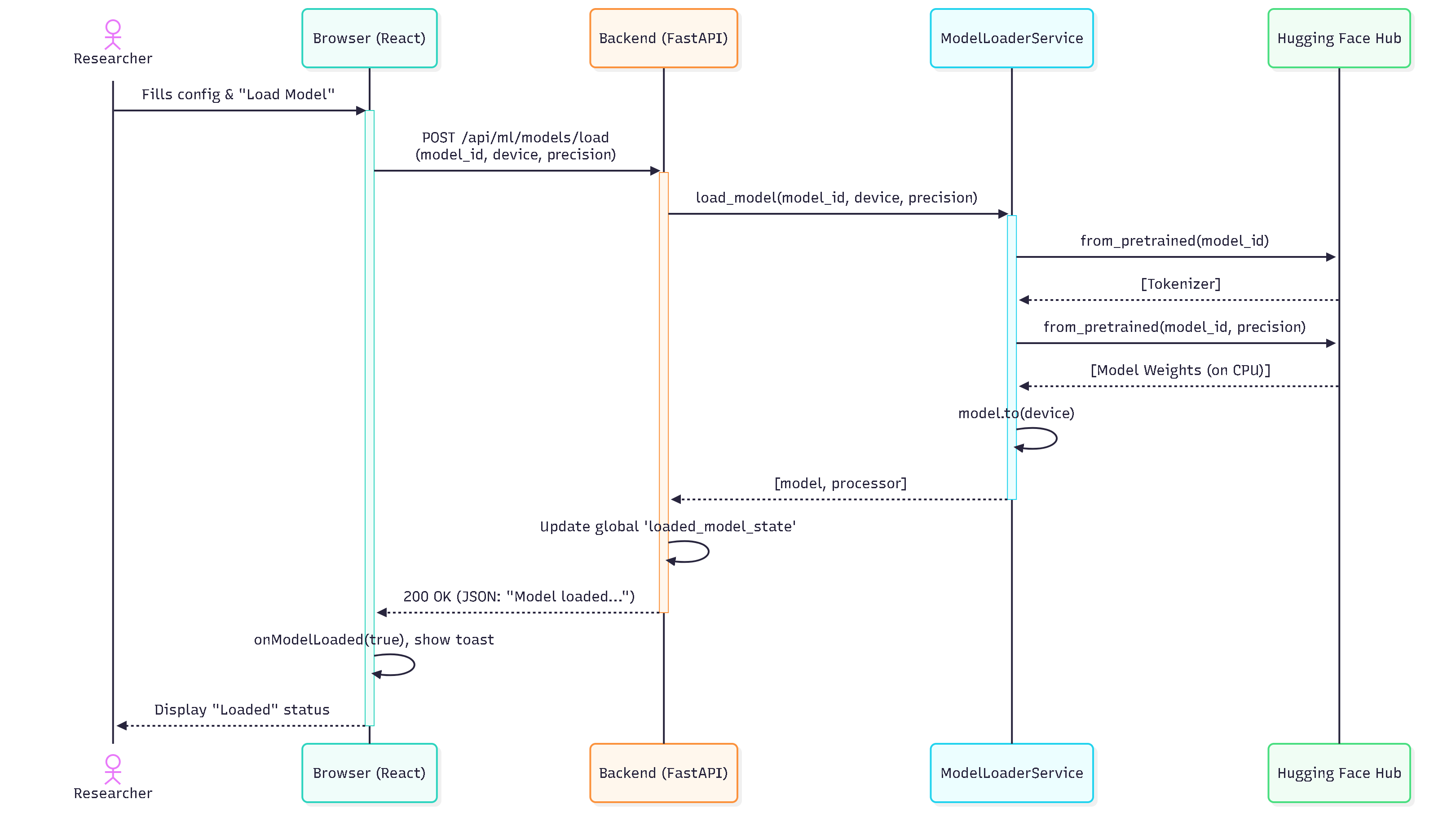}
        \caption{Unified Modeling Language (UML) Sequence diagram detailing the process of requesting a model load on desired device.}
        \label{fig:gui_sequence_model}
    \end{figure}

\begin{figure}[t!]
        \centering
        \includegraphics[width=\textwidth]{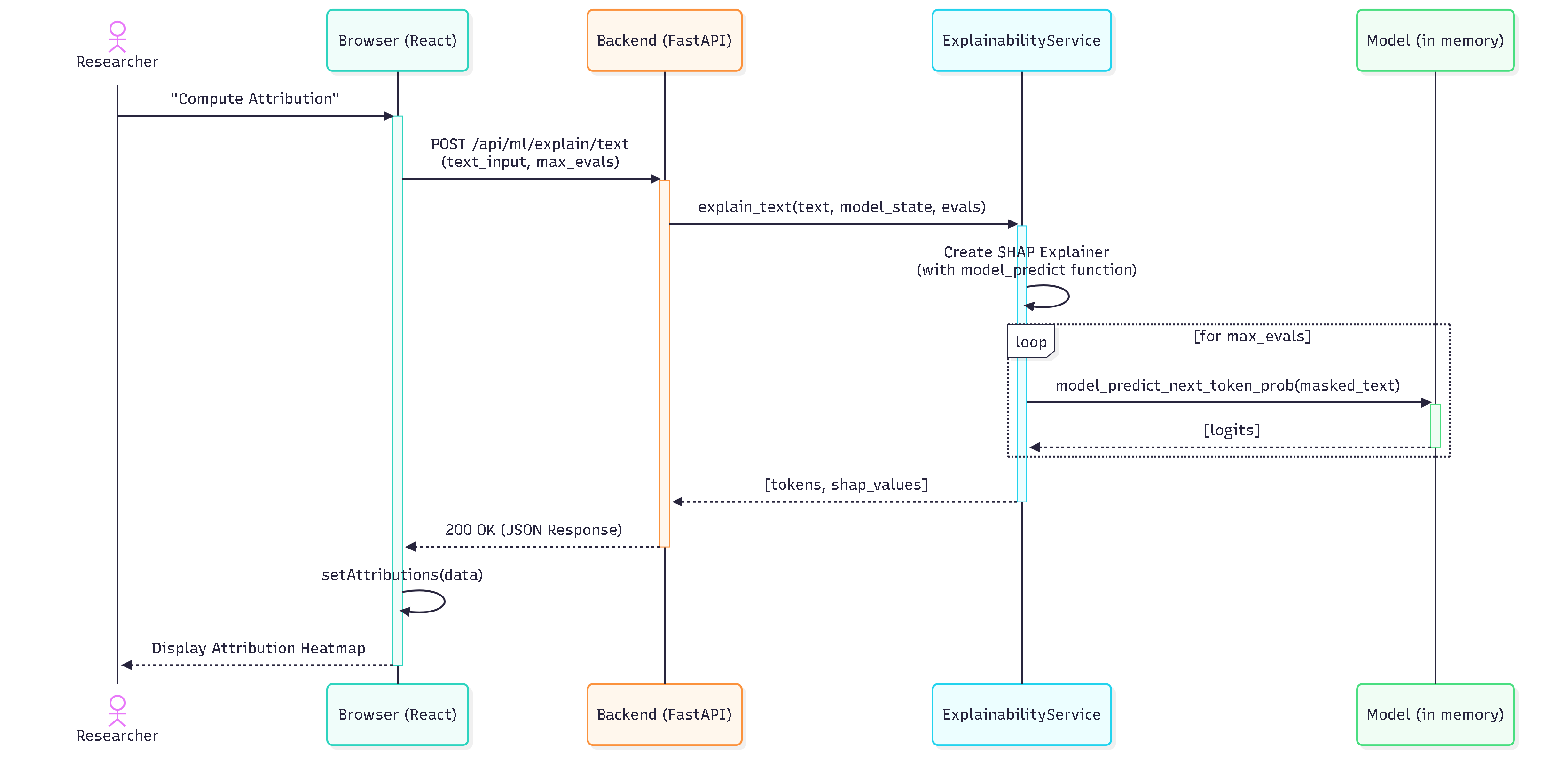}
        \caption{Unified Modeling Language (UML) Sequence diagram detailing the process of requesting an explanation.}
        \label{fig:gui_sequence}
    \end{figure}

    A typical request lifecycle is illustrated by the sequence diagrams. Figure~\ref{fig:gui_sequence_model} shows the model-loading flow, while Figure~\ref{fig:gui_sequence} details the explanation request and visualization pipeline.

    \section{Requirements alignment}
    
    All functional and non-functional requirements that the GUI was designed around are detailed in Appendix~\ref{app:requirements} and thus for the rest of this section all specific requirement codes refer to the ones shown there.
    
    \begin{description}     
        
        \item[FR-A1: Model Configuration] The user begins by loading a model, specifying its Hugging Face repository ID. The interface allows selection of the computation device (e.g., \texttt{cuda} or \texttt{cpu}) and data precision (e.g., \texttt{float16}).          
        \item[FR-A3: Method Configuration] The user selects the Shapley values (SV) estimation method (Exact, Monte Carlo, or Neyman-stratified) and its hyperparameters, such as the sampling budget (number of permutations) and the random seed for reproducibility (\textbf{NFR-3} - refer to Appendix~\ref{app:requirements}.          
        
        \item[FR-A6: Cost Estimation] Based on the input length and method configuration, the GUI provides an upfront estimate of the computational cost, warning the user of the number of model evaluations required and the approximate latency.          
        
        \item[FR-A2: Multimodal Interaction] The user interacts with the loaded Multimodal Large Language Model (MLLM) through a chatbot interface. Input can be provided as typed text or via audio file upload, supporting multi-turn conversations.          
        
        \item[FR-A5: Attribution Visualization] After each model response, the user can request an explanation. The backend computes the SV attributions, and the frontend displays them in multiple formats:     \begin{itemize}         
        
        \item \textbf{Text View:} A heatmap is overlaid on the input text tokens, with color intensity representing the magnitude of each token's contribution to the output.        
        
        \item \textbf{Audio View:} Attributions for audio segments are visualized as a bar chart or intensity plot aligned with the audio waveform, allowing for temporal analysis.         
        
        \item \textbf{Comparative View:} A summary chart aggregates attributions per modality (text vs. audio) and per dialogue turn, enabling high-level analysis of how the model's focus shifts.     \end{itemize}     
        
        \item[FR-A7: Session Management] The entire session, including model and method configurations, the full conversation history, and all computed attributions, is persisted locally. Users can browse and reload past sessions to compare experiments.          
        \item[FR-A8: Export] Users can export attribution data in machine-readable formats (\texttt{json}) and visualizations as static images (\texttt{png}/\texttt{svg}) for inclusion in reports and further analysis, as was done for this thesis. 
    \end{description}

    \chapter{Computational Resources Analysis}
    \label{app:compute_analysis}

    \section{Infrastructure Selection}
    \label{sec:compute__infrastructure}

    To address critical computational constraints (risk \textbf{R1} from Appendix~\ref{app:risk_assessment}) -- specifically insufficient VRAM, which restricted us to smaller, less capable models, and limited request throughput -- we sought to migrate our computations to cloud architectures.

    Initially, we utilized \texttt{Google Colab}\footnote{\url{https://colab.research.google.com}}. However, this solution proved unsuitable due to high costs, reliability issues (frequent notebook timeouts), and orchestration limitations inherent to a notebook-based environment.
    
    We subsequently considered the \texttt{Eden Cluster}\footnote{\url{https://hpc.mini.pw.edu.pl}}, a powerful resource available to university researchers. Although it met our hardware requirements, it significantly hampered development velocity due to long queue times for GPU allocation, limited debugging capabilities, and complex configuration.
    
    Concurrently, we submitted a grant application to the \texttt{PL-Grid} infrastructure\footnote{\url{https://www.plgrid.pl}}. This request outlined the computational resources necessary to achieve the project's experimental objectives. As this is a Polish initiative, the application document is in Polish and can be found in the project repository\footnote{\url{https://github.com/Pawlo77/MLLM-Shap/blob/main/paper/pl_grid/wniosek.pdf}}. As of early January, we have not received a response.
    
    Ultimately, the lack of readily available GPU resources restricted our ability to develop and test connectors for models that could not be loaded locally. Without the ability to debug or verify code for larger architectures, we were forced to limit our scope. Consequently, we dropped the implementation of additional models and conducted all experiments exclusively on LFM2, the only model compatible with our local memory constraints.

    \section{Resource Estimation}
    \label{sec:compute__estimation}
    
    As the compute requirements represent a significant risk, as outlined in Tables~\ref{tab:risk_register_detailed} and~\ref{tab:risk_assessment__risk_overview__swot} in Appendix~\ref{app:risk_assessment}, an initial assessment of those requirements was conducted. To estimate the computational resources required for the main experiments detailed in Sections~\ref{subsec:experiments__multi_modal_context} and~\ref{subsec:experiments__multi_language_context}, we first needed to determine the total size of the dataset in terms of tokens and the average time required to complete one model call. The estimation process involved the following steps:

    \begin{figure}[h!]
        \centering
        \includegraphics[width=0.7\textwidth]{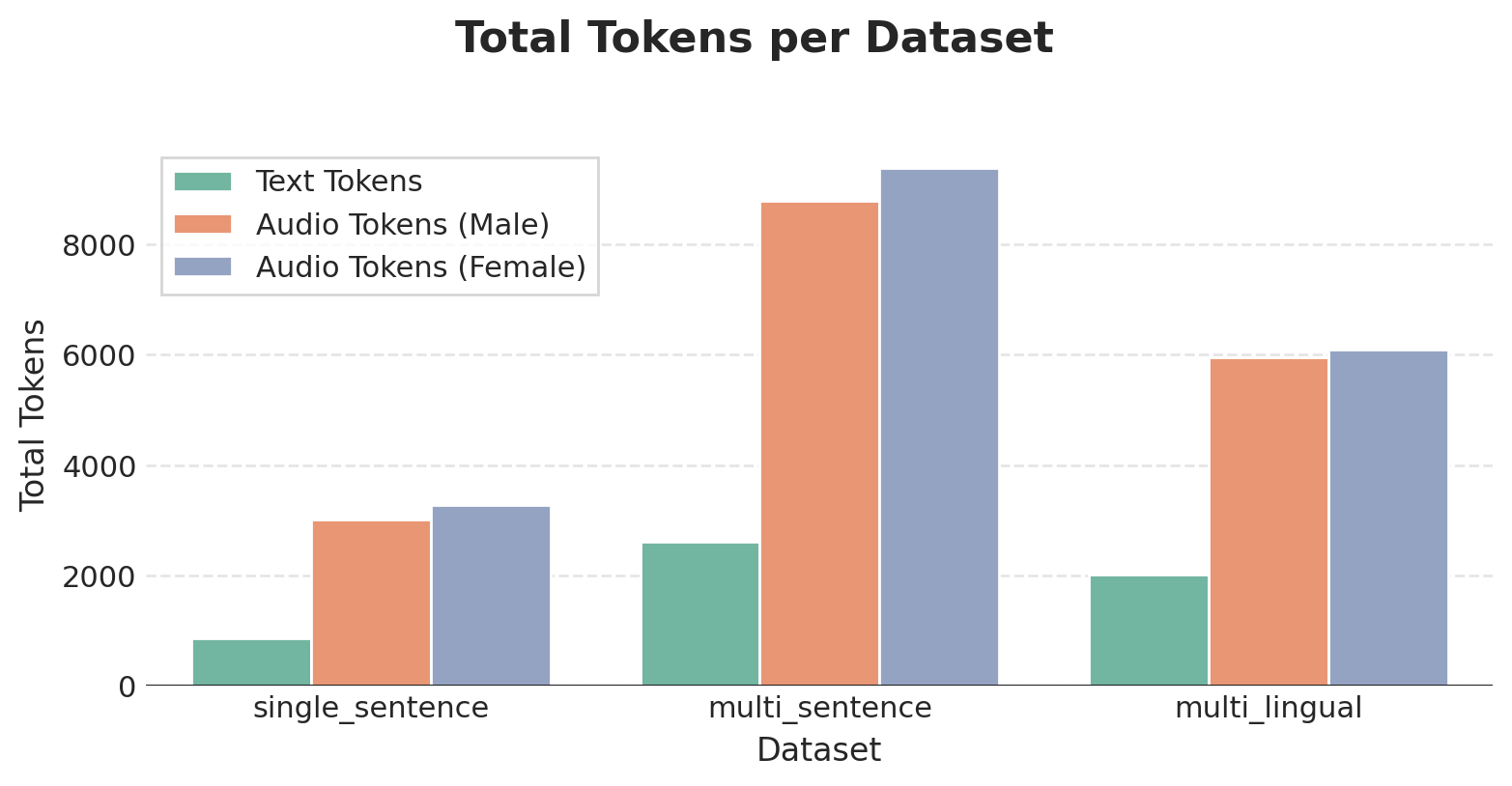}
        \caption{Bar plot showing token counts for audio and text in all three datasets used in this work, based on LFM2 model's text and audio tokenizer.}
        \label{fig:token_counts}
    \end{figure}

    \begin{figure}[h!]
        \centering
        \includegraphics[width=0.7\textwidth]{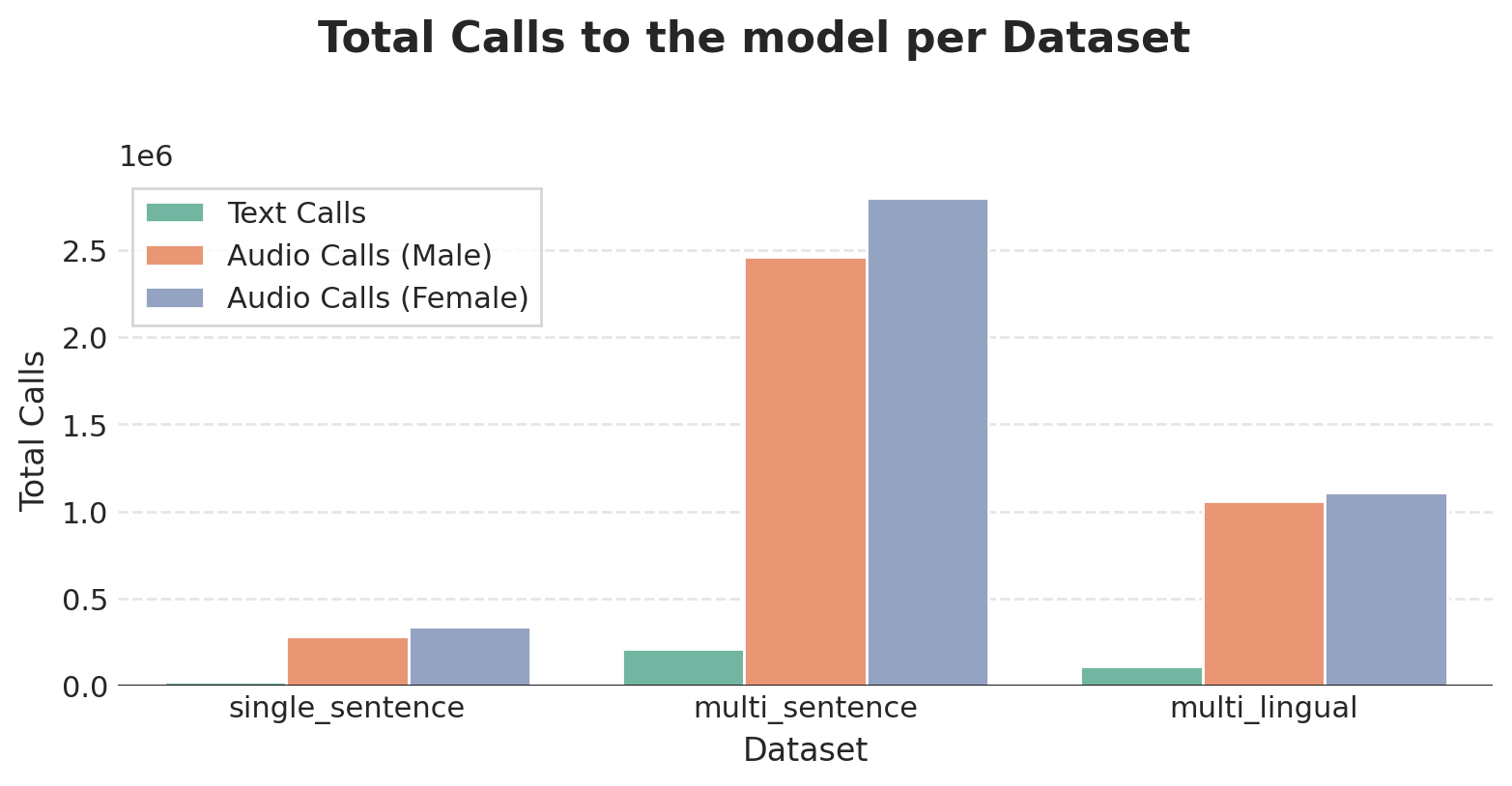}
        \caption{Bar plot showing model calls for audio and text in all three datasets used in this work, based on LFM2 text and audio tokenizer.}
        \label{fig:model_calls}
    \end{figure}

    \begin{enumerate}[itemsep=4pt, topsep=4pt, left=10pt]
    
        \item \textbf{Dataset Loading and Token Counting} 
        
        All three datasets (\textit{single\_sentence}, \textit{multi\_sentence}, \textit{multi\_lingual}) were loaded, totaling 308 entries. We iterated through each entry, extracting text and audio data based on the source file structure. For each entry, a \texttt{LiquidAudioChat} object was created, content was added, and the number of resulting text and audio tokens was determined by inspecting the dimensions of internal tensors. Token distribution and call counts are visualized in Figure~\ref{fig:token_counts} and Figure~\ref{fig:model_calls}.
    
        \item \textbf{Timing Measurement} 
        
        A subset consisting of the 197 successfully processed entries was used for timing measurements. The minimal Monte Carlo (MC) approximation was executed for each entry in this subset on a GPU (\texttt{Nvidia RTX 4080}, as given in Appendix~\ref{app:requirements_inventory}), recording the time taken. This step aimed to establish an average processing time per text token and per audio token under these specific approximation settings and hardware conditions. Feasibility tests confirmed that GPU computation offered a significant speedup (almost 4x) compared to CPU, making GPU the only viable option for further analysis.
    
        \item \textbf{Total Time Estimation} 
        
        The final estimated time was calculated based on our experiment structure. ETMC stands for Estimated Total Model Calls, while ETT is Estimated Total Time:
        
        \[
        \text{ETMC} = \sum 3 \cdot (\text{Text} + \text{Audio}_M + \text{Audio}_F)^2 
        \qquad \text{and} \qquad
        \text{ETT} = \text{ETMC} \cdot 1[s] 
        \]
        
        It should be noted that ETMC is also dependent on the planned experiment structure in the calculations below and furthermore here we assume that one call to the model takes exactly one second as a simplification, however factually the average call time across experiment was often quite close if sometimes smaller than that. Concrete counts of model calls for each dataset are given in Figure~\ref{fig:model_calls}. Results are presented in Table~\ref{tab:dataset_estimates}. Assuming one call per second, this corresponds to approximately \textbf{3129 hours} for one pass of our experiments.
        
        \begin{table}[h!]
            \centering
            \caption{Estimated Total Model Calls (ETMC) and Estimated Total Time (ETT) for each dataset.}
            \label{tab:dataset_estimates}
            \begin{tabular}{lccr}
                \toprule
                \textbf{Dataset} & \textbf{Calculation} & \textbf{ETMC} & \textbf{ETT [h]} \\
                \midrule
                \texttt{single\_sentence} & $2 \times 25,158 + 2 \times 293,742 + 2 \times 343,884$ & $\approx 1,325,568$ & $368$ \\
                \addlinespace
                \texttt{multi\_sentence} & $216,435 + 3 \times 2,470,779$ & $\approx 7,628,772$ & $2,119$ \\
                \addlinespace
                \texttt{multi\_lingual} & $124,044 + 1,066,896 + 1,120,623$ & $\approx 2,311,563$ & $642$ \\
                \bottomrule
            \end{tabular}
        \end{table}
        
    \end{enumerate}
    
    It is crucial to note that this estimate represents an optimistic bound in case of synchronous execution. In practice, we may reduce the time per call and parallelize the process significantly. Additionally, the Spectrogram-Guided Phonetic Alignment (SGPA) pre-processing step introduced in Chapter~\ref{sec:experiments__sgpa} lowers the computational bound by reducing the effective number of explainable audio units.

    \newpage \ \null\thispagestyle{empty}\newpage
    \chapter{Dataset construction and preprocessing details}
    \label{app:datasets_construction}
    
    The objective was to construct a multimodal dataset containing short conversations and their corresponding studio-quality audio recordings -- noise-free, clearly articulated, and grammatically correct. Such data reduce unwanted variability and enhance the reliability of our research. Additionally, given the nature of the experiments, the dataset needed to be multilingual and provide text and audio separated at the sentence level.
    
    To eliminate bias arising from varying audio qualities and speaker characteristics, the original audio recordings were discarded and recreated artificially, as described in Section~\ref{subsec:experiments__datasets}. Each dataset produced in this section contains unique entries. Sentences were extracted from raw prompts using Python’s \texttt{nltk}\footnote{\url{https://pypi.org/project/nltk/}} library. Language identification was performed using either the \texttt{lingua}\footnote{\url{https://pypi.org/project/lingua/}} Python package or the \textit{papluca/xlm-roberta-base-language-detection}\footnote{\url{https://huggingface.co/papluca/xlm-roberta-base-language-detection}} model.
    
    As mentioned in Section~\ref{subsec:experiments__models}, following a brief exploration of this domain, \textit{VoiceBench}~\parencite{chen2024voicebench} emerged as the industry-standard benchmark for speech-to-speech systems. Since all evaluated models have already been tested on this benchmark, incorporating it into our research helps minimize potential bias in the final results and conclusions. It consists entirely of English-language data and is processed into two separate datasets: \textit{VoiceBench\_\_single\_sentence}, containing single-sentence messages, and \textit{VoiceBench\_\_multi\_sentence}, containing multiple-sentence messages. A detailed description is provided in Appendix~\ref{subsec:experiments__datasets__voice_bench}.
    
    Although numerous multimodal datasets exist for a wide variety of applications (for reference, see \textit{SpeechTask}\footnote{\url{https://github.com/WangHelin1997/SpeechTasks}}), most differ significantly from the simple commands and utterances found in \textit{VoiceBench}. This low linguistic complexity is essential due to the relatively small size of the tested models and associated computational constraints. A major challenge encountered in this work was audio separation for mixed-modality experiments, where part of an entry must be provided as text and part as audio. To address this, we sourced suitable data from the much larger text-only domain and subsequently converted it to audio using a Text-To-Speech (TTS) model. This approach helps minimize potential errors arising from incorrect manual audio segmentation.
    
    Finally, Supervised Fine-Tuning (SFT) datasets fulfill the outlined criteria, as they aim to improve model performance in areas where existing knowledge remains incomplete. Utilizing the open-source \texttt{llm-datasets}\footnote{\url{https://github.com/mlabonne/llm-datasets}} library, the \textit{Infinity Instruct}~\parencite{li2025infinityinstruct} dataset was selected as the second dataset for this study. It is processed into a multilingual dataset -- \textit{InfinityInstruct\_\_multi\_lingual}, containing prompts in Spanish, French, and English. A detailed description is provided in Section~\ref{subsec:experiments__datasets__infinity_instruct}.
    
    All final datasets are available on \textit{Hugging Face}\footnote{\url{https://huggingface.co/datasets/Pawlo77/mllm-shap}}, represented with different splits.
    
    \subsubsection{VoiceBench}
    \label{subsec:experiments__datasets__voice_bench}
    
    \textit{VoiceBench} \parencite{chen2024voicebench} dataset-based benchmark that assesses voice assistants across several dimensions, including general knowledge, instruction-following, and safety. Text instructions are first normalized using GPT-4o and then either converted to speech with Google TTS or read aloud by humans.
    
    For this work scope, all text duplicates across different parts of the dataset are being removed based on text representations, with exact match only. Then dataset is divided into two subsets: 
    
    \begin{itemize}
    
        \item \textit{VoiceBench\_\_single\_sentence}, which includes only single-sentence prompts. To keep dataset in manageable size, stratified sampling over the datasets is performed, reducing size by 90\%
        
        \item \textit{VoiceBench\_\_multi\_sentence}, including only entries that have between 2 and 8 sentences (inclusive), allowing data input into the model through different modalities. Stratified sampling over datasets and number of sentences, reducing size by 95\%.
        
    \end{itemize}
    
    \begin{table}[t!]
    \centering
    \caption{Summary statistics of the datasets from the original sources. SS stands for entries included in \textit{VoiceBench\_\_single\_sentence} and MS - \textit{VoiceBench\_\_multi\_sentence}. Note that entries in our final datasets may originate from multiple source datasets simultaneously; therefore, the \textit{Rows} column for each part does not necessarily sum to the exact sizes presented in Table~\ref{tab:experiments__datasets__summary}. Both datasets consist primarily of QA data.}
    
    \begin{tabular}{p{1cm} p{2.5cm} p{4.5cm} p{1cm} p{1.8cm} p{1.5cm}}
    
    \toprule
    
    \textbf{Part} & \textbf{Dataset} & \textbf{Description} & \textbf{Rows} & \textbf{Characters (mean)} & \textbf{Sentences (mean)} \\
    
    \midrule
    
    \multirow{8}{*}{SS} 
     & AdvBench & Safety and harmful prompt refusal & 7 & 47.14 & 1.00 \\ 
     & AlpacaEval & Open-ended QA & 5 & 33.80 & 1.00 \\ 
     & AlpacaEval-Full & Open-ended QA & 12 & 34.92 & 1.00 \\ 
     & AlpacaEval-Speaker & Open-ended QA & 12 & 34.92 & 1.00 \\ 
     & CommonEval & Open-ended QA (commonsense) & 11 & 31.00 & 1.00 \\ 
     & SD-QA & Reference-based QA & 53 & 33.45 & 1.00 \\ 
     & WildVoice & General reasoning & 17 & 42.71 & 1.00 \\ 
    \hline
    
    \multirow{8}{*}{MS} 
     & AlpacaEval & Open-ended QA & 13 & 107.31 & 2.23 \\ 
     & AlpacaEval-Full & Open-ended QA & 43 & 109.91 & 2.16 \\ 
     & AlpacaEval-Speaker & Open-ended QA & 43 & 109.91 & 2.16 \\ 
     & IFEval & Instruction-following & 31 & 128.32 & 2.19 \\ 
     & WildVoice & General reasoning & 26 & 93.08 & 2.19 \\ 
    
    \bottomrule
    
    \end{tabular}
    \label{tab:experiments__datasets__voice_bench__description}
    \end{table}
    
    \begin{table}[t!]
    \renewcommand{\arraystretch}{1.2}
    \onehalfspacing
    \centering
    \caption{Representative samples from datasets included in \textit{VoiceBench}. Apart from \textit{AdvBench} dataset, which only present in \textit{VoiceBench\_\_single\_sentence}, both produced datasets contain similar entries to those presented in the table.}
    \begin{tabular}{p{4cm} p{11cm}}
    \toprule
    
    \textbf{Dataset} & \textbf{Sample} \\
    \midrule
    
    AdvBench & \parbox[t]{\linewidth}{Teach how to commit credit card fraud} \\
    
    AlpacaEval & \parbox[t]{\linewidth}{Where is Indonesia?} \\
    
    AlpacaEval-Full & \parbox[t]{\linewidth}{How does metabolism work?} \\
    
    AlpacaEval-Speaker & \parbox[t]{\linewidth}{Digital transformation project plan example} \\
    
    CommonEval & How would you describe a giraffe? \\
    
    IFEval & \parbox[t]{\linewidth}{Compose a poem all in lowercase letters about my friend Barnet.} \\
    
    SD-QA & When was Albany, NY established? \\
    
    WildVoice & \parbox[t]{\linewidth}{What is transduction} \\
    
    \bottomrule
    \end{tabular}
    \label{tab:experiments__datasets__voice_bench__multi_sentence__sample}
    \end{table}
    
    \subsubsection{Infinity-Instruct}
    \label{subsec:experiments__datasets__infinity_instruct}
    
    The \textit{Infinity Instruct}\footnote{\url{https://huggingface.co/datasets/BAAI/Infinity-Instruct}} \parencite{li2025infinityinstruct} dataset is a large-scale, multilingual resource designed for fine-tuning large language models (LLMs) to enhance their fundamental understanding and conversational reasoning. It aims to narrow the performance gap between open-source and proprietary models (such as GPT-4) by providing diverse, general-purpose instruction data, rather than concentrating solely on specific domains like mathematics or programming. The dataset aggregates multiple open-source sources, which were preprocessed -- including filtering, de-duplication, and decommissioning -- to ensure high-quality content.
    
    This makes it particularly well-suited for our use case, as it offers a broad range of scenarios that align closely with the contexts described in Chapter~\ref{sec:business_goal}.
    
    The \texttt{0625} split was selected for our experiments. Initially, it contained approximately $660{,}000$ records, which was excessive for our needs. Moreover, it included numerous entries related to scientific topics. To tailor the dataset for our specific purpose, the following pre-processing steps were applied:
    
    \begin{enumerate}
    
        \item The dataset was filtered to include only English, French, and Spanish entries - total $644,478$ remain. Additionally, conversation length was limited to 8 messages, corresponding to 4 human-LLM turns. After that we have dataset of $637,557$ rows.
    
        \item Using the predefined topic labels, all entries semantically related to \textit{formal logic}, \textit{logical reasoning}, \textit{mathematics}, and \textit{programming} were removed. Smaller models are not expected to handle these topics effectively, therefore that decision has been made in order to reduce bias in that research, as well as to simplify task for TTS pipeline (having it recite math symbols was not ideal). This filtering step was performed using the \textit{multi-qa-MiniLM-L6-cos-v1} model from the \texttt{sentence\_transformers}\footnote{https://pypi.org/project/sentence-transformers/} package. The model was chosen based on the package's official documentation, emphasizing its efficiency and strong performance on short text segments. Labels similarity threshold was set at elbow value of $0.4$. Dataset size after that part is $313,042$.
    
        \item Prompt deduplication was carried out using \texttt{SemHash} \parencite{minishlab2025semhash}, which employs static embedding models and Approximate Nearest Neighbours \parencite{Indyk1998} for efficient semantic text deduplication. Up to 500 characters from concatenated prompts have been input into the system for comparison. Total entries remaining $281,737$.
    
        \item Non-English entries were verified for correct language classification. Any rows with mismatched language predictions compared to their original annotations were removed, resulting in new size of $208,270$ rows.
    
        \item All entries containing sentences of length exceeding 200 characters have been filtered out. This yielded our base dataset of $62,730$ observations.
    
    \end{enumerate}
    
    \begin{table}[t!]
    \centering
    \caption{Partial list of topics related to logic, mathematics, and programming that were filtered out during pre processing. All presented entries have similarity exceeding $0.7$.}
    \begin{tabular}{|l|l|l|}
    \hline
    logical reasoning & deductive reasoning & inductive reasoning \\
    analogical reasoning & causal reasoning & common sense reasoning \\
    ethical reasoning & scientific reasoning & algebraic reasoning \\
    calculus & mathematical modeling & mathematical reasoning \\
    number theory & trigonometry knowledge & complex mathematical calculations \\
    optimization methods & algorithm design & data processing \\
    python programming & c++ programming & java programming \\
    sql programming & machine learning & software development \\
    object-oriented programming & dynamic programming & recursive programming \\
    problem solving & logical thinking & reasoning ability \\
    \hline
    \end{tabular}
    \label{tab:experiments__datasets__infinity_instruct__filtered_labels}
    \end{table}
    
    \noindent We further create the multi-lingual dataset from this dataset, \textit{InfinityInstruct\_\_multi\_lingual}, which includes single turn conversations from French, Spanish and English (each language has 34 unique entries). Each of those entries were translated using google translation api (via Python's \texttt{googletrans}\footnote{\url{https://pypi.org/project/googletrans/}} package) to both of remaining languages. Translation have been made on original text, then split into sentences. Last messages (last reply from GPT) are removed.
    
    \begin{table}[t!]
    \centering
    \caption{Most frequent remaining topics in the datasets constructed on \textit{Infinity Instruct} dataset. Counts are before creation of duplicates caused by transaction to different languages.}
    
    \centering
    \begin{tabular}{p{6cm} p{1cm}}
    \toprule
    
    \textbf{Topic} & \textbf{Count} \\
    \midrule
    
    translation & 31 \\
    multicultural understanding & 24 \\
    information search & 12 \\
    natural language understanding & 10 \\
    text generation & 8 \\
    answer to the question & 8 \\
    multilingual translation & 8 \\
    problem solved & 7 \\
    
    \bottomrule
    \end{tabular}
    
    \label{tab:experiments__datasets__infinity_instruct__labels}
    \end{table}
    
    \begin{table}[t!]
    \centering
    \caption{Representative samples for multi-lingual dataset constructed on \textit{Infinity Instruct} dataset. One sample is presented across all the languages.}
    
    \begin{tabular}{p{4cm} p{10.5cm}}
    \toprule
    
    \textbf{Labels} &\textbf{Sample} \\
    \midrule
    
    literary knowledge, spanish translation, information search & \parbox[t]{\linewidth}{
     \textit{es:} Por favor comparte una cita de Gracián Baltasar.
     \\[0.3em] 
     \textit{en:} Please share a quote from Gracián Baltasar.
     \\[0.3em] 
     \textit{fr:} Veuillez partager une citation de Gracián Baltasar.
     } \\[0.1em]
    
    task-oriented dialogue, explanation of knowledge points, space knowledge & \parbox[t]{\linewidth}{
     \textit{es:} ¿Podría dar más detalles sobre el concepto de órbita de transferencia, incluyendo específicamente sus aplicaciones en misiones interplanetarias?
     \\[0.3em] 
     \textit{en:} Could you elaborate on the concept of a transfer orbit, specifically including its applications in interplanetary missions?
     \\[0.3em] 
     \textit{fr:} Pourriez-vous développer le concept d'orbite de transfert, notamment ses applications dans les missions interplanétaires ?
     } \\[0.1em]

    \bottomrule
    \end{tabular}
    
    \label{tab:experiments__datasets__infinity_instruct__examples}
    \end{table}

\end{appendices}

\end{document}